\documentclass[journal]{IEEEtran}

\usepackage{cite}

\ifCLASSINFOpdf
  \usepackage[pdftex]{graphicx}
\else
  \usepackage[dvips]{graphicx}
\fi

\usepackage[cmex10]{amsmath}
\usepackage{amssymb}
\usepackage{float}
\usepackage{bm}
\usepackage{extarrows}
\usepackage{algorithm}
\usepackage{epstopdf}
\usepackage{mathtools}
\usepackage{mleftright}
\usepackage{color}
\usepackage[noend]{algpseudocode}
\makeatletter
\def\BState{\State\hskip-\ALG@thistlm}
\makeatother

\usepackage{tikz}
\usetikzlibrary{shapes,arrows,fit,positioning,shadows,calc}
\usetikzlibrary{plotmarks}
\usetikzlibrary{decorations.pathreplacing}
\usetikzlibrary{patterns}
\usetikzlibrary{automata}
\usepackage{pgfplots}
\pgfplotsset{compat=newest}
\usepackage{adjustbox}
\usepackage{caption}
\newcommand{\vertiii}[1]{{\left\vert\kern-0.25ex\left\vert\kern-0.25ex\left\vert #1
 \right\vert\kern-0.25ex\right\vert\kern-0.25ex\right\vert}}

\makeatletter

\newcommand{\Rmnum}[1]{\expandafter\@slowromancap\romannumeral #1@}
\makeatother

\newtheorem{theorem}{Theorem}

\newtheorem{remark}{Remark}

\newcommand*{\QEDB}{\hfill\ensuremath{\square}}%
\usepackage{footnote}
\makesavenoteenv{tabular}
\usepackage{hyperref}

\begin{document} 
%\linespread{1}
%\vspace{-0.5cm}
\title{Projection-based QLP Algorithm for Efficiently
Computing Low-Rank Approximation of Matrices}
\author{Maboud F. Kaloorazi,  \; Jie Chen, \IEEEmembership{Senior Member, IEEE} \\ 
\texttt{{kaloorazi}@xsyu.edu.cn}; \texttt{{jie.chen}@nwpu.edu.cn}}

\maketitle

%for Efficiently Computing Low-Rank Approximation of Matrices
\begin{abstract}
Matrices with low numerical rank are omnipresent
in many signal processing and data analysis applications. The
pivoted QLP (p-QLP) algorithm constructs a highly accurate
approximation to an input low-rank matrix. However, it is
computationally prohibitive for large matrices. In this paper,
we introduce a new algorithm termed Projection-based Partial
QLP (PbP-QLP) that efficiently approximates
the p-QLP with high accuracy. Fundamental in our work is the exploitation of
randomization and in contrast to the p-QLP, PbP-QLP does not
use the pivoting strategy. As such, PbP-QLP can harness modern
computer architectures, even better than competing randomized algorithms. The efficiency and effectiveness of our proposed PbP-QLP algorithm are investigated through various classes of synthetic and real-world data matrices.
\end{abstract}

\IEEEpeerreviewmaketitle

\section{Introduction}
\label{SectIntro}
\IEEEPARstart{M}{atrices} with low-rank structure are  omnipresent in signal processing, data analysis, scientific computing and machine learning applications including system identification \cite{FazelPST13}, subspace clustering \cite{CaiLi15}, matrix completion \cite{EftekhariYW18}, background subtraction \cite{AybatI15, MFKDeJSTSP18}, least-squares regression \cite{ClarWood2017}, hyperspectral imaging \cite{Rasti18, ErichsonMWK18}, anomaly detection \cite{FowlerDu12,KaDeICASSP17}, subspace estimation over networks \cite{ChenRS7859344}, genomics \cite{DarnellGME17, UbaruSaad19}, tensor decompositions \cite{7038247}, and sparse matrix problems \cite{DavisRS2016}.    
 
\stepcounter{footnote}\footnotetext {This paper has been accepted for publication in the IEEE Transactions on Signal Processing (submitted December 2019, accepted March 2021).}

\stepcounter{footnote}\footnotetext {Maboud F. Kaloorazi is with Xi'an Shiyou University, Xi'an, China. This work was done when he was a postdoctoral researcher with School of Marine Science and Technology, Northwestern Polytechnical University, Xi'an, China. J. Chen is with School of Marine Science and Technology, Northwestern Polytechnical University, Xi'an, China.}

\stepcounter{footnote}\footnotetext {Matlab code for this paper is available at \url{https://maboudfkaloorazi.com/codes/}}

The SVD (singular value decomposition) \cite{GolubVanLoan96}, CPQR (column-pivoted QR) \cite{Chan87}, \cite{GolubVanLoan96} and p-QLP (pivoted QLP) \cite{StewartQLP} provide a full factorization of an input matrix. The p-QLP is closely associated with CPQR, as it uses CPQR in its computation procedure. These methods are costly in arithmetic operations (though the last two are more efficient) as well as the communication cost, i.e., moving data between the slow and fast memory \cite{DemGGX15,DuerschGu2017, Dongarraetal18, MartinssonQH2019}. Traditional methods to compute a truncated version of the SVD, CPQR and p-QLP, though efficient in arithmetic cost, need to repeatedly access the data. This is a serious bottleneck, as the data access on advanced computational platforms is considerably more expensive compared with arithmetic operations.

Randomized methods \cite{Rokhlin09}, \cite{HMT2011}, \cite{Gu2015}, \cite{ClarWood2017}, \cite{MFKDeTSP18}, \cite{MFKDeJSTSP18}, \cite{Martinsson18}, \cite{Saibaba19}, \cite{MFKJC19}, \cite{MartinssonQH2019} provide an approximation to the foregoing deterministic decompositions. The general strategy of this type of methods is as follows: first, the input matrix is transformed to a lower-dimensional space by means of random sampling. Second, an SVD or CPQR is utilized to process the reduced-size matrix. Finally, the processed matrix is projected back to the original space.  Randomized methods are more efficient in arithmetic cost, highly effective, particularly if a low-rank approximation is desired, and harness parallel architectures better in comparison to their classical counterparts.

A bottleneck associated with existing randomized methods, however, is the utilization of the SVD or CPQR; as mentioned, standard methods to compute these factorizations encounter difficulties for parallel implementation, particularly for very large matrices. The goal of this paper is, thus, to devise a randomized matrix factorization method that averts this bottleneck by using only the QR factorization (without pivoting), that, compared with methods that utilize the SVD and CPQR, lends itself much readily to parallel processing.

\subsection{Our Contributions}
\label{subsec_I_Contrib}
In this paper, we devise a matrix factorization algorithm called PbP-QLP (projection-based partial QLP). PbP-QLP makes use of random sampling and the unpivoted QR factorization, and establishes an approximation to the p-QLP, primarily for matrices with low numerical rank. Given ${\bf A} \in \mathbb R^{n_1 \times n_2}$, a large and dense matrix with numerical rank $k$ and $n_1 \ge n_2$, and an integer $k \le d \le n_2$, PbP-QLP generates an approximation
$\hat{\bf A}_\text{PbP-QLP}$ such as:
\begin{equation}
\hat{\bf A}_\text{PbP-QLP}= {\bf QLP}^T,
\label{eq_PbPQLP}
\end{equation}
where ${\bf Q}\in \mathbb R^{n_1 \times d}$ and ${\bf P} \in \mathbb R^{n_2 \times d}$ are orthonormal matrices that constitute approximations to the numerical range of $\bf A$ and ${\bf A}^T$, respectively. ${\bf L}\in \mathbb R^{d \times d}$ is a lower triangular matrix, and its diagonals constitute approximations to the first $d$ singular values of $\bf A$. With theoretical analysis and numerical examples, we show that, analogous to the p-QLP, PbP-QLP is rank-revealing, that is, the rank of $\bf A$ is revealed in the $k \times k$ principal submatrix of $\bf L$. Due to incorporated randomization, as well as using only the unpivoted QR factorization as the decompositional tool, PbP-QLP (i) is efficient in arithmetic operations as its computation incurs the cost $\mathcal{O}(n_1n_2d)$, (ii) makes only a constant number of passes over $\bf A$, and (iii) can harness modern computing environments. We apply PbP-QLP to real data and data matrices from various applications.

\subsection{Notation}
We denote matrices/vectors by bold-face upper-case/lower-case
letters. For any matrix $\bf M$, ${\|{\bf M}\|_2}$ and ${\|{\bf M}\|_F}$ denote the spectral and the Frobenius norm, respectively. The $i$-th largest and the smallest singular value of $\bf M$ are indicated by $\sigma_i(\bf M)$ and  $\sigma_\text{min}({\bf M})$, respectively. The notations $\mathcal{R}({\bf M})$ and $\mathcal{N}({\bf M})$ indicate the numerical range and null space of $\bf M$, respectively. $\texttt{orth}({\bf M})$ constructs an orthonormal basis for the range of $\bf M$. $\texttt{qr}({\bf M})$, $\texttt{cpqr}({\bf M})$ and $\texttt{svd}({\bf M})$ give the unpivoted QR, column-pivoted QR and SVD of $\bf M$, respectively. $\texttt{randn}(\cdot)$ generates a matrix whose entries are independent, identically distributed Gaussian random variables of zero mean and variance one. ${\bf I}_r$ indicates an identity matrix of order $r$, and the dagger $\dagger$ indicates the Moore-Penrose inverse.

The remainder of this paper is organized as follows. Section
\ref{section_PriorWork} briefly surveys the related works. Section \ref{section_PbPQLP} describes our
proposed method (PbP-QLP). In Section \ref{section_PbPQLPAnalysis}, we establish a
thorough analysis for PbP-QLP. In Section \ref{sec_NumerExp}, we present and
discuss the results of our numerical experiments, and Section
\ref{section_concl} gives our conclusions.

\section{Related Works}
\label{section_PriorWork}
This section gives a review of related prior works: Section \ref{subsec_II_DeterMeth} presents traditional deterministic matrix factorization methods, while Section \ref{subsec_II_RandMeth} describes more recent methods
based on randomization for low-rank matrix approximations. In this work, we consider a matrix $\bf A$ defined in Section \ref{subsec_I_Contrib}.

\subsection{Deterministic Methods}
\label{subsec_II_DeterMeth}
Let matrix $\bf A$ have a decomposition as follows:
\begin{equation}
 {\bf A} = {\bf UTV}^T.
\label{eq_UTV_RelWork}
\end{equation}
\begin{itemize}
 \item  The Singular Value Decomposition (SVD). If ${\bf U}\in \mathbb R^{n_1 \times n_2}$ and ${\bf V} \in \mathbb R^{n_2 \times n_2}$ have orthonormal columns, and ${\bf T}\in \mathbb R^{n_2 \times n_2}$ is diagonal (with non-negative entries), the factorization in \eqref{eq_UTV_RelWork} is called the SVD \cite{GolubVanLoan96}. To be more
precise, an SVD of ${\bf A}$ is defined as:
\begin{equation}
\begin{aligned}
{\bf A} = {\bf U}{\bf \Sigma V}^T  = 
\begin{bmatrix} {{\bf U}_k \quad {\bf U}_\perp} \end{bmatrix}
\begin{bmatrix}
{\bf \Sigma}_k & {\bf 0}  \\
{\bf 0} & {\bf \Sigma}_\perp
\end{bmatrix}\begin{bmatrix}
{\bf V}_k^T  \\
{\bf V}_\perp^T
\end{bmatrix},
\end{aligned}
\label{eq_SVD}
\end{equation}
where the columns $\{{\bf u}_i\}_{i=1}^k$ of ${\bf U}_k \in \mathbb R^{n_1 \times k}$ and $\{{\bf u}_i\}_{i=k+1}^{n_2}$ of ${\bf U}_\perp \in \mathbb R^{n_1 \times (n_2-k)}$ span $\mathcal{R}({\bf A})$ and $\mathcal{N}({\bf A}^T)$, respectively. The diagonal matrix ${\bf \Sigma}$ contains the singular values $\sigma_i$'s in a nonincreasing order: ${\bf \Sigma}_k$ comprises the first $k$ and ${\bf \Sigma}_\perp$ the remaining $n_2-k$ singular values. The columns $\{{\bf v}_i\}_{i=1}^k$ of ${\bf V}_k \in \mathbb R^{n_2 \times k}$ and $\{{\bf v}_i\}_{i=k+1}^{n_2}$ of ${\bf V}_\perp \in \mathbb R^{n_2 \times (n_2-k)}$ span $\mathcal{R}({\bf A}^T)$ and $\mathcal{N}({\bf A})$, respectively. For any integer $1 \le r < n_2$, the SVD establishes the best rank-$r$ approximation ${\bf B}$ to ${\bf A}$:
\begin{equation}\notag
\begin{aligned}
& \underset{\text{rank}({\bf B})\le r}{\text{argmin}}
&& \|{\bf A} - {\bf B}\|_\xi = \|{\bf \Sigma}_{r+1}\|_\xi,
\end{aligned}
\end{equation}
where $\xi \in \{2, F\}$, and $\|{\bf \Sigma}_{r+1}\|_2=\sigma_{r+1}$, and $\|{\bf \Sigma}_{r+1}\|_F=\Big(\sum_{i=r+1}^{n_2}{\sigma_i^2}\Big)^{1/2}$.
\item The Column-Pivoted QR (CPQR). If ${\bf U} \in \mathbb R^{n_1 \times n_2}$ is orthonormal, ${\bf T} \in \mathbb R^{n_2 \times n_2}$ is upper triangular, and ${\bf V} \in \mathbb R^{n_2 \times n_2}$ is a permutation matrix, it is called the
CPQR or rank-revealing QR decomposition \cite{GolubVanLoan96, Chan87, Stewart98}. Columns of ${\bf U}$ span $\mathcal{R}({\bf A})$ and $\mathcal{N}({\bf A}^T)$, and diagonals of ${\bf T}$ constitute approximations to the singular values. Let ${\bf T}$ be partitioned as:
\begin{equation}\notag
{\bf T} =
\begin{bmatrix}
{\bf T}_{11} & {\bf T}_{12}  \\
{\bf 0} & {\bf T}_{22}
\end{bmatrix},
\end{equation}
where ${\bf T}_{11} \in \mathbb R^{k \times k}$, ${\bf T}_{12} \in \mathbb R^{k \times (n_2-k)}$, and ${\bf T}_{22} \in \mathbb R^{(n_2-k) \times (n_2-k)}$. The rank-revealing property of CPQR implies that $\sigma_\text{min}({\bf T}_{11})= \mathcal{O}(\sigma_k)$, and $\|{\bf T}_{22}\|_2 = \mathcal{O}(\sigma_{k+1})$.
\item The pivoted QLP (p-QLP). If ${\bf U} \in \mathbb R^{n_1 \times n_2}$ and ${\bf V} \in \mathbb R^{n_2 \times n_2}$ are orthonormal, and ${\bf T} \in \mathbb R^{n_2 \times n_2}$ is lower triangular, it is called the p-QLP \cite{StewartQLP}. The p-QLP is built on the CPQR, and addresses its drawbacks: the CPQR (i) establishes fuzzy approximations to the singular values, and (ii) does not explicitly establish orthogonal bases for $\mathcal{R}({\bf A}^T)$ and $\mathcal{N}({\bf A})$. Let $\bf A$ have a CPQR factorization as follows:
\begin{equation}
 {\bf A\Pi}_A = {\bf Q}_A{\bf R}_A.
 \label{eq_CPQR_1pQLP}
\end{equation}
The p-QLP is obtained by performing a CPQR on ${\bf R}_A^T$:
\begin{equation}
 {\bf R}_A^T\acute{\bf \Pi} = \acute{\bf P}\acute{\bf L}^T.
 \label{eq_CPQR_2pQLP}
\end{equation}
Thus, ${\bf A} = {\bf Q}_A\acute{\bf \Pi}\acute{\bf L}\acute{\bf P}^T{\bf \Pi}_A^T$. Orthogonal ${\bf Q}_A\acute{\bf \Pi}$ and ${\bf \Pi}_A\acute{\bf P}$ provide bases for the column space and row space of $\bf A$, respectively. Diagonals of  $\acute{\bf L}$ approximate the singular values of $\bf A$. The p-QLP is rank-revealing \cite{StewartQLP, MathiasStewart93, HuckabyChan05}.
\item Complete Orthogonal Decompositions (CODs). If ${\bf U} \in \mathbb R^{n_1 \times n_2}$ and ${\bf V} \in \mathbb R^{n_2 \times n_2}$ are orthonormal, and ${\bf T} \in \mathbb R^{n_2 \times n_2}$ is upper or lower triangular, it is called COD \cite{GolubVanLoan96}. CODs
are also called UTV decompositions \cite{ Stewart98, FierroHan97}. They are rank-revealing in the same sense as CPQR.
\end{itemize}

Computing a full factorization of $\bf A$ by these deterministic methods needs $\mathcal{O}(n_1n_2^2)$ arithmetic operations, which is demanding. However, computing a truncated version of them, say a rank $k$ factorization, which is done by terminating the computation procedure after $k$ steps, incurs the cost $\mathcal{O}(n_1n_2k)$. Though this is regarded as efficient, the significant drawback encountered in computing traditional methods is the data movement, as they require to repeatedly access the date \cite{DemGGX15, DuerschGu2017, Dongarraetal18}. To be more specific, the SVD of a matrix is usually computed using the classical bidiagonalization method in two stages: (i) the matrix is reduced to a bidiagonal form, and (ii) the bidiagonal form is reduced to a diagonal form. In the first stage, a large portion of the operations are performed in terms of level-1 and level-2 BLAS (the Basic Linear Algebra Subprograms) routines. While the method only utilizes level-1 BLAS routines in the second stage \cite{Dongarraetal18}, \cite{MartinssonQH2019}. On the other hand, roughly half of the operations of CPQR and most operations of the unpivoted QR factorization are in level-3 BLAS \cite{DuerschGu2017}, \cite{MartinssonQH2019}. Level-1 and level-2 BLAS routines are memory-bound, and as such can not attain high performance on modern computing platforms. Level-3 BLAS routines, however, are CPU-bound; they can take advantage of the data locality in advanced architectures, and thus attain higher performance, very close to the peak performance rendered by the processors.

\subsection{Randomized Methods}
\label{subsec_II_RandMeth}

Recently, matrix approximation methods based on randomized sampling have gained momentum. These methods
combine the ideas from random matrix theory and classical
methods to approximate a matrix, especially matrices with low numerical rank. The advantages offered by randomized methods are (i) computational efficiency, and (ii) parallelization on modern computers. Using randomized methods to factor a matrix may lead to the loss of accuracy. However, the optimality is not necessary in many practical applications.

The earliest works were based on subset selection or
pseudo-skeleton approximation \cite{GoreinovTZ97, FriezeKVS04}. These methods are referred to as CUR decompositions \cite{MahDri09, DeshpandeV2006, RudelsonV07, Friedland11, BoutsidisWood17}, where matrices $\bf C$ and $\bf R$ are formed by the $k$ (actual) columns and rows (selected according to a probability distribution) of an input matrix, respectively, and matrix $\bf U$ is subsequently obtained via various procedures. The works in \cite{ClarWood2017, Sarlos06, PapaRTV00, NelsonNg2013} were based on the concept of \emph{random projections} \cite{AilonCh09} in which the authors argued that projecting a low-rank matrix onto a random subspace produces a good approximation. This is due to the linear dependency of the matrix's rows. The work presented in \cite{Rokhlin09} made use of random Gaussian matrix to firstly compress the input matrix. The rank-$k$ approximation is subsequently obtained by means of an SVD on the small matrix. In \cite{HMT2011}, Halko, Martinsson, and Tropp devised the R-SVD (randomized SVD) algorithm based on random column sampling that approximates the SVD of a given matrix. R-SVD constructs a factorization to matrix $\bf A$ as described in Algorithm \ref{Alg_RSVD}.

\begin{algorithm}
	\caption{Randomized SVD (R-SVD)}
	\begin{algorithmic}[1]
		\renewcommand{\algorithmicrequire}{\textbf{Input:}}
		\Require
		\Statex $n_1 \times n_2$ matrix ${\bf A}$, and integers $1\le k\le d \le n_2$.
		\renewcommand{\algorithmicrequire}{\textbf{Output:}}
		\Require
		\Statex Orthonormal $\widetilde{\bf U} \in \mathbb R^{n_1 \times d}$ that approximates $\mathcal{R}({\bf A})$, diagonal $\widetilde{\bf \Sigma}\in \mathbb R^{d \times d}$ that approximates the $d$ leading singular values of $\bf A$, orthonormal $\widetilde{\bf V} \in \mathbb R^{n_2 \times d}$ that approximates $\mathcal{R}({\bf A}^T)$ and, accordingly, an approximation constructed as $\hat{\bf A} = \widetilde{\bf U}\widetilde{\bf \Sigma}\widetilde{\bf V}^T$.
		\Function{[$\widetilde{\bf U}, \widetilde{\bf \Sigma}, \widetilde{\bf V}$]}{}$=$\texttt{{R\_SVD}}(${\bf A}, d$)
		\State ${\bf \Omega}=\texttt{randn}(n_2, d)$
		\State ${\bf F} = {\bf A}{\bf \Omega}$
		\State $\bar{\bf U} = \texttt{orth}({\bf F})$
		\State ${\bf G} = \bar{\bf U}^T{\bf A}$
		\State $[\Bar{\Bar{\bf U}}, \widetilde{\bf \Sigma}, \widetilde{\bf V}] = \texttt{svd}({\bf G}) \rightarrow \widetilde{\bf U}\coloneqq \bar{\bf U}\Bar{\Bar{\bf U}}$
		\BState \textbf{end function}
		\EndFunction
	\end{algorithmic}
\label{Alg_RSVD}
\end{algorithm}

Gu \cite{Gu2015} modified R-SVD, by using a truncated SVD, and presented a new error analysis. He applied the modified R-SVD to improve the subspace iteration methods. The work in \cite{MFKDeTSP18} proposed a \textit{fixed-rank} matrix approximation method,
that is, a rank-$k$ factorization of an input matrix, termed SOR-SVD (subspace-orbit randomized SVD). Through the random column and row sampling, SOR-SVD first reduces the dimension of the input matrix. Then, the matrix is transformed into a lower dimensional space (compared with matrix ambient dimensions). Lastly, a truncated SVD follows to construct an approximate SVD of the given matrix. The work in \cite{MFKDeJSTSP18} presented a rank-revealing algorithm using the randomized scheme termed compressed randomized UTV (CoR-UTV) decomposition. This algorithm is described in Algorithm \ref{Alg_CoRUTV}.

\begin{algorithm}
	\caption{Compressed Randomized UTV (CoR-UTV)}\begin{algorithmic}[1]
		\renewcommand{\algorithmicrequire}{\textbf{Input:}}
		\Require
		\Statex $n_1 \times n_2$ matrix ${\bf A}$, and integers $1\le k\le d \le n_2$.
		\renewcommand{\algorithmicrequire}{\textbf{Output:}}
		\Require
		\Statex $\widetilde{\bf U} \in \mathbb R^{n_1 \times d}$ that approximates $\mathcal{R}({\bf A})$, upper triangular ${\bf T}\in \mathbb R^{d \times d}$ whose diagonals approximate the $d$ leading singular values of $\bf A$, orthonormal $\widetilde{\bf V} \in \mathbb R^{n_2 \times d}$ that approximates $\mathcal{R}({\bf A}^T)$ and, accordingly, an approximation constructed as $\hat{\bf A} = \widetilde{\bf U}\widetilde{\bf T}\widetilde{\bf V}^T$.
		\Function{[$\widetilde{\bf U}, \widetilde{\bf T}, \widetilde{\bf V}$]}{}$=$\texttt{{CoR\_UTV}}(${\bf A}, d$)
		\State ${\bf \Omega}=\texttt{randn}(n_2, d)$
		\State ${\bf F}_1 = {\bf A}{\bf \Omega}$
		\State ${\bf F}_2 = {\bf A}^T{\bf F}_1$
		\State $\bar{\bf U} = \texttt{orth}({\bf F}_1)$
		\State $\bar{\bf V} = \texttt{orth}({\bf F}_2)$
		\State ${\bf G} = \bar{\bf U}^T{\bf A}\bar{\bf V}$ $\rightarrow$ \big(An approximation to $\bf G$, which does\\ \hspace{.4cm} not need a pass over $\bf A$, can be computed as follows: \\
		\hspace{.4cm} $\hat{\bf G}= \bar{\bf U}^T{\bf F}_1(\bar{\bf V}^T{\bf F}_2)^\dagger)$.\big)
		\State $[\Bar{\Bar{\bf U}}, \widetilde{\bf T}, \Bar{\Bar{\bf \Pi}}] = \texttt{cpqr}({\bf G}) \rightarrow \widetilde{\bf U}\coloneqq \bar{\bf U}\Bar{\Bar{\bf U}}$, and $ \widetilde{\bf V} \coloneqq \bar{\bf V}\Bar{\Bar{\bf \Pi}}$
		\BState \textbf{end function}
		\EndFunction
	\end{algorithmic}
\label{Alg_CoRUTV}
\end{algorithm}

The work in \cite{XiaoGL17} presented the spectrum-revealing QR factorization (SRQR) algorithm for low-rank matrix approximation. The workhorse of the algorithm is a randomized version of CPQR \cite{DuerschGu2017}. The accuracy of SRQR can match that of CPQR. (For a comparison of CPQR, the SVD and randomized methods in terms of approximation accuracy, see, e.g., \cite{DuerschGu2017}, \cite{MFKDeJSTSP18}.) The authors in \cite{FengXG19} presented the Flip-Flop SRQR factorization algorithm. The algorithm first applies SSQR to compute a partial CPQR of the given matrix. Next, an LQ factorization is performed on the R-factor obtained. The final approximation is then given through an SVD of the transformed matrix. The authors made use of Flip-Flop SRQR for robust principal component analysis \cite{MFKJC19} and tensor decomposition applications. The work in \cite{WuX20} presented randomized QLP decomposition algorithm; the authors have replaced the SVD in R-SVD \cite{HMT2011} by p-QLP \cite{StewartQLP} and presented error bounds (in expected value) for the approximate leading singular values of the matrix. 

The current randomized algorithms make use of the SVD or CPQR algorithms in order to factor the transformed (or reduced) input matrix. When dealing with large matrices, the drawback of such deterministic factorization algorithms, as mentioned earlier, is that standard methods to compute them can not efficiently exploit modern parallel architectures, owning to the utilization of level-1 and level-2 BLAS routines \cite{DemGGX15, DuerschGu2017, Dongarraetal18}. In this paper, we develop a randomized matrix factorization algorithm that uses only the unpivoted QR factorization. As most operations of this factorization are in level-3 BLAS, the proposed algorithm thereby weeds out the drawback associated with existing randomized algorithms and in turn lends itself much easily to parallel implementation.

\section{Projection-based Partial QLP (PbP-QLP)}
\label{section_PbPQLP}

We describe in this section our proposed PbP-QLP  algorithm. It approximates the p-QLP \cite{StewartQLP}, providing a factorization in the form of \eqref{eq_PbPQLP}. PbP-QLP is rank-revealing, and its development has been motivated by the p-QLP’s procedure. However, in contrast to the p-QLP, PbP-QLP (i) uses randomization, and (ii) does not use CPQR; it only employs the unpivoted QR as the decompositional apparatus. In addition to the basic form of PbP-QLP described in Section \ref{subsec_SimplePbP}, we present in Section \ref{subsec_PIPbPQLP} a variant of the algorithm that utilizes the power iteration scheme that enhances its accuracy and robustness.

\subsection{Computation of PbP-QLP}
\label{subsec_SimplePbP}
Given matrix ${\bf A}$ and an integer $1\le k\le d\le n_2$, we generate in the first step of computing PbP-QLP an $n_1 \times d$ standard Gaussian matrix ${\bf \Phi}=\texttt{randn}(n_1, d)$. We then transform ${\bf A}$ to a low dimensional space using ${\bf \Phi}$:
\begin{equation}
{\bf C} = {\bf A}^T{\bf \Phi}.
\label{eq_c}
\end{equation}
We next construct an orthonormal basis $\bar{\bf P}\in \mathbb R^{n_2 \times d}$ for $\mathcal{R}({\bf C})$:
\begin{equation}
\bar{\bf P} = \texttt{orth}({\bf C}).
\label{eq_Pbar}
\end{equation}
This step can efficiently be performed by a call to a packaged
QR decomposition. Then, we form a matrix ${\bf D}\in \mathbb R^{n_1 \times d}$ by right-multiplying ${\bf A}$ with $\bar{\bf P}$:
\begin{equation}
{\bf D} = {\bf A}\bar{\bf P}.
\label{eq_D}
\end{equation}
Afterwards, we use the unpivoted QR to factor $\bf D$:
\begin{equation}
{\bf D}= {\bf Q}{\bf R}.
\label{eq_DQR}
\end{equation}
We now carry out another QR factorization on ${\bf R}^T\in \mathbb R^{d \times d}$:
\begin{equation}
{\bf R}^T = \widetilde{\bf P}\widetilde{\bf R}.
\label{eq_RTPtRt}
\end{equation}
Lastly, we construct an approximation $\hat{\bf A}_\text{PbP-QLP}$ to $\bf A$:
\begin{equation}
\hat{\bf A}_\text{PbP-QLP} = {\bf Q}\widetilde{\bf R}^T \widetilde{\bf P}^T \bar{\bf P}^T \coloneqq {\bf QLP}^T,
\label{eq_Ahat_simplePbP}
\end{equation}
where ${\bf L}\coloneqq \widetilde{\bf R}^T$, and ${\bf P} \coloneqq \bar{\bf P}\widetilde{\bf P}$. Orthonormal matrices ${\bf Q}\in \mathbb R^{n_1 \times d}$ and ${\bf P}\in \mathbb R^{n_2 \times d}$ approximate $\mathcal{R}({\bf A})$ and $\mathcal{R}({\bf A}^T)$, respectively. ${\bf L}\in \mathbb R^{d \times d}$ is lower triangular, where its diagonals approximate the first $d$ singular values of $\bf A$, and its $k\times k$ leading block reveals the numerical rank $k$ of $\bf A$.

\noindent \textbf{Justification and Relation to p-QLP.} As expounded in Section \ref{subsec_II_DeterMeth}, the p-QLP for $\bf A$ involves two steps, each performing one CPQR factorization (equations \eqref{eq_CPQR_1pQLP} and \eqref{eq_CPQR_2pQLP}). Stewart \cite{StewartQLP} argues that in the second step of the computation, column pivoting is not necessary, due to pivoting done in the first step. To relate it to PbP-QLP, equations \eqref{eq_c}-\eqref{eq_DQR} emulate (up to $d$ columns) the first step of p-QLP. Since $\bar{\bf P}$ \eqref{eq_Pbar} approximates $\mathcal{R}({\bf A}^T)$ and $\bf D$ is formed as in \eqref{eq_D}, matrix $\bf R$ \eqref{eq_DQR} reveals the numerical rank of $\bf A$. As such, we proceed with an unpivoted QR on ${\bf R}^T$ for the second step in PbP-QLP’s computation. We will show matrix $\bf R$ reveals the gap in $\bf A$’s spectrum through theoretical analysis and simulations. 

\subsection{Power Iteration-coupled PbP-QLP}
\label{subsec_PIPbPQLP} 
PbP-QLP provides highly accurate approximations to the
SVD of matrices with rapidly spectral decay. However, it
may produce less accurate approximations for matrices whose singular values decay slowly. To enhance the accuracy of approximations, we thus utilize the power iteration (PI) scheme \cite{Rokhlin09, HMT2011, Gu2015, VoroninMartin16, YuLi18}. The reasoning behind using the PI scheme is to replace the input matrix with an intimately related matrix whose singular values decay more rapidly. In particular, ${\bf A}^T$ in \eqref{eq_c} is substituted by ${\bf A}_\text{PI}$ defined as:
\begin{equation}\notag
{\bf A}_\text{PI} = ({\bf A}^T{\bf A})^q{\bf A}^T,
\end{equation}
where $q$ is the PI factor. It is seen that ${\bf A}^T$ and ${\bf A}_\text{PI}$ have the same singular vectors, whereas the latter has singular values $\sigma_i^{2q+1}$. It should be noted that the performance improvement brought by the PI technique incurs an extra cost, as PbP-QLP needs more arithmetic operations and also more passes through $\bf A$, if $\bf A$ is stored externally.

The computation of ${\bf A}_\text{PI}$ in floating point arithmetic is prone to round-off errors. To put it precisely, let $\epsilon_\text{machine}$ be the machine precision. Then, any singular component less than $\sigma_1 \epsilon_\text{machine}^{1/(2q+1)}$ will be lost. To compensate this loss of accuracy, an orthonormalization of the sample matrix between each application of ${\bf A}^T$ and ${\bf A}$ is needed \cite{HMT2011,Gu2015}. The resulting PbP-QLP algorithm is presented in Algorithm \ref{Alg_PbPQLP_PI}.
\begin{algorithm}
	\caption{Power Iteration-coupled PbP-QLP}
	\begin{algorithmic}[1]
		\renewcommand{\algorithmicrequire}{\textbf{Input:}}
		\Require
		\Statex $n_1 \times n_2$ matrix ${\bf A}$, and integers $1\le k\le d \le n_2$, and $q\ge 1$.
		\renewcommand{\algorithmicrequire}{\textbf{Output:}}
		\Require  
		\Statex Orthonormal ${\bf Q} \in \mathbb R^{n_1 \times d}$ that approximates $\mathcal{R}({\bf A})$, lower triangular ${\bf L}\in \mathbb R^{d \times d}$ whose diagonals approximate the $d$ leading singular values of $\bf A$, orthonormal ${\bf P} \in \mathbb R^{n_2 \times d}$ that approximates $\mathcal{R}({\bf A}^T)$ and, accordingly, an approximation constructed as $\hat{\bf A} = {\bf Q}{\bf L}{\bf P}^T$.
		\Function{[${\bf Q}, {\bf L}, {\bf P}$]}{}$=$\texttt{{PbP\_QLP}}(${\bf A}, d$)
		\State ${\bf \Phi}=\texttt{randn}(n_1, d)$
		\State ${\bf C} = {\bf A}^T{\bf \Phi}$
		\State $\bar{\bf P} = \texttt{orth}({\bf C})$
		\State \textbf{for} $i=1:q$
		\State \quad ${\bf C} = {\bf A}\bar{\bf P}$; $\bar{\bf P} = \texttt{orth}({\bf C})$
		\State \quad ${\bf C} = {\bf A}^T\bar{\bf P}$; $\bar{\bf P} = \texttt{orth}({\bf C})$
		\State \textbf{end for}
		\State ${\bf D} = {\bf A}\bar{\bf P}$
		\State $[{\bf Q}, {\bf R}] = \texttt{qr}({\bf D})$
		\State $[\widetilde{\bf P}, \widetilde{\bf R}] = \texttt{qr}({\bf R}^T) \rightarrow {\bf P}\coloneqq \bar{\bf P}\widetilde{\bf P}, {\bf L}\coloneqq \widetilde{\bf R}^T$
		\BState \textbf{end function}
		\EndFunction
	\end{algorithmic}
\label{Alg_PbPQLP_PI}
\end{algorithm}

\section{Theoretical Analysis of PbP-QLP}
\label{section_PbPQLPAnalysis}
This section provides a detailed account of theoretical
analysis for the PbP-QLP algorithm.

\subsection{Rank of $\bf A$ is revealed in $\bf R$}
Here, we show that matrix $\bf R$ generated by the PbP-QLP
algorithm (equation \eqref{eq_DQR}, or Step 10 of Algorithm \ref{Alg_PbPQLP_PI}) reveals the numerical rank of $\bf A$. To be precise, let $\bf A$ have an SVD as defined in \eqref{eq_SVD}, and the matrix $\bf R$ be partitioned as:
\begin{equation} 
{\bf R} =
\begin{bmatrix}
{\bf R}_{11} & {\bf R}_{12}  \\
{\bf 0} & {\bf R}_{22}
\end{bmatrix},
\label{eq_Rparti}
\end{equation}
where ${\bf R}_{11} \in \mathbb R^{k \times k}$, ${\bf R}_{12} \in \mathbb R^{k \times (d-k)}$, and ${\bf R}_{22} \in \mathbb R^{(d-k) \times (d-k)}$. We call the diagonals of $\bf R$, R-values. The definition of rank-revealing factorizations in the literature \cite{Chan87, Stewart98, ChandrasekaranIpsen94} implies that
\begin{equation} \label{eq_sigmak_R11}
\sigma_\text{min}({\bf R}_{11})= \mathcal{O}(\sigma_k).
\end{equation}
\begin{equation} \label{eq_2normR22}
\|{\bf R}_{22}\|_2 = \mathcal{O}(\sigma_{k+1}).
\end{equation}

We prove \eqref{eq_2normR22} first, and then \eqref{eq_sigmak_R11}.

\begin{theorem}
	Let ${\bf A} \in \mathbb R^{n_1\times n_2}$ be an input matrix with an SVD defined in \eqref{eq_SVD}, $\bf R$ be generated by PbP-QLP partitioned as in \eqref{eq_Rparti}, integer $p\ge 0$, and $k+p\le d$. Defining
\begin{equation}\notag
	\begin{bmatrix}
	{\bf \Phi}_1 \\
	{\bf \Phi}_2
	\end{bmatrix} \coloneqq {\bf U}^T{\bf \Phi},
	\label{eqG12_ThrLRA}
	\end{equation}
	where ${\bf \Phi}_1 \in \mathbb R^{(d-p)\times d}$ and  ${\bf \Phi}_2 \in \mathbb R^{(n_2-d+p)\times d}$, we then have
\begin{equation}\label{eq_2nR22_bound}
\begin{aligned}
\|{\bf R}_{22}\|_2 & \le \sigma_{k+1} + \frac{\delta_k^{2q+2}\sigma_1\|{\bf \Phi}_2\|_2\|{\bf \Phi}_1^\dagger\|_2}{\sqrt{1 + \delta_k^{4q+4}\|{\bf \Phi}_2\|_2^2\|{\bf \Phi}_1^\dagger\|_2^2}},
\end{aligned}
\end{equation}
where $\delta_k = \frac{\sigma_{d-p+1}}{\sigma_k}$.
\label{Th_R22bound}
\end{theorem}

\textit{Proof.} The proof is given in Appendix \ref{Proo_R22bound}.

\begin{remark}
The relation \eqref{eq_2nR22_bound} implies that one of the conditions through which the rank of $\bf A$ is revealed by matrix $\bf R$ depends on the ratio $\frac{\sigma_{d-p+1}}{\sigma_k}$. Provided that (i) there exists a large gap
in the singular values of $\bf A$ ($\sigma_k \gg \sigma_{k+1}$), and (ii) $\sigma_k$ is not substantially smaller than $\sigma_1$, for any $p \ge 0$ the trailing block of $\bf R$ is sufficiently
small in magnitude. In addition, utilizing the PI technique
exponentially derives down the quotient in \eqref{eq_2nR22_bound} to zero. It is therefore expected that PI considerably improves the accuracy of PbP-QLP in forming a rank-revealing factorization.
\end{remark}

The following theorem bounds the minimum singular value of the leading block of $\bf R$.

\begin{theorem}\label{Th_boundssigmak}
 Under the notation and hypotheses of Theorem \ref{Th_R22bound}, we have
 \begin{equation}\label{eq_Th_boundssigmak}
  \sigma_{k+1} \le \sigma_k({\bf R}_{11}) \le \sigma_k + \sigma_{k+1}
 \end{equation}
\end{theorem}
\textit{Proof.} The proof is given in Appendix \ref{Proo_boundssigmak}.

\begin{remark}
The bounds presented in this theorem together with that of Theorem 1 assert that if there is a substantial gap in the singular values of $\bf A$, and $\sigma_1$ is not substantially larger than $\sigma_k$, the rank of $\bf A$ is revealed in $\bf R$.
\end{remark}

\subsection{Approximate Subspaces and Low-Rank Approximation}
The closeness of any two subspaces, say ${\bf S}_1$ and ${\bf S}_2$, is measured by the \textit{distance} between them \cite{GolubVanLoan96}. Denoted by $\text{dist}({\bf S}_1, {\bf S}_2)$, it corresponds to the largest canonical angle between the two subspaces. The following theorem establishes the distances between ${\bf U}_k$ and ${\bf V}_k$  (the principal left and right singular vectors of $\bf A$, respectively) and their corresponding approximations ${\bf Q}_1$ and ${\bf P}_1$ by PbP-QLP, formed by the first $k$ columns of ${\bf Q}$ and ${\bf P}$.

\begin{theorem}	\label{Th_distUkVkQ1P1}
Let ${\bf A} \in \mathbb R^{n_1\times n_2}$ be an input matrix with an SVD defined in \eqref{eq_SVD}, and ${\bf Q}$ and ${\bf P}$ be generated by PbP-QLP. Let further
	\begin{equation}\notag
	\begin{aligned}
		\begin{bmatrix}
	{\bf \Phi}_{11} & {\bf \Phi}_{12} \\
	{\bf \Phi}_{21} & {\bf \Phi}_{22}
	\end{bmatrix} \coloneqq {\bf U}^T{\bf \Phi},
	\end{aligned}
	\end{equation}
where ${\bf \Phi}_{11} \in \mathbb R^{k \times k}$, ${\bf \Phi}_{21} \in \mathbb R^{(n_2-k) \times k}$, ${\bf \Phi}_{12} \in \mathbb R^{k \times (d-k)}$ and ${\bf \Phi}_{22} \in \mathbb R^{(n_2-k) \times (d-k)}$. Let ${\bf \Phi}_{11}$ be full rank, and ${\bf Q}_1$ and ${\bf P}_1$ be formed by the first $k$ columns of ${\bf Q}$ and ${\bf P}$. Then
	\begin{equation}\label{eq_Th_distance1}
    \text{dist}(\mathcal{R}({\bf U}_k), \mathcal{R}({\bf Q}_1)) \le \Big(\frac{\sigma_{k+1}}{\sigma_k}\Big)^{2q+2}\|{\bf \Phi}_{21}{\bf \Phi}_{11}^{-1}\|_2.
	\end{equation}
	\begin{equation}\label{eq_Th_distance2}
    \text{dist}(\mathcal{R}({\bf V}_k), \mathcal{R}({\bf P}_1)) \le \Big(\frac{\sigma_{k+1}}{\sigma_k}\Big)^{2q+1}\|{\bf \Phi}_{21}{\bf \Phi}_{11}^{-1}\|_2.
\end{equation}
	\end{theorem}
\textit{Proof.} The proof is provided in Appendix \ref{Proo_distance}.
 
\begin{remark} \label{Remark_Basis}
Provided ${\sigma_k}>{\sigma_{k+1}}$, Theorem \ref{Th_distUkVkQ1P1} shows that the subspaces $\mathcal{R}({\bf Q}_1)$ and $\mathcal{R}({\bf P}_1)$ converge respectively to $\mathcal{R}({\bf U}_k)$ and $\mathcal{R}({\bf V}_k)$ at a rate proportional to $\Big(\frac{\sigma_{k+1}}{\sigma_k}\Big)^q$. The convergence rate also depends on the distribution of the random matrix employed. The virtue of powering the input matrix laid out by Theorem \ref{Th_distUkVkQ1P1} can be seen in the accuracy of PI-coupled PbP-QLP: for large enough $q$, ${\bf Q}_1^T{\bf U}_\perp \approx 0$, which vanishes the first term in \eqref{eq_2n2ndTermR22}, as well as the second term in the right-hand side of the last relation in \eqref{eq_R11_exp}. This, simply, implies that with an appropriately chosen $q$, the rank of the input matrix is revealed in submatrix $\bf R$ (Step 6 of Alg. \ref{Alg_PbPQLP_PI}), regardless of the gap in the spectrum being substantial or not.
\end{remark}

\begin{theorem}\label{Th_LRapprox}
 Under the notation of Theorem \ref{Th_R22bound}, let further $\bf Q$ be constructed by PbP-QLP. Then
\begin{equation}\label{eq_2nLRAErrbound}
\begin{aligned}
\|({\bf I} -  {\bf QQ}^T){\bf A}\|_2 & \le \|{\bf \Sigma}_\perp\|_2 + \frac{\delta_k^{2q+2}\sigma_1\|{\bf \Phi}_2\|_2\|{\bf \Phi}_1^\dagger\|_2}{\sqrt{1 + \delta_k^{4q+4}\|{\bf \Phi}_2\|_2^2\|{\bf \Phi}_1^\dagger\|_2^2}}.
\end{aligned}
\end{equation}
\end{theorem}

\textit{Proof.} The proof is provided in Appendix \ref{Proo_LRapprox}.

\begin{remark}
This theorem shows the PI scheme considerably enhances
the approximation quality of PbP-QLP: the accuracy of
a low-rank approximation generated by PbP-QLP coupled with PI depends on the ratio $\frac{\sigma_{d-p+1}}{\sigma_k}$. Provided that $\sigma_k > \sigma_{k+1}$, for any $p \ge 0$, the extra quotient in the bound \eqref{eq_2nLRAErrbound} is driven to zero exponentially fast.
\end{remark}

\subsection{Estimated Singular Values and Rank-Revealing Property of PbP-QLP}
Here, we establish the relation between the first $k$ estimated singular values of $\bf A$ computed by PbP-QLP, i.e., diagonals of $\bf L$, and the first $k$ singular values given by the SVD.

\begin{theorem}
Under the notation of Theorem \ref{Th_R22bound}, let further $\bf D$ be generated by PbP-QLP. Then, for $i = 1,..., k$, we have
\begin{equation}\label{eq_Th_SinValD_bound}
	\sigma_i \ge \sigma_i({\bf D}) \ge \frac{\sigma_i}{\sqrt{1 + \delta_i^{4q+4}\|{\bf \Phi}_2\|_2^2\|{\bf \Phi}_1^\dagger\|_2^2}},
	\end{equation}
	where $\delta_i= \frac{\sigma_{d-p+1}}{\sigma_i}$.
	\label{Theorem_SinValueApprox}
\end{theorem}

\textit{Proof.} The proof is given in Appendix \ref{Proo_SinValueApprox1}.

\begin{theorem} \label{Th_SinValueApprox2}
 Under the notation of Theorem \ref{Th_R22bound}, let further $\bf D$ and ${\bf L}$ be generated by PbP-QLP, and ${\bf L}$ be partitioned as:
\begin{equation} \label{eq_Lparti}
\begin{aligned}
{\bf L} =\begin{bmatrix}
{\bf L}_{11} & {\bf 0} \\
{\bf L}_{21}  & {\bf L}_{22}
\end{bmatrix},
\end{aligned}
\end{equation}
where ${\bf L}_{11} \in \mathbb R^{k \times k}$, ${\bf L}_{21} \in \mathbb R^{(d-k) \times k}$, and ${\bf L}_{22} \in \mathbb R^{(d-k) \times (d-k)}$. Then, for $i=1,..., k$, we have
\begin{equation}\label{eq_sigmaL11sigmai}
\begin{aligned}
\frac{{\sigma_i({\bf L}_{11})}}{{\sigma_i}} \ge
\frac{\Bigg[1 - \mathcal{O}\Bigg(\frac{\|{\bf L}_{12}\|_2^2}{(1-\rho^2)\sigma_k^2({\bf L}_{11})}\Bigg)\Bigg]}{\sqrt{1 + \delta_i^{4q+4}\|{\bf \Phi}_2\|_2^2\|{\bf \Phi}_1^\dagger\|_2^2}}.
\end{aligned}
\end{equation}
where $\rho = \frac{\|{\bf L}_{22}\|_2}{\sigma_k({\bf L}_{11})}$.
\end{theorem}
 
\textit{Proof.}  The proof is given in Appendix \ref{Proo_SinValueApprox2}.

We have thus far shown matrix $\bf R$ generated by PbP-QLP reveals the numerical rank of $\bf A$. The rank-revealing property of PbP-QLP is guaranteed by Theorem 2.1 of \cite{MathiasStewart93}: let $\bf R$ and $\bf L$ be constructed by PbP-QLP and partitioned as in \eqref{eq_Rparti} and \eqref{eq_Lparti}, respectively. Then, we have
\begin{equation}\notag
\sigma_\text{min}({\bf L}_{11}) \ge \sigma_\text{min}({\bf R}_{11}),
\end{equation}
\begin{equation}\notag
\sigma_1({\bf L}_{22}) \le \sigma_1({\bf R}_{22}).
\end{equation}

\begin{theorem}\label{Th_boundsA'ssigmak}
Under the notation of Theorem \ref{Th_distUkVkQ1P1}, let further $\bf L$ be constructed by PbP-QLP and partitioned as in \eqref{eq_Lparti}. Then
\begin{equation}\notag
\sigma_{\text{min}}({\bf L}_{11}) \le \sigma_k \le \sigma_{\text{min}}({\bf L}_{11}) + \sigma_1\Big(\frac{\sigma_{k+1}}{\sigma_k}\Big)^{2q+2}\|{\bf \Phi}_{21}{\bf \Phi}_{11}^{-1}\|_2.
\end{equation}
\end{theorem}

\textit{Proof.} The proof is provided in Appendix \ref{Proo_boundsA'ssigmak}.

\subsection{High-Probability Error Bounds}
The following theorem provides high-probability bounds
which give further insights into the accuracy of PbP-QLP.

\begin{theorem}\label{Th_SinVal_PbP}
 Let ${\bf A} \in \mathbb R^{n_1\times n_2}$ be an input matrix with an SVD defined in \eqref{eq_SVD}, $0\le p\le d-k$, and $\bf R$ and $\hat{\bf A}_\text{PbP-QLP}$ be generated by PbP-QLP. Let further $0<\Upsilon \ll 1$, and define
 \begin{equation}\label{key}
C_\Upsilon = \frac{e\sqrt{d}}{p+1}\Big(\frac{2}{\Upsilon}\Big)^\frac{1}{p+1}\Big(\sqrt{n-d+p}+\sqrt{d} + \sqrt{2\text{log}\frac{2}{\Upsilon}}\Big).
\end{equation}
Then with probability not less than $1 - \Upsilon$, we have
\begin{equation}\label{eq_HighProB_R22}
\sigma_1({\bf R}_{22}) \le \sigma_k +\delta_k^{2q+2}\sigma_1C_\Upsilon,
\end{equation}
\begin{equation}\label{eq_HighProB_SVs}
\begin{aligned}
\frac{{\sigma_i(\hat{\bf A}_\text{PbP-QLP})}}{{\sigma_i}} \ge
\frac{\Bigg[1 - \mathcal{O}\Bigg(\frac{\|{\bf L}_{12}\|_2^2}{(1-\rho^2)\sigma_k^2({\bf L}_{11})}\Bigg)\Bigg]}{\sqrt{1 + \delta_i^{4q+4}C_\Upsilon^2}},
\end{aligned}
\end{equation}
and
\begin{equation}\label{eq_HighProB_LRAE}
\begin{aligned}
\|({\bf I} - {\bf QQ}^T){\bf A}\|_2 & \le \|{\bf \Sigma}_\perp\|_2 + \delta_k^{2q+2}\sigma_1C_\Upsilon.
\end{aligned}
\end{equation}
\end{theorem}

\textit{Proof.} The proof is given in Appendix \ref{Proo_SinVal_PbP}.

For the distance between approximate subspaces $\mathcal{R}({\bf Q}_1)$ and $\mathcal{R}({\bf P}_1)$ and the corresponding singular subspaces, we have with high probability 
\begin{equation}\notag
    \text{dist}(\mathcal{R}({\bf U}_k), \mathcal{R}({\bf Q}_1)) \le c\sqrt{k(n_1-k)}\Big(\frac{\sigma_{k+1}}{\sigma_k}\Big)^{2q+2}.
	\end{equation}
	\begin{equation}\notag
\text{dist}(\mathcal{R}({\bf V}_k), \mathcal{R}({\bf P}_1)) \le c\sqrt{k(n_1-k)}\Big(\frac{\sigma_{k+1}}{\sigma_k}\Big)^{2q+1}.
	\end{equation}
Here $c = c_1c_2$, where $c_1$ and $c_2$ are positive absolute constants. These results follow by applying the bounds for standard Gaussian matrices ${\bf \Phi}_{11}$ and ${\bf \Phi}_{21}$ devised in \cite{Edelman88, Szarek91}: 
\begin{equation}\notag 
\|{\bf \Phi}_{11}^{-1}\|_2 \le c_1\sqrt{k}, \text{and} \quad \|{\bf \Phi}_{21}\|_2 \le  c_2\sqrt{n_1 - k}.                                                                                                                                                                                                   \end{equation}

\subsection{Computational Cost}
\label{subsec_computcost}
Computation of the simple form of PbP-QLP for matrix $\bf A$ needs the following arithmetic operations:
\begin{itemize}
    \item Generating a matrix $\bf \Phi$ with Gaussian  random variables costs $\mathcal{O}(n_1d)$.
	\item Forming matrix $\bf C$ in \eqref{eq_c} costs $\mathcal{O}(n_1n_2d)$.
	\item Constructing the basis $\bar{\bf P}$ in \eqref{eq_Pbar} costs $\mathcal{O}(n_2d^2)$.
	\item Forming matrix $\bf D$ in \eqref{eq_D} costs $\mathcal{O}(n_1n_2d)$.
	\item  Constructing matrices ${\bf Q}$ and ${\bf R}$ in \eqref{eq_DQR} costs $\mathcal{O}(n_1d^2)$.
	\item Constructing matrices $\widetilde{\bf P}$ and $\widetilde{\bf R}$ in \eqref{eq_RTPtRt} costs $\mathcal{O}(d^3)$.
	\item Forming matrix $\bf P$ costs $\mathcal{O}(n_2d^2)$.
\end{itemize}

The above cost is dominated by multiplications of $\bf A$ and ${\bf A}^T$ by associated matrices. Thus,
\begin{equation}\notag
\mathcal{C}_\text{PbP-QLP} = \mathcal{O}(n_1n_2d).
\label{eq_CostBasicRPTSOD}
\end{equation}

The computation of PbP-QLP is more efficient provided that the matrix $\bf A$ is sparse. In this case, the flop count is proportional to the number of non-zero entries $z$ of $\bf A$ and satisfies $\mathcal{O}(zd)$. Considering $\bf A$ to be \emph{large}, that is, $\bf A$ is stored out-of-memory, the simple form of PbP-QLP makes two passes over $\bf A$ to construct an approximation. However, if PbP-QLP is coupled with the PI scheme, it needs $2q+2$ passes over $\bf A$ and, as a result, its flop count satisfies $(q+1)\mathcal{C}_\text{PbP-QLP}$.

In addition to the arithmetic cost, another cost imposed on any algorithm is the communication cost \cite{DemGGX15, DuerschGu2017}. This cost, which constitutes  the primary factor for traditional methods to be unsuitable on high performance computing devices, is associated with the exploitation of level-1, 2 and 3 BLAS routines: it is determined by moving data between processors working in parallel, and also data movement between different levels of the memory hierarchy. On modern devices, the communication cost considerably dominates the cost of an algorithm. This has motivated researchers to incorporate randomized sampling paradigm into matrix factorization algorithms \cite{DemGGX15, DuerschGu2017, MartinssonQH2019}, where the share of level-3 BLAS operations is higher. In addition, among classical methods applied to the reduced matrices by randomized methods, the unpivoted QR factorization algorithm needs the least data access (the least communication cost), as the large majority of its operations are in terms of level-3 BLAS. As PbP-QLP utilizes only the QR factorization to construct the approximation, its operations can be cast almost completely in terms of level-3 BLAS operations. Consequently, it can be executed more efficiently on highly parallel machines. 
 
\section{Nunerical Simulations}
\label{sec_NumerExp}
This section reports our numerical results obtained by conducting a set of experiments with randomly generated data as well as real-world data. They show the performance behavior of the PbP-QLP algorithm in comparison with several existing algorithms. The experiments are run in MATLAB on a 4-core Intel Core i7 CPU running at 1.8 GHz and 8GB RAM.

\subsection{Runtime Comparison}
We compare the speed of proposed PbP-QLP against two most representative randomized algorithms, namely R-SVD \cite{HMT2011} (Algorithm \ref{Alg_RSVD}) and CoR-UTV \cite{MFKDeJSTSP18} (Algorithm \ref{Alg_CoRUTV}), in factoring input matrices with various dimensions. We generate square, dense matrices of order $n$, and consider three cases for the sampling size parameter $d$, specifically $d=0.04n$, $d=0.2n$ and $d=0.3n$. Note that for the runtime comparison, the distribution of singular values of matrices is immaterial. In this experiment, we have excluded deterministic algorithms for two reasons: firstly, they are considerably slower than their randomized counterparts due to computational cost. Secondly, there are no optimized MATLAB implementations for a truncated version of these algorithms. This, accordingly, enables us to clearly display the behavior of considered algorithms. The results for the three algorithms with and without the power iteration technique are reported in Tables \ref{Table1RuntimePbPQLP}-\ref{Table3RuntimePbPQLP}. For each runtime measurement, the result is averaged over 10 independent runs. It is seen that for the first scenario, where $d=0.04n$ (Table \ref{Table1RuntimePbPQLP}), PbP-QLP shows similar performance as R-SVD and CoR-UTV in most cases. However, as the dimension of the input matrix grows, PbP-QLP begins to edge out. For the second and third scenarios (Tables \ref{Table2RuntimePbPQLP} and \ref{Table3RuntimePbPQLP}), we observe that PbP-QLP outperforms the other two algorithms in all cases. Moreover, for larger matrices we observe larger discrepancies in execution time. This is because, as mentioned earlier, PbP-QLP only makes use of the unpivoted QR factorization (to factor the reduced-size matrix) whose vast majority of operations (contrary to the SVD and CPQR) are in level-3 BLAS operations. These results simply show that the ``quality" of operations matters, and factorization algorithms that incorporate more BLAS-3 routines consume less time. As such, if implemented on a highly parallel machine, PbP-QLP would show yet better results compared to R-SVD and CoR-UTV. (Compared to R-SVD and PbP-QLP, CoR-UTV needs more arithemtic operations, but almost all of them are in level-3 BLAS. Thus, we expect it shows better performance if implemented in parallel.)

\begin{table}[!htb]
\caption{Computational time (in seconds) for different randomized algorithms. $d$ is the sampling size parameter, and $q$ is the power iteration factor.}
\begin{tabular}{p{1.2cm} p{0.5cm} p{2.5cm} p{2.5cm}} 
\noindent\rule{8.1cm}{0.4pt}\\
Algorithms &  &
 \begin{tabular}{p{0.65cm} p{0.65cm} p{0.65cm} p{0.65cm} p{0.65cm}}
 \multicolumn{5}{c}{$n$ $(d=0.04n)$} \\
\hline
 5000 & 10000 & 15000 & 20000 & 25000 \\
 \end{tabular} \\
\noindent\rule{8.1cm}{0.4pt}\\
R-SVD & 
\begin{tabular}{|p{0.5cm} p{.8cm} p{0.65cm} p{0.6cm} p{0.65cm} p{0.5cm}}
$q$=0 & 0.29 & 2.3 & 7.2  & 16.9 & 35.1  \\
$q$=1 & 0.51 & 4.1 & 13.0 & 29.7 & 59.5 \\
$q$=2 & 0.74 & 5.9 & 18.6 & 42.8 & 84.1 \\
		\end{tabular} \\ \\
CoR-UTV & 
\begin{tabular}{|p{0.5cm} p{.8cm} p{0.65cm} p{0.6cm} p{0.65cm} p{0.5cm}}
$q$=0 & 0.32 & 2.3 & 7.5  & 17.0 & 34.4 \\
$q$=1 & 0.58 & 4.2 & 13.2 & 29.6 & 59.3 \\
$q$=2 & 0.80 & 6.0 & 18.8 & 42.7 & 83.7  \\
		\end{tabular} \\\\ 
PbP-QLP & 
\begin{tabular}{|p{0.5cm} p{.8cm} p{0.65cm} p{0.6cm} p{0.65cm} p{0.5cm}}
$q$=0 & 0.29 & 2.2 & 7.1  & 16.0 & 31.6 \\
$q$=1 & 0.53 & 4.1 & 12.7 & 28.7 & 56.6 \\
$q$=2 & 0.81 & 5.9 & 18.6 & 41.4 & 81.7 \\
		\end{tabular} \\
\noindent\rule{8.1cm}{0.4pt}\\
\end{tabular}
\label{Table1RuntimePbPQLP}
\end{table}

\begin{table}[!htb]
\caption{Computational time (in seconds) for different randomized algorithms. $d$ is the sampling size parameter, and $q$ is the power iteration factor.}
\begin{tabular}{p{1.2cm} p{0.5cm} p{2.5cm} p{2.5cm}} 
\noindent\rule{8.1cm}{0.4pt}\\
Algorithms &  &
 \begin{tabular}{p{0.65cm} p{0.65cm} p{0.65cm} p{0.65cm} p{0.65cm}}
 \multicolumn{5}{c}{$n$ $(d=0.2n)$} \\
\hline
 5000 & 10000 & 15000 & 20000 & 25000 \\
 \end{tabular} \\
\noindent\rule{8.1cm}{0.4pt}\\
R-SVD & 
\begin{tabular}{|p{0.5cm} p{.75cm} p{0.65cm} p{0.6cm} p{0.65cm} p{0.5cm}}
$q$=0 & 2.4 & 20.2 & 65.7 & 195 & 429  \\
$q$=1 & 3.6 & 29.1 & 94.8 & 280 & 613 \\
$q$=2 & 4.7 & 37.9 & 124  & 364 & 782 \\
		\end{tabular} \\ \\
CoR-UTV & 
\begin{tabular}{|p{0.5cm} p{.75cm} p{0.65cm} p{0.6cm} p{0.65cm} p{0.5cm}}
$q$=0 & 2.7 & 22.7 & 74.4 & 236 & 469 \\
$q$=1 & 3.8 & 31.7 & 103  & 316 & 767 \\
$q$=2 & 4.9 & 39.7 & 133  & 408 & 888 \\
		\end{tabular} \\\\
PbP-QLP & 
\begin{tabular}{|p{0.5cm} p{.75cm} p{0.65cm} p{0.6cm} p{0.65cm} p{0.5cm}}
$q$=0 & 2.1 & 15.4 & 48.7 & 160 & 319  \\
$q$=1 & 3.3 & 24.1 & 77.5 & 248 & 496  \\
$q$=2 & 4.3 & 33.4 & 107  & 332 & 654  \\
		\end{tabular} \\
\noindent\rule{8.1cm}{0.4pt}\\
\end{tabular}
\label{Table2RuntimePbPQLP}
\end{table}

\begin{table}[!htb]
\caption{Computational time (in seconds) for different randomized algorithms. $d$ is the sampling size parameter, and $q$ is the power iteration factor.}
\begin{tabular}{p{1.2cm} p{0.5cm} p{2.5cm} p{2.5cm}} 
\noindent\rule{8.1cm}{0.4pt}\\
Algorithms &  &
 \begin{tabular}{p{0.65cm} p{0.65cm} p{0.65cm} p{0.65cm} p{0.65cm}}
 \multicolumn{5}{c}{$n$ $(d=0.3n)$} \\
\hline
 5000 & 10000 & 15000 & 20000 & 25000 \\
 \end{tabular} \\
\noindent\rule{8.1cm}{0.4pt}\\
R-SVD & 
\begin{tabular}{|p{0.5cm} p{.75cm} p{0.65cm} p{0.6cm} p{0.65cm} p{0.5cm}}
$q$=0 & 4.7 & 40.5 & 142 & 387 & 856  \\
$q$=1 & 6.3 & 53.2 & 193 & 504 & 1148 \\
$q$=2 & 8.0 & 66.6 & 244 & 622 & 1423 \\
		\end{tabular} \\ \\
CoR-UTV & 
\begin{tabular}{|p{0.5cm} p{.75cm} p{0.65cm} p{0.6cm} p{0.65cm} p{0.5cm}}
$q$=0 & 5.9 & 48.8 & 179 & 481 & 993 \\
$q$=1 & 7.7 & 61.7 & 230 & 618 & 1325 \\
$q$=2 & 9.5 & 74.3 & 280 & 732 & 1626 \\
		\end{tabular} \\\\
PbP-QLP & 
\begin{tabular}{|p{0.5cm} p{.75cm} p{0.65cm} p{0.6cm} p{0.65cm} p{0.5cm}}
$q$=0 & 3.5 & 25.9 & 95.0 & 285 & 601  \\
$q$=1 & 5.2 & 38.7 & 146  & 402 & 896  \\
$q$=2 & 7.0 & 51.5 & 197  & 530 & 1201  \\
		\end{tabular} \\
\noindent\rule{8.1cm}{0.4pt}\\ 
\end{tabular}
\label{Table3RuntimePbPQLP}
\end{table}

\begin{figure}[t]
	\begin{center}
			% This file was created by matlab2tikz v0.4.7 running on MATLAB 8.3.
% Copyright (c) 2008--2014, Nico Schlömer <nico.schloemer@gmail.com>
% All rights reserved.
% Minimal pgfplots version: 1.3
% 
% The latest updates can be retrieved from
%   http://www.mathworks.com/matlabcentral/fileexchange/22022-matlab2tikz
% where you can also make suggestions and rate matlab2tikz.
% 
%
% defining custom colors
\usetikzlibrary{positioning,calc}

\definecolor{mycolor1}{rgb}{0.00000,1.00000,1.00000}%
\definecolor{mycolor2}{rgb}{1.00000,0.00000,1.00000}%

\pgfplotsset{every axis label/.append style={font=\footnotesize},
every tick label/.append style={font=\footnotesize}
}

\begin{tikzpicture}[font=\footnotesize] 

\begin{axis}[%
name=ber,
ymode=log,
width  = 0.35\columnwidth,%5.63489583333333in,
height = 0.4\columnwidth,%4.16838541666667in,
scale only axis,
xmin  = 1,
xmax  = 29,
xlabel= {$d$},
xmajorgrids,
ymin = 0.0,
ymax = 1.1,
xtick       ={10,20,25},
xticklabels ={$10$,$20$, $25$},
ylabel={Magnitude},
ymajorgrids,
legend entries = {SVD,CPQR,p-QLP, R-SVD,CoR-UTV,R-values,PbP-QLP},
legend style={at={(0.56,0.66)},anchor=north east,draw=black,fill=white,legend cell align=left,font=\tiny},
]

%% SVD 
\addplot+[smooth,color=black,loosely dotted, every mark/.append style={solid}, mark=x]
table[row sep=crcr]{
1	1.00012118475465\\
3	0.998158766939078\\
5	0.996025885647185\\
7	0.993989379158925\\
9	0.992053275575964\\
11	0.989864975808377\\
13	0.988157442347616\\
15	0.986019198329047\\
17	0.984206589840957\\
19	0.981895396901780\\
20  0.980956298901488\\
21	0.00483157408311632\\
23	0.00476515872308047\\
25	0.00473124221714669\\
27	0.00469411573292995\\
29	0.00465798947102966 \\
};
%% QRP
\addplot+[smooth,color=gray,loosely dotted, every mark/.append style={solid}, mark=+]
table[row sep=crcr]{
1	0.203650567182692\\
3	0.196463657035048\\
5	0.187558718347019\\
7	0.176706357549629\\
9	0.163869890028070\\
11	0.158334051362534\\
13	0.153848232961766\\
15	0.137179663617259\\
17	0.125376825644912\\
19	0.114288582269537\\
20  0.0963521687770062\\
21	0.00534846053641404\\
23	0.00461425384945296\\
25	0.00409147638985509\\
27	0.00380758553489865\\
29	0.00369461500778390 \\
};

%% p-QLP
\addplot+[smooth,color=red,loosely dotted, every mark/.append style={solid}, mark=pentagon]
table[row sep=crcr]{
1	0.994656828982110\\
3	0.992359824589945\\
5	0.991096828618124\\
7	0.990992277587761\\
9	0.990667786037168\\
11	0.990114633012273\\
13	0.989533791833799\\
15	0.989388086725326\\
17	0.988963477780713\\
19	0.987222430536951\\
20  0.986354150691751\\
21	0.00356965394967211\\
23	0.00350857432696402\\
25	0.00348173284896542\\
27	0.00346124185321653\\
29	0.00344810059730215 \\
  };

%%% R-SVD
\addplot+[smooth,color=teal,loosely dotted, every mark/.append style={solid}, mark=diamond]
table[row sep=crcr]{
1	0.999977100150511\\
3	0.998025349909413\\
5	0.995893568062639\\
7	0.993856689253287\\
9	0.991920737610294\\
11	0.989717765043172\\
13	0.987995708054390\\
15	0.985874609927315\\
17	0.984034631038748\\
19	0.981737509989671\\
20  0.980815683430395\\
21	0.00371703783032968\\
23	0.00363065309573130\\
25	0.00356032378123204\\
27	0.00349966051939147\\
29	0.00344214400332072 \\
};

%%% CoR-UTV
\addplot+[smooth,color=blue,loosely dotted, every mark/.append style={solid}, mark=star]
table[row sep=crcr]{
1	0.993557144664490\\
3	0.992413437624261\\
5	0.991910679195596\\
7	0.991316337326227\\
9	0.990768378218776\\
11	0.990275385951799\\
13	0.989762756441939\\
15	0.989164056503674\\
17	0.988568306457743\\
19	0.987674978894729\\
20  0.986960678215478\\
21	0.00351730691535859\\
23	0.00347692792889298\\
25	0.00344778951157409\\
27	0.00343035404304804\\
29	0.00341676224000812 \\
};

%%% R-values
\addplot+[smooth,color=orange,loosely dotted, every mark/.append style={solid}, mark=-]
table[row sep=crcr]{
1	0.989952334384600\\
3	0.989875411997172\\
5	0.990817771774048\\
7	0.990412338520818\\
9	0.989717936490270\\
11	0.990621902187237\\
13	0.989195151794162\\
15	0.988972926558660\\
17	0.981672589407076\\
19	0.950797506799222\\
20  0.733907220961611\\
21	0.00882913893835920\\
23	0.00348138705465785\\
25	0.00342639175730215\\
27	0.00341024520569232\\
29	0.00340446555676689 \\
};

%%% PbP-QLP
\addplot+[smooth,color=green,loosely dotted, every mark/.append style={solid}, mark=triangle]
table[row sep=crcr]{
1	0.990111800572219\\
3	0.990017877010518\\
5	0.991012576994644\\
7	0.990621478704719\\
9	0.990146101734716\\
11	0.991202077967465\\
13	0.990434522437949\\
15	0.990881005503767\\
17	0.989873473514404\\
19	0.990395064604219\\
20  0.990327903222744\\
21	0.00342421487240781\\
23	0.00343594874941445\\
25	0.00342003683919225\\
27	0.00340897685969797\\
29	0.00340289715162827 \\
};

\end{axis}

\begin{axis}[%
name=SumRate,
at={($(ber.east)+(35,0em)$)},
		anchor= west,
ymode=log,
width  = 0.35\columnwidth,%5.63489583333333in,
height = 0.4\columnwidth,%4.16838541666667in,
scale only axis,
xmin  = 1,
xmax  = 29,
xlabel= {$d$},
xmajorgrids,
xtick       ={10,20,25},
xticklabels ={$10$,$20$, $25$},
ymin = 0.003,
ymax = 1.1,
ylabel={},
ymajorgrids,
%title = {\texttt{HighNoiseLowRank}}]
%ytick       ={0.0973831, 0.0973830 , 0.0973829, 0.0973828},
%yticklabels ={$9.73831$, $9.73830$ , $9.73829$, $9.73828$},
%legend entries={SVD,CPQR,p-QLP, R-SVD,CoR-UTV,R-values,PbP-QLP}, 
%legend style={at={(1,0.82)},anchor=north east,draw=black,fill=white,legend cell align=left,font=\tiny, legend columns=2}
]
%% SVD 
\addplot+[smooth,color=black,loosely dotted, every mark/.append style={solid}, mark=x]
table[row sep=crcr]{
1	1.00011894510508 \\
3	0.998016775192999 \\
5	0.996019828454719 \\
7	0.994226164611526 \\
9	0.992149474486267 \\
11	0.989908574539390 \\
13	0.988012885516550 \\
15	0.986107606763717 \\
17	0.984179489185476 \\
19	0.982187827457008 \\
20  0.980995010048426\\
21	0.00482236284807617 \\
23	0.00478660623923592 \\
25	0.00473536868547800 \\
27	0.00468807940398825 \\
29	0.00467023272482854 \\
};
%% QRP
\addplot+[smooth,color=gray,loosely dotted, every mark/.append style={solid}, mark=+]
table[row sep=crcr]{
1	0.229687175471309 \\
3	0.212585158555365 \\
5	0.191380099883652 \\
7	0.179116025547853 \\
9	0.169224765673710 \\
11	0.158664552703920 \\
13	0.144330324988094 \\
15	0.136314783675545 \\
17	0.122439501330723 \\
19	0.106909826500025 \\
20  0.0983226619517222\\
21	0.00585985582057391 \\
23	0.00466758237180545 \\
25	0.00402776325128298 \\
27	0.00389899975905609 \\
29	0.00368672886835435 \\
};

%% p-QLP
\addplot+[smooth,color=red,loosely dotted, every mark/.append style={solid}, mark=pentagon]
table[row sep=crcr]{
1	0.993720380934942 \\
3	0.992200729554874 \\
5	0.991847504217045 \\
7	0.991586632940638 \\
9	0.990858171209560 \\
11	0.990289103069491 \\
13	0.989439457083098 \\
15	0.988870221975360 \\
17	0.988498054292553 \\
19	0.988204613028161 \\
20  0.986541106955725\\
21	0.00354026600816926 \\
23	0.00351389115569408 \\
25	0.00348123477693136 \\
27	0.00347502236920650 \\
29	0.00346306092592276 \\
};

%%% R-SVD
\addplot+[smooth,color=teal,loosely dotted, every mark/.append style={solid}, mark=diamond]
table[row sep=crcr]{
1	1.00011894510508 \\
3	0.998016775192999 \\
5	0.996019828454718 \\
7	0.994226164611526 \\
9	0.992149474486267 \\
11	0.989908574539390 \\
13	0.988012885516550 \\
15	0.986107606763718 \\
17	0.984179489185476 \\
19	0.982187827457008 \\
20  0.980995010048426\\
21	0.00452587469194591 \\
23	0.00444467692827492 \\
25	0.00438728110051331 \\
27	0.00433394935529910 \\
29	0.00428203046860493 \\
};

%%% CoR-UTV
\addplot+[smooth,color=blue,loosely dotted, every mark/.append style={solid}, mark=star]
table[row sep=crcr]{
1	0.994058569304353 \\
3	0.992775247705414 \\
5	0.991868461341696 \\
7	0.991244880267562 \\
9	0.990789478524290 \\
11	0.990353532820881 \\
13	0.989869324717432 \\
15	0.989364778210360 \\
17	0.988878043133062 \\
19	0.988127903416585 \\
20  0.980995010048426\\
21	0.00435346340668038 \\
23	0.00431084280751085 \\
25	0.00429075162891308 \\
27	0.00427657376877998 \\
29	0.00425908481662400 \\
};

%%% R-values
\addplot+[smooth,color=orange,loosely dotted, every mark/.append style={solid}, mark=-]
table[row sep=crcr]{
1	0.991166983260341 \\
3	0.991007977438451 \\
5	0.990365692954629 \\
7	0.991096452563043 \\
9	0.990120477852585 \\
11	0.989728667520394 \\
13	0.990631813493913 \\
15	0.990763045836648 \\
17	0.990971898625437 \\
19	0.989814477492647 \\
20  0.990716374072198\\
21	0.00431072221479091 \\
23	0.00429920791427345 \\
25	0.00426407737850509 \\
27	0.00427073619783205 \\
29	0.00426293198962369 \\
};

%%% PbP-QLP
\addplot+[smooth,color=green,loosely dotted, every mark/.append style={solid}, mark=triangle]
table[row sep=crcr]{
1	0.991227311627852 \\
3	0.991053340501338 \\
5	0.990398603887648 \\
7	0.991118698096485 \\
9	0.990131937720391 \\
11	0.989728978396361 \\
13	0.990616051557443 \\
15	0.990732035985370 \\
17	0.990930079603557 \\
19	0.989764245080998 \\
20  0.990656117551894\\
21	0.00432140990530510 \\
23	0.00430676780620482 \\
25	0.00427079381246436 \\
27	0.00427397466336615 \\
29	0.00426583644018532 \\
};

\end{axis}

\end{tikzpicture}%
	\captionsetup{justification=centering,font=scriptsize}
		\caption{Singular value approximations for \texttt{LowRankLargeGap}. Left: basic PbP-QLP. Right: PI-coupled PbP-QLP.}
		\label{fig_SVM1_LG}       % Give a unique label
	\end{center}
\end{figure}
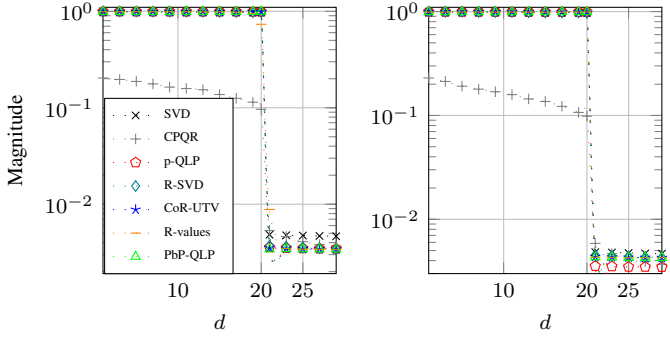

\subsection{Test Matrices with Randomly Generated Variables}
\label{subsec_LRA_SynData}
We construct four classes of input matrices. The first two
classes contain one or multiple gaps in the spectrum and are particularly designed to investigate the rank-revealing property of PbP-QLP. The second two classes have fast and slow decay singular values. For Matrices 1, 3 and 4, we further investigate the low-rank approximation accuracy of PbP-QLP in relation to those of the optimal SVD and some existing methods. We generate square matrices of order $n = 1000$.
\begin{itemize}
 \item Matrix 1 (low-rank plus noise). This rank-$k$ matrix, with $k = 20$, is formed as follows:
 \begin{equation}\label{eq_A1PlusA2}
  {\bf A} = {\bf A}_1 + {\bf A}_2,
 \end{equation}
where ${\bf A}_1 = {\bf U\Sigma V}^T$. Matrices $\bf U$ and $\bf V$ are random orthogonal, and $\bf \Sigma$ is diagonal whose entries ($\sigma_i$s) decrease linearly from 1 to $10^{-25}$, and $\sigma_{k+1} =...=\sigma_n= 0$. ${\bf A}_2 = \mu \sigma_k {\bf N}$, where $\bf N$ is a normalized Gaussian matrix. We consider two cases for $\mu$:
\begin{itemize}
 \item [i)] $\mu = 0.005$ in which the matrix has a gap $\approx 200$. This matrix is denoted by \texttt{LowRankLargeGap}.
 \item [ii)] $\mu = 0.02$ in which the matrix has a gap $\approx 50$. This matrix is denoted by \texttt{LowRankSmallGap}.
\end{itemize}
\item Matrix 2 (the devil’s stairs \cite{StewartQLP}). This challenging matrix has multiple gaps in its spectrum. The singular values are arranged analogues to a descending staircase with each step consisting of $\ell = 15$ equal singular values.
\item Matrix 3 (fast decay). This matrix is generated as ${\bf A}_1$ in \eqref{eq_A1PlusA2}, however the diagonal elements of $\bf \Sigma$ have the form $\sigma_i = e^{-i/6}$, for $i=1,..., n$.
\item Matrix 4 (slow decay). This matrix is also formed as ${\bf A}_1$, but the diagonal entries of $\bf \Sigma$ take the form $\sigma_i = i^{-2}$, for $i=1,..., n$.
\end{itemize}

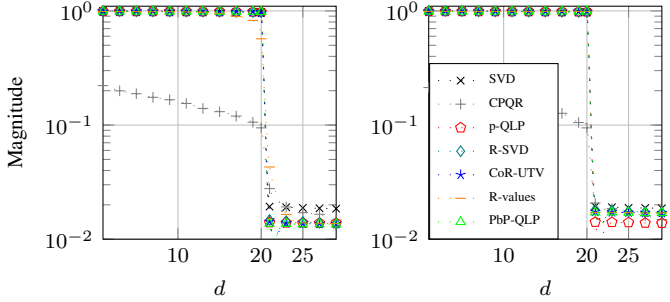
\begin{figure}[t]
	\begin{center}
			% This file was created by matlab2tikz v0.4.7 running on MATLAB 8.3.
% Copyright (c) 2008--2014, Nico Schlömer <nico.schloemer@gmail.com>
% All rights reserved.
% Minimal pgfplots version: 1.3
% 
% The latest updates can be retrieved from
%   http://www.mathworks.com/matlabcentral/fileexchange/22022-matlab2tikz
% where you can also make suggestions and rate matlab2tikz.
% 
%
% defining custom colors
\usetikzlibrary{positioning,calc}

\definecolor{mycolor1}{rgb}{0.00000,1.00000,1.00000}%
\definecolor{mycolor2}{rgb}{1.00000,0.00000,1.00000}%

\pgfplotsset{every axis label/.append style={font=\footnotesize},
every tick label/.append style={font=\footnotesize}
}

\begin{tikzpicture}[font=\footnotesize] 

\begin{axis}[%
name=ber,
ymode=log,
width  = 0.35\columnwidth,%5.63489583333333in,
height = 0.35\columnwidth,%4.16838541666667in,
scale only axis,
xmin  = 1,
xmax  = 29,
xlabel= {$d$},
xmajorgrids,
ymin = 0.01,
ymax = 1.1,
xtick       ={10,20,25},
xticklabels ={$10$,$20$, $25$},
ylabel={Magnitude},
ymajorgrids,
]

%% SVD 
\addplot+[smooth,color=black,loosely dotted, every mark/.append style={solid}, mark=x]
table[row sep=crcr]{
1	1.00029151034606\\
3	0.998209196831690\\
5	0.996436791451527\\
7	0.994321651223700\\
9	0.991958583918690\\
11	0.990457001948143\\
13	0.988555106509835\\
15	0.986569103538169\\
17	0.983032165595380\\
19	0.981346164578819\\
20  0.980471305004484\\
21	0.0193037579332143\\
23	0.0190576669713212\\
25	0.0188027460094711\\
27	0.0186483605369577\\
29	0.0185198573931089 \\
};
%% QRP
\addplot+[smooth,color=gray,loosely dotted, every mark/.append style={solid}, mark=+]
table[row sep=crcr]{
1	0.221637348904866\\
3	0.199623334812860\\
5	0.188318455282380\\
7	0.174961640500226\\
9	0.166786699634162\\
11	0.154659493200792\\
13	0.139800104784481\\
15	0.131109168856735\\
17	0.119860167406710\\
19	0.105916320976489\\
20  0.0945279547933352\\
21	0.0277474206756068\\
23	0.0191767591037053\\
25	0.0170799857534401\\
27	0.0165228603813520\\
29	0.0146908232655151 \\
};

%% p-QLP
\addplot+[smooth,color=red,loosely dotted, every mark/.append style={solid}, mark=pentagon]
table[row sep=crcr]{
1	0.993356212149863\\
3	0.990816535584202\\
5	0.989830282831982\\
7	0.988810715640499\\
9	0.988647886745223\\
11	0.988048290715561\\
13	0.986637274246497\\
15	0.986149130069432\\
17	0.983183740713667\\
19	0.980834675704706\\
20  0.978088312530385 \\
21	0.0139897353857838\\
23	0.0139345331666024\\
25	0.0138896515500266\\
27	0.0138324258113918\\
29	0.0138118777234345 \\
  };

%%% R-SVD
\addplot+[smooth,color=teal,loosely dotted, every mark/.append style={solid}, mark=diamond]
table[row sep=crcr]{
1	0.998713125574313 \\
3	0.996580293961922\\
5	0.994373544841473\\
7	0.992433306796869\\
9	0.990272049201759\\
11	0.988276092052161\\
13	0.986188366970167\\
15	0.983565573673486\\
17	0.980830051649742\\
19	0.978521280964726\\
20  0.975701861881481\\
21	0.0147512422726313\\
23	0.0144125513648555\\
25	0.0141457162434774\\
27	0.0139142244291505\\
29	0.0136806430858104 \\
};

%%% CoR-UTV
\addplot+[smooth,color=blue,loosely dotted, every mark/.append style={solid}, mark=star]
table[row sep=crcr]{
1	0.992237657077224\\
3	0.990958231291621\\
5	0.990070095201139\\
7	0.989431181717889\\
9	0.988918870664942\\
11	0.988337672288648\\
13	0.987778251452721\\
15	0.987114917456257\\
17	0.986322789663784\\
19	0.984860755078314\\
20  0.965492073531398\\
21	0.0142533781097825\\
23	0.0137988066875114\\
25	0.0137105257546444\\
27	0.0136399112989832\\
29	0.0135635428086902 \\
};

%%% R-values
\addplot+[smooth,color=orange,loosely dotted, every mark/.append style={solid}, mark=-]
table[row sep=crcr]{
1	0.988388934301841\\
3	0.986972212457466\\
5	0.985837781456779\\
7	0.985105811583646\\
9	0.984919225193267\\
11	0.981960818025625\\
13	0.973867402903590\\
15	0.952912034956957\\
17	0.895700330342339\\
19	0.825465478744916\\
20  0.569999409748461 \\
21	0.0429168177508519\\
23	0.0164837794735170\\
25	0.0138791632397389\\
27	0.0137836481693879\\
29	0.0137218231806900 \\
};

%%% PbP-QLP
\addplot+[smooth,color=green,loosely dotted, every mark/.append style={solid}, mark=triangle]
table[row sep=crcr]{
1	0.989774870526251\\
3	0.988944449403373\\
5	0.989011708514057\\
7	0.989033411925786\\
9	0.989935185211640\\
11	0.988240776472382\\
13	0.988764436602389\\
15	0.987568469065075\\
17	0.986833787449354\\
19	0.986199483907392\\
20  0.980936170946759\\
21	0.0137056883994996\\
23	0.0136490966099900\\
25	0.0135336670168294\\
27	0.0135532099335007\\
29	0.0135195323421444 \\
};

\end{axis}

\begin{axis}[%
name=SumRate,
at={($(ber.east)+(35,0em)$)},
		anchor= west,
ymode=log,
width  = 0.35\columnwidth,%5.63489583333333in,
height = 0.35\columnwidth,%4.16838541666667in,
scale only axis,
xmin  = 1,
xmax  = 29,
xlabel= {$d$},
xmajorgrids,
xtick       ={10,20,25},
xticklabels ={$10$,$20$, $25$},
ymin = 0.01,
ymax = 1.1,
ylabel={},
ymajorgrids,
%title = {\texttt{HighNoiseLowRank}}]
%ytick       ={0.0973831, 0.0973830 , 0.0973829, 0.0973828},
%yticklabels ={$9.73831$, $9.73830$ , $9.73829$, $9.73828$},
legend entries={SVD,CPQR,p-QLP, R-SVD,CoR-UTV,R-values,PbP-QLP}, 
legend style={at={(0.557,0.756)},anchor=north east,draw=black,fill=white,legend cell align=left,font=\tiny, }%legend columns=2}
]
%% SVD 
\addplot+[smooth,color=black,loosely dotted, every mark/.append style={solid}, mark=x]
table[row sep=crcr]{
1	1.00041502866528 \\
3	0.998317503067605 \\
5	0.996381676975817 \\
7	0.993588573834765 \\
9	0.992172025576997 \\
11	0.990701864434061 \\
13	0.988064237272210 \\
15	0.986505695753385 \\
17	0.983296630913779 \\
19	0.982271422373951 \\
20  0.980278493878066\\
21	0.0193428257276055 \\
23	0.0190558624588639 \\
25	0.0188199295058889 \\
27	0.0187706838163657 \\
29	0.0186452047331613 \\
};
%% QRP
\addplot+[smooth,color=gray,loosely dotted, every mark/.append style={solid}, mark=+]
table[row sep=crcr]{
1	0.213137013009497 \\
3	0.205704024706918\\
5	0.192630172323273\\
7	0.181076733613953\\
9	0.168607678816843\\
11	0.156122564049959\\
13	0.147638495868835\\
15	0.132575870518650\\
17	0.126710185564912\\
19	0.105233116609252\\
20  0.0945279547933352\\
21	0.0207274532606905\\
23	0.0185238767144438\\
25	0.0173497740564944\\
27	0.0160542824595758\\
29	0.0148538553378357 \\
};

%% p-QLP
\addplot+[smooth,color=red,loosely dotted, every mark/.append style={solid}, mark=pentagon]
table[row sep=crcr]{
1	0.992020535808872 \\
3	0.990345901234738 \\
5	0.989999766529641 \\
7	0.989245268228891 \\
9	0.988242349844915 \\
11	0.987471035304831 \\
13	0.987400144721693 \\
15	0.986686892103137 \\
17	0.986310158772386 \\
19	0.983247478969001 \\
20  0.982699453216876\\
21	0.0141343302356208 \\
23	0.0140726421751099 \\
25	0.0140183975991521 \\
27	0.0138789513728708 \\
29	0.0138338278261158 \\
};

%%% R-SVD
\addplot+[smooth,color=teal,loosely dotted, every mark/.append style={solid}, mark=diamond]
table[row sep=crcr]{
1	1.00041502866528 \\
3	0.998317503067606 \\
5	0.996381676975817 \\
7	0.993588573834765 \\
9	0.992172025576996 \\
11	0.990701864434061 \\
13	0.988064237272210 \\ 
15	0.986505695753385 \\
17	0.983296630913779 \\
19	0.982271422373952 \\
20  0.980278493878065\\
21	0.0181424210950975 \\
23	0.0177642789152181 \\
25	0.0175415177019094 \\
27	0.0173119961467002 \\
29	0.0171133844656735 \\
};

%%% CoR-UTV
\addplot+[smooth,color=blue,loosely dotted, every mark/.append style={solid}, mark=star]
table[row sep=crcr]{
1	0.993974326356511 \\
3	0.992670264018833 \\
5	0.992046504913958 \\
7	0.991505638623750 \\
9	0.991064895165792 \\
11	0.990555614687913 \\
13	0.990015581547711 \\
15	0.989275835599433 \\
17	0.988630001543205 \\
19	0.987932207775993 \\
20  0.987330297661792\\
21	0.0173980355937417 \\
23	0.0172449506680259 \\
25	0.0171533563632456 \\
27	0.0170708344930266 \\
29	0.0169877347164649 \\
};

%%% R-values
\addplot+[smooth,color=orange,loosely dotted, every mark/.append style={solid}, mark=-]
table[row sep=crcr]{
1	0.991078619747638 \\
3	0.991475414282189 \\
5	0.990170620890864 \\
7	0.990279214786389 \\
9	0.989827882334690 \\
11	0.991394444073334 \\
13	0.989882472836051 \\
15	0.989931372476657 \\
17	0.991106190684721 \\
19	0.990253028630001 \\
20  0.990349147183858\\
21	0.0171591174055811 \\
23	0.0171714934100704 \\
25	0.0170833337991692 \\
27	0.0169746045316788 \\
29	0.0170177413111755  \\
};

%%% PbP-QLP
\addplot+[smooth,color=green,loosely dotted, every mark/.append style={solid}, mark=triangle]
table[row sep=crcr]{
1	0.991140261099242 \\
3	0.991533451725674 \\
5	0.990209073493214 \\
7	0.990304820293339 \\
9	0.989834433938222 \\
11	0.991391436968374 \\
13	0.989867347051642 \\
15	0.989909333523002 \\
17	0.991054561840040 \\
19	0.990189317810404 \\
20  0.990282870701196\\
21	0.0172016890456978 \\
23	0.0172083289563207 \\
25	0.0171048684599449 \\
27	0.0169902004897114 \\
29	0.0170188587153099 \\
};

\end{axis}

\end{tikzpicture}%
	\captionsetup{justification=centering,font=scriptsize}
		\caption{Singular value approximations of \texttt{LowRankSmallGap}. Left: basic PbP-QLP. Right: PI-coupled PbP-QLP with $q=2$.}
		\label{fig_SVM1_SG}       % Give a unique label
	\end{center}
\end{figure}
 
\begin{figure}[t]
	\begin{center}     
		\input{Plots/ASV_Mat2}
	\captionsetup{justification=centering,font=scriptsize} 
		\caption{Singular value approximations for Matrix 2. Left: basic PbP-QLP. Right: PI-coupled PbP-QLP with $q=2$.}
		\label{fig_SVM2}       % Give a unique label
	\end{center}
\end{figure}

\begin{figure}[t]
	\begin{center}
			% This file was created by matlab2tikz v0.4.7 running on MATLAB 8.3.
% Copyright (c) 2008--2014, Nico Schlömer <nico.schloemer@gmail.com>
% All rights reserved.
% Minimal pgfplots version: 1.3
% 
% The latest updates can be retrieved from
%   http://www.mathworks.com/matlabcentral/fileexchange/22022-matlab2tikz
% where you can also make suggestions and rate matlab2tikz.
% 
%
% defining custom colors
\usetikzlibrary{positioning,calc}

\definecolor{mycolor1}{rgb}{0.00000,1.00000,1.00000}%
\definecolor{mycolor2}{rgb}{1.00000,0.00000,1.00000}%

\pgfplotsset{every axis label/.append style={font=\footnotesize},
every tick label/.append style={font=\footnotesize}
}

\begin{tikzpicture}[font=\footnotesize] 

\begin{axis}[%
name=ber,
ymode=log,
width  = 0.35\columnwidth,%5.63489583333333in,
height = 0.35\columnwidth,%4.16838541666667in,
scale only axis,
xmin  = 1,
xmax  = 100,
xlabel= {$d$},
xmajorgrids,
ymin = 1e-08 ,
ymax = 0.85,
xtick       ={20,50,80},
xticklabels ={$20$,$50$, $80$},
ylabel={Magnitude},
ymajorgrids,
]

%% SVD 
\addplot+[smooth,color=black,loosely dotted, every mark/.append style={solid}, mark=x]
table[row sep=crcr]{
1	0.846481724890614\\
10	0.188875602837562\\
19	0.0421438435092764\\
28	0.00940356255149520\\
37	0.00209821841808090\\
46	0.000468175811652779\\
55	0.000104464143831706\\
64	2.33091011429374e-05\\
73	5.20096347093910e-06\\
82	1.16049181219835e-06\\
91	2.58940723907174e-07\\
100	5.77774851941163e-08 \\
};
%% QRP
\addplot+[smooth,color=gray,loosely dotted, every mark/.append style={solid}, mark=+]
table[row sep=crcr]{
1	0.101796945113610\\
10	0.0290979448855278\\
19	0.00772310296705444\\
28	0.00201400043728236\\
37	0.000507164357744638\\
46	0.000106476675132339\\
55	3.21999881801175e-05\\
64	5.82513237175237e-06\\
73	1.49970145838478e-06\\
82	4.12296906462167e-07\\
91	9.25999790752558e-08\\
100	1.65373442919269e-08 \\
};

%% p-QLP
\addplot+[smooth,color=red,loosely dotted, every mark/.append style={solid}, mark=pentagon]
table[row sep=crcr]{
1	0.770280388514038\\
10	0.207018433190476\\
19	0.0407589172966153\\
28	0.00875675179389580\\
37	0.00195836789266118\\
46	0.000456444763975982\\
55	0.000113934079539479\\
64	2.33238245655319e-05\\
73	5.65788746114508e-06\\
82	1.17351061094282e-06\\
91	2.99101618858826e-07\\
100	6.75517152401559e-08\\
  };

%%% R-SVD
\addplot+[smooth,color=teal,loosely dotted, every mark/.append style={solid}, mark=diamond]
table[row sep=crcr]{
1	0.846481724890612 \\
10	0.188875602837552\\
19	0.0421438435092483\\
28	0.00940356255133278\\
37	0.00209821841747312\\
46	0.000468175808069742\\
55	0.000104464133953652\\
64	2.33090411210092e-05\\
73	5.20074958100938e-06\\
82	1.15904403502148e-06\\
91	2.52240523828576e-07\\
100	3.81430833876704e-08 \\
};

%%% CoR-UTV
\addplot+[smooth,color=blue,loosely dotted, every mark/.append style={solid}, mark=star]
table[row sep=crcr]{
1	0.741748878255895\\
10	0.187095538557308\\
19	0.0416887305096727\\
28	0.00953867321730872\\
37	0.00205235982218562\\
46	0.000474624882184885\\
55	0.000101837988632738\\
64	2.32613945695046e-05\\
73	5.04409835684428e-06\\
82	1.14489101314248e-06\\
91	2.56991399470615e-07\\
100	4.24548712734014e-08 \\
};

%%% PbP-QLP
\addplot+[smooth,color=green,loosely dotted, every mark/.append style={solid}, mark=triangle]
table[row sep=crcr]{
1	0.710344011656983\\
10	0.189320621831445\\
19	0.0441841204202094\\
28	0.00908668266879675\\
37	0.00216564559496690\\
46	0.000473751033025878\\
55	0.000113559766349505\\
64	2.58174634257633e-05\\
73	5.15358608423556e-06\\
82	1.30293051074057e-06\\
91	2.65329124999001e-07\\
100	4.64648821499765e-08 \\
};

\end{axis}

\begin{axis}[%
name=SumRate,
at={($(ber.east)+(35,0em)$)},
		anchor= west,
ymode=log,
width  = 0.35\columnwidth,%5.63489583333333in,
height = 0.35\columnwidth,%4.16838541666667in,
scale only axis,
xmin  = 1,
xmax  = 100,
xlabel= {$d$},
xmajorgrids,
ymin = 1e-08 ,
ymax = 0.85,
xtick       ={20,50,80},
xticklabels ={$20$,$50$, $80$},
ylabel={},
ymajorgrids,
%title = {\texttt{HighNoiseLowRank}}]
%ytick       ={0.0973831, 0.0973830 , 0.0973829, 0.0973828},
%yticklabels ={$9.73831$, $9.73830$ , $9.73829$, $9.73828$},
%legend entries={SVD,CPQR,p-QLP, R-SVD,CoR-UTV,R-values,PbP-QLP}, 
%legend style={at={(1,0.82)},anchor=north east,draw=black,fill=white,legend cell align=left,font=\tiny, legend columns=2}
]
%% SVD 
\addplot+[smooth,color=black,loosely dotted, every mark/.append style={solid}, mark=x]
table[row sep=crcr]{
1	0.846481724890614\\
10	0.188875602837562\\
19	0.0421438435092764\\
28	0.00940356255149521\\
37	0.00209821841808090\\
46	0.000468175811652777\\
55	0.000104464143831705\\
64	2.33091011429353e-05\\
73	5.20096347093670e-06\\
82	1.16049181219677e-06\\
91	2.58940723906965e-07\\
100	5.77774851944854e-08\\
};
%% QRP
\addplot+[smooth,color=gray,loosely dotted, every mark/.append style={solid}, mark=+]
table[row sep=crcr]{
1	0.113330138369121\\
10	0.0282555529691867\\
19	0.00700953629324309\\
28	0.00193729704305795\\
37	0.000488250602449067\\
46	0.000109603233269805\\
55	3.00444658232892e-05\\
64	6.29283895582144e-06\\
73	1.53356379286459e-06\\
82	3.58482814224810e-07\\
91	9.63799398375781e-08\\
100	2.23856038817618e-08\\
};

%% p-QLP
\addplot+[smooth,color=red,loosely dotted, every mark/.append style={solid}, mark=pentagon]
table[row sep=crcr]{
1	0.791904216532134\\
10	0.186892928555194\\
19	0.0478808560383592\\
28	0.0119594405036101\\
37	0.00224402666218702\\
46	0.000383935468330080\\
55	0.000115213679405789\\
64	2.25072622890028e-05\\
73	5.38709127771417e-06\\
82	1.13090670818724e-06\\
91	2.96141415931028e-07\\
100	5.98625258236031e-08 \\
};

%%% R-SVD
\addplot+[smooth,color=teal,loosely dotted, every mark/.append style={solid}, mark=diamond]
table[row sep=crcr]{
1	0.846481724890614\\
10	0.188875602837562\\
19	0.0421438435092764\\
28	0.00940356255149521\\
37	0.00209821841808090\\
46	0.000468175811652778\\
55	0.000104464143831706\\
64	2.33091011429358e-05\\
73	5.20096347093825e-06\\
82	1.16049181219608e-06\\
91	2.58940710726748e-07\\
100	5.28493833752479e-08 \\
};

%%% CoR-UTV
\addplot+[smooth,color=blue,loosely dotted, every mark/.append style={solid}, mark=star]
table[row sep=crcr]{
1	0.825147343366495\\
10	0.191170180164398\\
19	0.0433526694800770\\
28	0.00945198044897501\\
37	0.00214729782367113\\
46	0.000473380146745765\\
55	0.000107540576557069\\
64	2.37360550659197e-05\\
73	5.23950554754779e-06\\
82	1.15723492523238e-06\\
91	2.52857613375741e-07\\
100	5.40301246755440e-08\\
};

%%% PbP-QLP
\addplot+[smooth,color=green,loosely dotted, every mark/.append style={solid}, mark=triangle]
table[row sep=crcr]{
1	0.799483726363520\\
10	0.187152015364375\\
19	0.0411539763435390\\
28	0.00944412395659602\\
37	0.00204788697583449\\
46	0.000472318880277797\\
55	0.000100619419162877\\
64	2.32024083418059e-05\\
73	5.03543102420083e-06\\
82	1.21270455241704e-06\\
91	2.56390569272461e-07\\
100	5.76087698156951e-08 \\
};

\end{axis}

\end{tikzpicture}%
	\captionsetup{justification=centering,font=scriptsize}
		\caption{Singular value approximations for Matrix 3. Left: basic PbP-QLP. Right: PI-coupled PbP-QLP with $q=2$.}
		\label{fig_SVM3}       % Give a unique label
	\end{center}    
\end{figure}
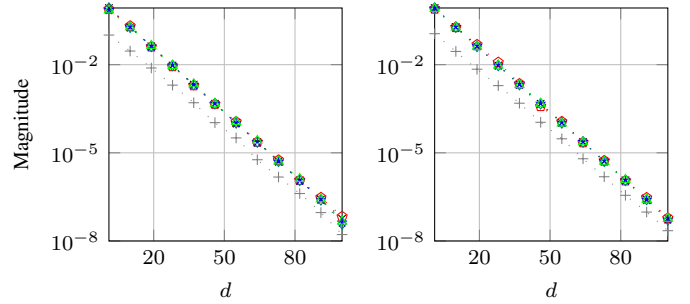

\begin{figure}[t]
	\begin{center}
			% This file was created by matlab2tikz v0.4.7 running on MATLAB 8.3.
% Copyright (c) 2008--2014, Nico Schlömer <nico.schloemer@gmail.com>
% All rights reserved.
% Minimal pgfplots version: 1.3
% 
% The latest updates can be retrieved from
%   http://www.mathworks.com/matlabcentral/fileexchange/22022-matlab2tikz
% where you can also make suggestions and rate matlab2tikz.
% 
%
% defining custom colors
\usetikzlibrary{positioning,calc}

\definecolor{mycolor1}{rgb}{0.00000,1.00000,1.00000}%
\definecolor{mycolor2}{rgb}{1.00000,0.00000,1.00000}%

\pgfplotsset{every axis label/.append style={font=\footnotesize},
every tick label/.append style={font=\footnotesize}
}

\begin{tikzpicture}[font=\footnotesize] 

\begin{axis}[%
name=ber,
ymode=log,
width  = 0.35\columnwidth,%5.63489583333333in,
height = 0.35\columnwidth,%4.16838541666667in,
scale only axis,
xmin  = 1,
xmax  = 100,
xlabel= {$d$},
xmajorgrids,
ymin = 4e-05 ,
ymax = 1.0,
xtick       ={20,50,80},
xticklabels ={$20$,$50$, $80$},
ylabel={Magnitude},
ymajorgrids,
legend entries = {SVD,CPQR,p-QLP, R-SVD,CoR-UTV,PbP-QLP},
legend style={at={(1,1)},anchor=north east,draw=black,fill=white,legend cell align=left,font=\tiny},
]

%% SVD 
\addplot+[smooth,color=black,loosely dotted, every mark/.append style={solid}, mark=x]
table[row sep=crcr]{
1	1\\
10	0.0100000000000000 \\
19	0.00277008310249308 \\
28	0.00127551020408164\\
37	0.000730460189919651\\
46	0.000472589792060491\\
55	0.000330578512396694\\
64	0.000244140624999999\\
73	0.000187652467629948\\
82	0.000148720999405116\\
91	0.000120758362516605\\
100	0.000100000000000001 \\
};
%% QRP
\addplot+[smooth,color=gray,loosely dotted, every mark/.append style={solid}, mark=+]
table[row sep=crcr]{
1	0.108147807494043\\
10	0.00137819571704012\\
19	0.000570534333952616\\
28	0.000303183852821756\\
37	0.000199025712414852\\
46	0.000139753062092784\\
55	0.000105707492831590\\
64	8.24438309823321e-05\\
73	6.95054748117974e-05\\
82	5.77336374292005e-05\\
91	4.84031546154651e-05\\
100	4.24582956485411e-05 \\
};

%% p-QLP
\addplot+[smooth,color=red,loosely dotted, every mark/.append style={solid}, mark=pentagon]
table[row sep=crcr]{
1	0.998783234537722\\
10	0.00867094200232960\\
19	0.00256768929607535\\
28	0.00110014815666561\\
37	0.000689430913300433\\
46	0.000410348663339486\\
55	0.000300750267573210\\
64	0.000226876736575713\\
73	0.000169294873904128\\
82	0.000136453247927060\\
91	0.000105271321682450\\
100	8.96108934510124e-05\\
  };

%%% R-SVD
\addplot+[smooth,color=teal,loosely dotted, every mark/.append style={solid}, mark=diamond]
table[row sep=crcr]{
1	0.999999992799706\\
10	0.00999925583155168\\
19	0.00276748203902944\\
28	0.00126918492059314\\
37	0.000720971586473956\\
46	0.000459250618745927\\
55	0.000312350459498625\\
64	0.000222319809000909\\
73	0.000160687779542780\\
82	0.000119233117442842\\
91	8.65956471828448e-05\\
100	5.67603200777976e-05 \\
};

%%% CoR-UTV
\addplot+[smooth,color=blue,loosely dotted, every mark/.append style={solid}, mark=star]
table[row sep=crcr]{
1	0.978769639513302\\
10	0.00961869647269553\\
19	0.00264065588375796\\
28	0.00123638426238920\\
37	0.000706775355292714\\
46	0.000448012714337468\\
55	0.000305046983762519\\
64	0.000217571271650668\\
73	0.000161499527547124\\
82	0.000122784578220171\\
91	9.38371241836711e-05\\
100	6.85492166540257e-05 \\
};

%%% PbP-QLP
\addplot+[smooth,color=green,loosely dotted, every mark/.append style={solid}, mark=triangle]
table[row sep=crcr]{
1	0.996431189055630\\
10	0.00920161742909442\\
19	0.00267477209173145\\
28	0.00125621723164286\\
37	0.000663052507838673\\
46	0.000432788926497546\\
55	0.000303489240264600\\
64	0.000213744101920590\\
73	0.000156416146818766\\
82	0.000124011482593289\\
91	9.51041937864342e-05\\
100	7.40639362626209e-05 \\
};

\end{axis}

\begin{axis}[%
name=SumRate,
at={($(ber.east)+(35,0em)$)},
		anchor= west,
ymode=log,
width  = 0.35\columnwidth,%5.63489583333333in,
height = 0.35\columnwidth,%4.16838541666667in,
scale only axis,
xmin  = 1,
xmax  = 100,
xlabel= {$d$},
xmajorgrids,
ymin = 4e-05 ,
ymax = 1.0,
xtick       ={20,50,80},
xticklabels ={$20$,$50$, $80$},
ylabel={},
ymajorgrids,
%title = {\texttt{HighNoiseLowRank}}]
%ytick       ={0.0973831, 0.0973830 , 0.0973829, 0.0973828},
%yticklabels ={$9.73831$, $9.73830$ , $9.73829$, $9.73828$},
%legend entries={SVD,CPQR,p-QLP, R-SVD,CoR-UTV,R-values,PbP-QLP}, 
%legend style={at={(1,0.82)},anchor=north east,draw=black,fill=white,legend cell align=left,font=\tiny, legend columns=2}
]

%% SVD 
\addplot+[smooth,color=black,loosely dotted, every mark/.append style={solid}, mark=x]
table[row sep=crcr]{
1	1.00000000000000  \\
10	0.0100000000000000 \\
19	0.00277008310249308 \\
28	0.00127551020408163 \\
37	0.000730460189919647 \\
46	0.000472589792060491 \\
55	0.000330578512396693 \\
64	0.000244140625000000 \\
73	0.000187652467629951 \\
82	0.000148720999405115 \\
91	0.000120758362516603 \\
100	0.000100000000000001 \\
};
%% QRP
\addplot+[smooth,color=gray,loosely dotted, every mark/.append style={solid}, mark=+]
table[row sep=crcr]{
1	0.106477024447873\\
10	0.00151192462803805\\
19	0.000539714761594633\\
28	0.000304823468044742\\
37	0.000186048631382711\\
46	0.000143453422366423\\
55	0.000103707014473828\\
64	8.42416340334205e-05\\
73	6.97895101325440e-05\\
82	5.76945303925486e-05\\
91	4.88085727860534e-05\\
100	4.29625663423422e-05 \\
};

%% p-QLP
\addplot+[smooth,color=red,loosely dotted, every mark/.append style={solid}, mark=pentagon]
table[row sep=crcr]{
1	0.993394637105414 \\
10	0.0102124310096709\\
19	0.00255511526858538\\
28	0.00107451077313198\\
37	0.000660748049519463\\
46	0.000430970522823606\\
55	0.000295766115908191\\
64	0.000216876666181241\\
73	0.000183005077985006\\
82	0.000139383376023749\\
91	0.000107073493439239\\
100	8.79893711263225e-05 \\
};

%%% R-SVD
\addplot+[smooth,color=teal,loosely dotted, every mark/.append style={solid}, mark=diamond]
table[row sep=crcr]{
1	1.0000000000000000\\
10	0.0100000000000000\\
19	0.00277008310249307\\
28	0.00127551020407554\\
37	0.000730460188823872\\
46	0.000472589745010977\\
55	0.000330577599652967\\
64	0.000244129047492918\\
73	0.000187545836134509\\
82	0.000147841459058512\\
91	0.000117690064422867\\
100	8.87065510216358e-05 \\
};

%%% CoR-UTV
\addplot+[smooth,color=blue,loosely dotted, every mark/.append style={solid}, mark=star]
table[row sep=crcr]{
1	0.999811874704634\\
10	0.0100157206806797\\
19	0.00277521895261323\\
28	0.00126649254322882\\
37	0.000724498677636001\\
46	0.000469576316014261\\
55	0.000329502283232737\\
64	0.000243061675340583\\
73	0.000187764036964957\\
82	0.000146685149003985\\
91	0.000119952487066360\\
100	9.40620985357365e-05 \\
};

%%% PbP-QLP
\addplot+[smooth,color=green,loosely dotted, every mark/.append style={solid}, mark=triangle]
table[row sep=crcr]{
1	0.999999122176778\\
10	0.0100221968784361\\
19	0.00275093962708318\\
28	0.00129780293022359\\
37	0.000703665310021298\\
46	0.000468912132141954\\
55	0.000326633633322268\\
64	0.000242464213692858\\
73	0.000187156712108497\\
82	0.000149077242847306\\
91	0.000119575541749383\\
100	9.78190073517281e-05 \\
};

\end{axis}

\end{tikzpicture}%
	\captionsetup{justification=centering,font=scriptsize}
		\caption{Singular value approximations for Matrix 4. Left: basic PbP-QLP. Right: PI-coupled PbP-QLP with $q=2$.}
		\label{fig_SVM4}       % Give a unique label
	\end{center}
\end{figure}
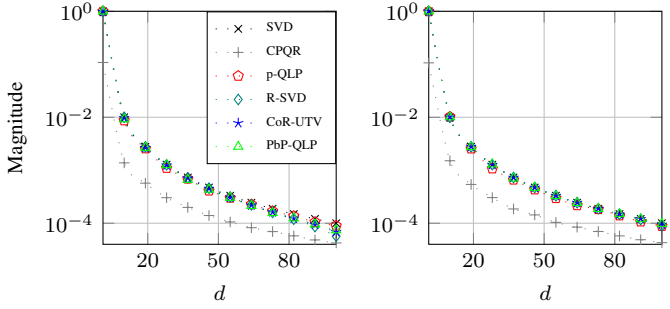

The results for singular values estimation are plotted in Figs. \ref{fig_SVM1_LG}-\ref{fig_SVM4}. We make several observations:
\begin{enumerate} 
 \item The numerical rank of both matrices \texttt{LowRankLargeGap} and \texttt{LowRankSmallGap} is strongly revealed in $\bf R$ generated by PbP-QLP with no PI technique. This is due to the fact that the gaps in the spectrums of these matrices are well-defined. R-values of PI-coupled PbP-QLP are as accurate as those of the optimal SVD. Though CPQR reveals the gaps, it substantially underestimates the principal singular values of the two matrices. Figs. \ref{fig_SVM1_LG} and \ref{fig_SVM1_SG} show that PbP-QLP is a rank-revealer.
 \item For Matrix 2, R-values of basic PbP-QLP do not clearly disclose the gaps in matix's spectrum. This is because the gaps are not substantial. However, basic PbP-QLP strongly reveals the gaps, which shows the procedure that leads to the formation $\bf R$ provides a good first step for PbP-QLP. R-values of PI-coupled PbP-QLP clearly disclose that gaps, due to the effect of PI (see Remark \ref{Remark_Basis}). CPQR reveals the gaps, however it underestimates the singular values. 
 \item For Matrices 3 and 4, PbP-QLP, though it only uses QR factorization, provides highly accurate singular values, showing similar performance as R-SVD and CoR-UTV.
 \end{enumerate}

 The results presented in Figs. \ref{fig_SVM1_LG}-\ref{fig_SVM4} demonstrate the applicability of PbP-QLP in accurately estimating the singular values of matrices from different classes.
 
For Matrices 1 (\texttt{LowRankSmallGap}), 3 and 4, we now compare the accuracy of PbP-QLP, together with other considered algorithms, for rank-$d$ approximations, where $d\ge 1$. We calculate the error as follow:
\begin{equation}\notag 
 \text{Error} = \|{\bf A} - {\bf A}_\text{Approx}\|_2.
\end{equation} 

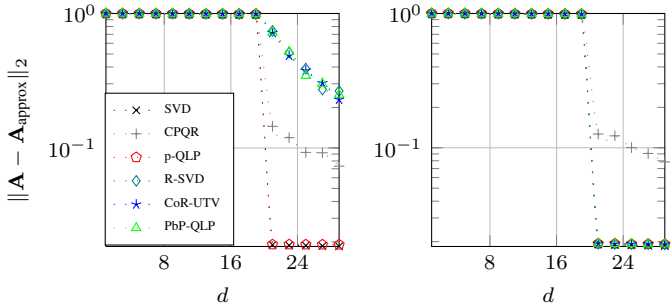
\begin{figure}[t]
	\begin{center}
			% This file was created by matlab2tikz v0.4.7 running on MATLAB 8.3.
% Copyright (c) 2008--2014, Nico Schlömer <nico.schloemer@gmail.com>
% All rights reserved.
% Minimal pgfplots version: 1.3
%
% The latest updates can be retrieved from
%   http://www.mathworks.com/matlabcentral/fileexchange/22022-matlab2tikz
% where you can also make suggestions and rate matlab2tikz.
%
%
% defining custom colors
\usetikzlibrary{positioning,calc}

\definecolor{mycolor1}{rgb}{0.00000,1.00000,1.00000}%
\definecolor{mycolor2}{rgb}{1.00000,0.00000,1.00000}%

\pgfplotsset{every axis label/.append style={font=\footnotesize},
every tick label/.append style={font=\footnotesize}
}

\begin{tikzpicture}[font=\footnotesize]

\begin{axis}[%
name=ber,
ymode=log,
width  = 0.35\columnwidth,%5.63489583333333in,
height = 0.35\columnwidth,%4.16838541666667in,
scale only axis,
xmin  = 1,
xmax  = 29,
xlabel= {$d$},
xmajorgrids,
ymin = 0.0185 ,
ymax = 1.0005,
xtick       ={8,16,24},
xticklabels ={$8$,$16$, $24$},
ylabel={$\|{\bf A}- {\bf A}_{\text{approx}}\|_2$},
ymajorgrids,
legend entries = {SVD,CPQR,p-QLP, R-SVD,CoR-UTV,PbP-QLP},
legend style={at={(0.55,0.66)},anchor=north east,draw=black,fill=white,legend cell align=left,font=\tiny},
]

%% SVD
\addplot+[smooth,color=black,loosely dotted, every mark/.append style={solid}, mark=x]
table[row sep=crcr]{
1	0.999570569302477 \\
3	0.997252853345501 \\
5	0.995071412143212 \\
7	0.993075710682820 \\
9	0.990657229677111 \\
11	0.989223971621773 \\
13	0.987467355432080 \\
15	0.985534909565476 \\
17	0.983152504845188 \\
19	0.980636336825429 \\
21	0.0190293715067733 \\
23	0.0188565599296432 \\
25	0.0187922189970834 \\
27	0.0186408570383796 \\
29	0.0185586724385111 \\
};
%% QRP
\addplot+[smooth,color=gray,loosely dotted, every mark/.append style={solid}, mark=+]
table[row sep=crcr]{
1	1.00052294912575  \\
3	1.00047098959949 \\
5	0.998318741880043 \\
7	0.997764787966101 \\
9	0.997556577653874 \\
11	0.994699167477404 \\
13	0.994658501589592 \\
15	0.993510402072746 \\
17	0.992528061994484 \\
19	0.990595706122273 \\
21	0.144758724851488 \\
23	0.119306896018860 \\
25	0.0926600847362123 \\
27	0.0920623899939014 \\
29	0.0732504664976558 \\
};
%% p-QLP
\addplot+[smooth,color=red,loosely dotted, every mark/.append style={solid}, mark=pentagon]
table[row sep=crcr]{
1	0.999572092403808 \\
3	0.998423776690191\\
5	0.997948528512090\\
7	0.996457434321819\\
9	0.995260699633246\\
11	0.994689117004576\\
13	0.993753650636961\\
15	0.992426521068703\\
17	0.990827853360338\\
19	0.990546601301733\\
21	0.0191280209905398\\
23	0.0191207548967000\\
25	0.0191019302979717\\
27	0.0190865104289324\\
29	0.0190773132584625 \\
  };

%%% R-SVD
\addplot+[smooth,color=teal,loosely dotted, every mark/.append style={solid}, mark=diamond]
table[row sep=crcr]{
1	1.00040237075777 \\
3	0.999700807588296\\
5	0.999260158758416\\
7	0.998115144190958\\
9	0.998026977587902\\
11	0.996694324407468\\
13	0.996125015251198\\
15	0.995104668537244\\
17	0.993196427195315\\
19	0.991310190101145\\
21	0.742243792413221\\
23	0.511931907711211\\
25	0.389212059718362\\
27	0.271777967217203\\
29	0.265344010583449 \\
};

%%% CoR-UTV
\addplot+[smooth,color=blue,loosely dotted, every mark/.append style={solid}, mark=star]
table[row sep=crcr]{
1	1.00035888607325  \\
3	0.999775088438290 \\
5	0.999184507455002 \\
7	0.998357355881379 \\
9	0.997963370498242 \\
11	0.996795567252846 \\
13	0.995824111781642 \\
15	0.994959098969532 \\
17	0.993223685219906 \\
19	0.990589504110027 \\
21	0.717558505928820 \\
23	0.486302312885089 \\
25	0.376885048291056 \\
27	0.303963651458986 \\
29	0.228881922948365 \\
};

%%% PbP-QLP
\addplot+[smooth,color=green,loosely dotted, every mark/.append style={solid}, mark=triangle]
table[row sep=crcr]{
1	1.00024529600329  \\
3	0.999684040077883 \\
5	0.999433536500737 \\
7	0.998564158771771 \\
9	0.997551107322819 \\
11	0.996866174765849 \\
13	0.996132190508707 \\
15	0.995106409693556 \\
17	0.993048048716674 \\
19	0.989683335730791 \\
21	0.722805821858772 \\
23	0.518530753061284 \\
25	0.344800143678519 \\
27	0.299914529889557 \\
29	0.249195385969955 \\
};

\end{axis}

\begin{axis}[%
name=SumRate,
at={($(ber.east)+(35,0em)$)},
		anchor= west,
ymode=log,
width  = 0.35\columnwidth,%5.63489583333333in,
height = 0.35\columnwidth,%4.16838541666667in,
scale only axis,
xmin  = 1,
xmax  = 29,
xlabel= {$d$},
xmajorgrids,
ymin = 0.0184 ,
ymax = 1.0008,
xtick       ={8,16,24},
xticklabels ={$8$,$16$, $24$},
ylabel={},
ymajorgrids,
%title = {\texttt{HighNoiseLowRank}}]
%ytick       ={0.0973831, 0.0973830 , 0.0973829, 0.0973828},
%yticklabels ={$9.73831$, $9.73830$ , $9.73829$, $9.73828$},
%legend entries={SVD,CPQR,p-QLP, R-SVD,CoR-UTV,R-values,PbP-QLP},
%legend style={at={(1,0.82)},anchor=north east,draw=black,fill=white,legend cell align=left,font=\tiny, legend columns=2}
]

%% SVD
\addplot+[smooth,color=black,loosely dotted, every mark/.append style={solid}, mark=x]
table[row sep=crcr]{
1	0.999633599102499  \\
3	0.998247120222202 \\
5	0.995405127522539 \\
7	0.992896234759186 \\
9	0.991260493527266 \\
11	0.988980242971499 \\
13	0.987585047743827 \\
15	0.984703369842752 \\
17	0.982514774111086 \\
19	0.981045361812937 \\
21	0.0191640349407404 \\
23	0.0188776207721358 \\
25	0.0187553118793279 \\
27	0.0186449687917534 \\
29	0.0185462705451452 \\
};
%% QRP
\addplot+[smooth,color=gray,loosely dotted, every mark/.append style={solid}, mark=+]
table[row sep=crcr]{
1	1.00079790845205  \\
3	1.00022043828371 \\
5	0.999706637689464 \\
7	0.998925110791300 \\
9	0.998547154339126 \\
11	0.997047743907160 \\
13	0.996963178914108 \\
15	0.996041619698579 \\
17	0.994605735926618 \\
19	0.989115504257425 \\
21	0.126583051743095 \\
23	0.122795175231528 \\
25	0.100316578281645 \\
27	0.0906823049781166 \\
29	0.0784930727642078 \\
};

%% p-QLP
\addplot+[smooth,color=red,loosely dotted, every mark/.append style={solid}, mark=pentagon]
table[row sep=crcr]{
1	1.00064249824974  \\
3	1.00056597425233 \\
5	1.00042314428791 \\
7	0.997320027277147 \\
9	0.996511319682661 \\
11	0.995511084836389 \\
13	0.994281897778221 \\
15	0.992277409836060 \\
17	0.990507332477179 \\
19	0.989035583398824 \\
21	0.0193014432465545 \\
23	0.0192978874808731 \\
25	0.0191765902523812 \\
27	0.0191746660062797 \\
29	0.0191349471890114 \\
};

%%% R-SVD
\addplot+[smooth,color=teal,loosely dotted, every mark/.append style={solid}, mark=diamond]
table[row sep=crcr]{
1	1.00060277763423 \\
3	1.00014052809995\\
5	0.999419542831050\\
7	0.998893642933415\\
9	0.998318964109689\\
11	0.997026553848247\\
13	0.995829764006396\\
15	0.995096579439550\\
17	0.993104452499908\\
19	0.990917574981743\\
21	0.0192952677123299\\
23	0.0192356654382750\\
25	0.0191709744155895\\
27	0.0191416147628088\\
29	0.0190916389532006  \\
};

%%% CoR-UTV
\addplot+[smooth,color=blue,loosely dotted, every mark/.append style={solid}, mark=star]
table[row sep=crcr]{
1	1.00058999557122  \\
3	0.999963926442666 \\
5	0.999711241076539 \\
7	0.998985911952049 \\
9	0.998207382489931 \\
11	0.997065911525804 \\
13	0.996505451235520 \\
15	0.995059001146945 \\
17	0.992911866277017 \\
19	0.990262839556902 \\
21	0.0192961172442404 \\
23	0.0192347873643400 \\
25	0.0192119834111648 \\
27	0.0191420840607589 \\
29	0.0190826681270129 \\
};

%%% PbP-QLP
\addplot+[smooth,color=green,loosely dotted, every mark/.append style={solid}, mark=triangle]
table[row sep=crcr]{
1	1.00063078368498  \\
3	1.00002754704790 \\
5	0.999388180354851 \\
7	0.998904567497390 \\
9	0.997916705586803 \\
11	0.997132170346866 \\
13	0.995916779749176 \\
15	0.994378711350089 \\
17	0.993255776086086 \\
19	0.990102622014336 \\
21	0.0193123185181832 \\
23	0.0192491348023251 \\
25	0.0192181297862450 \\
27	0.0191166073030227 \\
29	0.0191092473925143  \\
};

\end{axis}

\end{tikzpicture}% 
	\captionsetup{justification=centering,font=scriptsize}
		\caption{Low-rank approximation errors for Matrix 1 (\texttt{LowRankSmallGap}). Left: basic PbP-QLP. Right: PI-coupled PbP-QLP with $q=2$.}
		\label{fig_LRAE_M1}     % Give a unique label
	\end{center}
\end{figure}

\begin{figure}[t]
	\begin{center}
			% This file was created by matlab2tikz v0.4.7 running on MATLAB 8.3.
\usetikzlibrary{positioning,calc}

\definecolor{mycolor1}{rgb}{0.00000,1.00000,1.00000}%
\definecolor{mycolor2}{rgb}{1.00000,0.00000,1.00000}%

\pgfplotsset{every axis label/.append style={font=\footnotesize},
every tick label/.append style={font=\footnotesize}
}

\begin{tikzpicture}[font=\footnotesize] 

\begin{axis}[%
name=ber,
ymode=log,
width  = 0.35\columnwidth,%5.63489583333333in,
height = 0.35\columnwidth,%4.16838541666667in,
scale only axis,
xmin  = 1,
xmax  = 100,
xlabel= {$d$},
xmajorgrids,
ymin = 9.5e-05 ,
ymax = 0.51,
xtick       ={20,50,80},
xticklabels ={$20$,$50$, $80$},
ylabel={$\|{\bf A}- {\bf A}_{\text{approx}}\|_2$},
ymajorgrids,
legend entries = {SVD,CPQR,p-QLP, R-SVD,CoR-UTV,PbP-QLP},
legend style={at={(1,1)},anchor=north east,draw=black,fill=white,legend cell align=left,font=\tiny},
]

%% SVD 
\addplot+[smooth,color=black,loosely dotted, every mark/.append style={solid}, mark=x]
table[row sep=crcr]{
1	0.250000000000000  \\
10	0.00826446280991735 \\
19	0.00250000000000000 \\
28	0.00118906064209275 \\
37	0.000692520775623268 \\
46	0.000452693526482572 \\
55	0.000318877551020410 \\
64	0.000236686390532544 \\
73	0.000182615047479913 \\
82	0.000145158949049209 \\
91	0.000118147448015123 \\
100	9.80296049406922e-05 \\
};
%% QRP
\addplot+[smooth,color=gray,loosely dotted, every mark/.append style={solid}, mark=+]
table[row sep=crcr]{
1	0.275297093634743  \\
10	0.0119908439883389 \\
19	0.00418149664931698 \\
28	0.00205275548803517 \\
37	0.00141316810230250 \\
46	0.000872199060097471 \\
55	0.000625147511290899 \\
64	0.000526201829317107 \\
73	0.000394464222591332 \\
82	0.000310229059573695 \\
91	0.000256160483947917 \\
100	0.000237836845282966 \\
};

%% p-QLP
\addplot+[smooth,color=red,loosely dotted, every mark/.append style={solid}, mark=pentagon]
table[row sep=crcr]{
1	0.251672111757427  \\
10	0.0101163846306846 \\
19	0.00275921281792495 \\
28	0.00150931517371248 \\
37	0.000874628328144282 \\
46	0.000556547321730776 \\
55	0.000418297426770026 \\
64	0.000329243038450481 \\
73	0.000230293963866043 \\
82	0.000196059352233120 \\
91	0.000152818592512397 \\
100	0.000134264036705895 \\
  };

%%% R-SVD
\addplot+[smooth,color=teal,loosely dotted, every mark/.append style={solid}, mark=diamond]
table[row sep=crcr]{
1	0.464989995688163 \\
10	0.0231992223942671 \\
19	0.00715256291696953 \\
28	0.00381801592783104 \\
37	0.00239755656639599 \\
46	0.00163570548018619 \\
55	0.00115702914866237 \\
64	0.000804860387504065 \\
73	0.000632327160728511 \\
82	0.000520799387054417 \\
91	0.000423163134047128 \\
100	0.000344600229044653 \\
};

%%% CoR-UTV
\addplot+[smooth,color=blue,loosely dotted, every mark/.append style={solid}, mark=star]
table[row sep=crcr]{
1	0.431509229855940  \\
10	0.0283585573905169 \\
19	0.00771081676093407 \\
28	0.00394878950564482 \\
37	0.00229128366243253 \\
46	0.00165651230358545 \\
55	0.00113809515985395 \\
64	0.000809911963566789 \\
73	0.000646221944214623 \\
82	0.000500376435436657 \\
91	0.000418310436665040 \\
100	0.000346047693326696  \\
};

%%% PbP-QLP
\addplot+[smooth,color=green,loosely dotted, every mark/.append style={solid}, mark=triangle]
table[row sep=crcr]{
1	0.500286601830746  \\
10	0.0283944490494524 \\
19	0.00868531578453396 \\
28	0.00397309565453211 \\
37	0.00236721266623928 \\
46	0.00154932664914339 \\
55	0.00108424518728737 \\
64	0.000877210369284242 \\
73	0.000639922383492003 \\
82	0.000514694375014609 \\
91	0.000422181123118470 \\
100	0.000343682912335181 \\
};

\end{axis}

\begin{axis}[%
name=SumRate,
at={($(ber.east)+(35,0em)$)},
		anchor= west,
ymode=log,
width  = 0.35\columnwidth,%5.63489583333333in,
height = 0.35\columnwidth,%4.16838541666667in,
scale only axis,
xmin  = 1,
xmax  = 100,
xlabel= {$d$},
xmajorgrids,
ymin = 9.5e-05 ,
ymax = 0.26,
xtick       ={20,50,80},
xticklabels ={$20$,$50$, $80$},
ylabel={},
ymajorgrids,
%title = {\texttt{HighNoiseLowRank}}]
%ytick       ={0.0973831, 0.0973830 , 0.0973829, 0.0973828},
%yticklabels ={$9.73831$, $9.73830$ , $9.73829$, $9.73828$},
%legend entries={SVD,CPQR,p-QLP, R-SVD,CoR-UTV,R-values,PbP-QLP}, 
%legend style={at={(1,0.82)},anchor=north east,draw=black,fill=white,legend cell align=left,font=\tiny, legend columns=2}
]
%% SVD 
\addplot+[smooth,color=black,loosely dotted, every mark/.append style={solid}, mark=x]
table[row sep=crcr]{
1	0.250000000000000  \\
10	0.00826446280991736 \\
19	0.00250000000000000 \\
28	0.00118906064209275 \\
37	0.000692520775623267 \\
46	0.000452693526482571 \\
55	0.000318877551020409 \\
64	0.000236686390532544 \\
73	0.000182615047479912 \\
82	0.000145158949049209 \\
91	0.000118147448015123 \\
100	9.80296049406915e-05 \\
};
%% QRP
\addplot+[smooth,color=gray,loosely dotted, every mark/.append style={solid}, mark=+]
table[row sep=crcr]{
1	0.257254524101364  \\
10	0.0112002084648865 \\
19	0.00432834910971239 \\
28	0.00209964270144735 \\
37	0.00132798489953713 \\
46	0.000934559771538730 \\
55	0.000680010473254488 \\
64	0.000479999583008566 \\
73	0.000380688887369375 \\
82	0.000321690594475772 \\
91	0.000262345374536544 \\
100	0.000242998920650419 \\
};

%% p-QLP
\addplot+[smooth,color=red,loosely dotted, every mark/.append style={solid}, mark=pentagon]
table[row sep=crcr]{
1	0.250446424381605  \\
10	0.00999812589135079 \\
19	0.00273119844804561 \\
28	0.00139224856639113 \\
37	0.000845216172831954 \\
46	0.000583310361366650 \\
55	0.000395590847506074 \\
64	0.000325030997514352 \\
73	0.000236637808522179 \\
82	0.000190657723643803 \\
91	0.000162847139491166 \\
100	0.000133228264020756 \\
};

%%% R-SVD
\addplot+[smooth,color=teal,loosely dotted, every mark/.append style={solid}, mark=diamond]
table[row sep=crcr]{
1	0.250004512292909  \\
10	0.00893388776064934 \\
19	0.00267989449257045 \\
28	0.00133997402531261 \\
37	0.000795458824544106 \\
46	0.000521261393352945 \\
55	0.000366386440294138 \\
64	0.000266487171614199 \\
73	0.000213592720057495 \\
82	0.000168704871215453 \\
91	0.000135702035340741 \\
100	0.000113945592325572 \\
};

%%% CoR-UTV
\addplot+[smooth,color=blue,loosely dotted, every mark/.append style={solid}, mark=star]
table[row sep=crcr]{
1	0.250002406081024  \\
10	0.00899441518056383 \\
19	0.00287723114317497 \\
28	0.00135722358284567 \\
37	0.000800468875595704 \\
46	0.000520328794437049 \\
55	0.000364741586520966 \\
64	0.000267833381688544 \\
73	0.000211398821589178 \\
82	0.000170933676626514 \\
91	0.000135231529829450 \\
100	0.000114460336904841 \\
};

%%% PbP-QLP
\addplot+[smooth,color=green,loosely dotted, every mark/.append style={solid}, mark=triangle]
table[row sep=crcr]{
1	0.250017404472443  \\
10	0.00895433734303437 \\
19	0.00278820129263587 \\
28	0.00130833296209100 \\
37	0.000794897220200099 \\
46	0.000527858742584453 \\
55	0.000371669519560640 \\
64	0.000274474707872770 \\
73	0.000210127605203899 \\
82	0.000167447898799825 \\
91	0.000135852568032571 \\
100	0.000114445894796163 \\
};

\end{axis}

\end{tikzpicture}% 
	\captionsetup{justification=centering,font=scriptsize}
		\caption{Low-rank approximation errors for Matrix 4. Left: basic PbP-QLP. Right: PI-coupled PbP-QLP with $q=2$.}
		\label{fig_LRAE_M4}       % Give a unique label
	\end{center}
\end{figure}
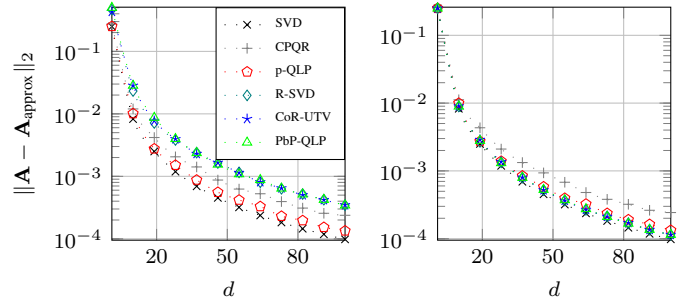

Here, ${\bf A}_\text{Approx}$ is an approximation constructed by each algorithm. The results are shown in Figs. \ref{fig_LRAE_M1}-\ref{fig_LRAE_M4}. We make two observations:
\begin{enumerate}
 \item For Matrix 1, approximations by basic PbP-QLP for $1\le d \le 20$ are highly accurate, while for $d>20$, the approximations, similar to those of R-SVD and CoR-UTV, are poorer than the CPQR, p-QLP, and SVD. However, PI-coupled PbP-QLP with $q=2$ produces approximations as accurate as p-QLP and the SVD for any rank parameter $d$.
 \item For Matrices 3 and 4 CPQR shows a better performance compared with R-SVD, CoR-UTV and PbP-QLP methods with no PI. While PI-coupled PbP-QLP with $q=2$ constructs approximations as good as the optimal SVD.
 \end{enumerate}

\begin{figure}[t]
	\begin{center}
			% This file was created by matlab2tikz v0.4.7 running on MATLAB 8.3.
\usetikzlibrary{positioning,calc}

\definecolor{mycolor1}{rgb}{0.00000,1.00000,1.00000}%
\definecolor{mycolor2}{rgb}{1.00000,0.00000,1.00000}%

\pgfplotsset{every axis label/.append style={font=\footnotesize},
every tick label/.append style={font=\footnotesize}
}

\begin{tikzpicture}[font=\footnotesize] 

\begin{axis}[%
name=ber,
ymode=log,
width  = 0.35\columnwidth,%5.63489583333333in,
height = 0.35\columnwidth,%4.16838541666667in,
scale only axis,
xmin  = 1,
xmax  = 100,
xlabel= {$d$},
xmajorgrids,
ymin = 4.8e-08 ,
ymax = 0.81,
xtick       ={20,50,80},
xticklabels ={$20$,$50$, $80$},
ylabel={$\|{\bf A}- {\bf A}_{\text{approx}}\|_2$},
ymajorgrids,
]

%% SVD 
\addplot+[smooth,color=black,loosely dotted, every mark/.append style={solid}, mark=x]
table[row sep=crcr]{
1	0.716531310573789  \\
10	0.159879746079694  \\
19	0.0356739933472524 \\
28	0.00795994384870645 \\
37	0.00177610354573438 \\
46	0.000396302268599906 \\
55	8.84269886598842e-05 \\
64	1.97307281411224e-05 \\
73	4.40252052997505e-06 \\
82	9.82335110908031e-07 \\
91	2.19188590617503e-07 \\
100	4.89075853280066e-08 \\
};
%% QRP
\addplot+[smooth,color=gray,loosely dotted, every mark/.append style={solid}, mark=+]
table[row sep=crcr]{
1	0.723432302758466  \\
10	0.225294368552873 \\
19	0.0508731965230759 \\
28	0.0132904724218676 \\
37	0.00353648044736736 \\
46	0.000676163868368664 \\
55	0.000186322309934491 \\
64	5.16598530056625e-05 \\ 
73	1.11168997619283e-05 \\
82	2.86583728937964e-06 \\
91	7.26457915621270e-07 \\
100	1.41151426169204e-07 \\
};

%% p-QLP
\addplot+[smooth,color=red,loosely dotted, every mark/.append style={solid}, mark=pentagon]
table[row sep=crcr]{
1	0.721267162452223  \\
10	0.177518498574665 \\
19	0.0365398682042816 \\
28	0.00831905502556898 \\
37	0.00209909166765973 \\
46	0.000426855750644204 \\
55	0.000101056012374994 \\
64	2.18287017309652e-05 \\
73	4.98882079657525e-06 \\
82	1.12408243321253e-06 \\
91	2.59860834696290e-07 \\
100	5.26192732284805e-08 \\
  };

%%% R-SVD
\addplot+[smooth,color=teal,loosely dotted, every mark/.append style={solid}, mark=diamond]
table[row sep=crcr]{
1	0.784104618338314  \\
10	0.348728469127229 \\
19	0.121702773013965 \\
28	0.0296358536082815 \\
37	0.00903698770155054 \\
46	0.00209199382810149 \\
55	0.000581644553106194 \\
64	0.000117402250818065 \\
73	3.06017893095629e-05 \\
82	7.58380206120062e-06 \\
91	1.48581676693754e-06 \\
100	3.90410337351319e-07 \\
};

%%% CoR-UTV
\addplot+[smooth,color=blue,loosely dotted, every mark/.append style={solid}, mark=star]
table[row sep=crcr]{
1	0.801169883481126  \\
10	0.383178154701134 \\
19	0.110631965018738 \\
28	0.0351505557381136 \\
37	0.00841210451042115 \\
46	0.00180582853812589 \\
55	0.000529200048841984 \\
64	0.000125134805541480 \\
73	2.85839226123951e-05 \\
82	7.16956042544369e-06 \\
91	1.96430611651256e-06 \\
100	4.00741724028489e-07 \\
};

%%% PbP-QLP
\addplot+[smooth,color=green,loosely dotted, every mark/.append style={solid}, mark=triangle]
table[row sep=crcr]{
1	0.792081049235836 \\
10	0.360362616506266 \\
19	0.110639502075947 \\
28	0.0335897008341522 \\
37	0.00839333787026068 \\
46	0.00205391568117148 \\
55	0.000647776692501045 \\
64	0.000122645692654966 \\
73	3.04040600727926e-05 \\
82	6.93667872344955e-06 \\
91	1.58124056408724e-06 \\
100	3.57337923343843e-07 \\
};

\end{axis}

\begin{axis}[%
name=SumRate,
at={($(ber.east)+(35,0em)$)},
		anchor= west,
ymode=log,
width  = 0.35\columnwidth,%5.63489583333333in,
height = 0.35\columnwidth,%4.16838541666667in,
scale only axis,
xmin  = 1,
xmax  = 100,
xlabel= {$d$},
xmajorgrids,
ymin = 4.8e-08 ,
ymax = 0.75,
xtick       ={20,50,80},
xticklabels ={$20$,$50$, $80$},
ylabel={},
ymajorgrids,
%title = {\texttt{HighNoiseLowRank}}]
%ytick       ={0.0973831, 0.0973830 , 0.0973829, 0.0973828},
%yticklabels ={$9.73831$, $9.73830$ , $9.73829$, $9.73828$},
%legend entries={SVD,CPQR,p-QLP, R-SVD,CoR-UTV,R-values,PbP-QLP}, 
%legend style={at={(1,0.82)},anchor=north east,draw=black,fill=white,legend cell align=left,font=\tiny, legend columns=2}
]
%% SVD 
\addplot+[smooth,color=black,loosely dotted, every mark/.append style={solid}, mark=x]
table[row sep=crcr]{
1	0.716531310573790  \\
10	0.159879746079694 \\
19	0.0356739933472524 \\
28	0.00795994384870645 \\
37	0.00177610354573438 \\
46	0.000396302268599905 \\ 
55	8.84269886598835e-05 \\ 
64	1.97307281411237e-05 \\
73	4.40252052997380e-06 \\
82	9.82335110909691e-07 \\
91	2.19188590617607e-07 \\
100	4.89075853275695e-08 \\
};
%% QRP
\addplot+[smooth,color=gray,loosely dotted, every mark/.append style={solid}, mark=+]
table[row sep=crcr]{
1	0.721836547841102  \\
10	0.197938296415483 \\
19	0.0583056491884956 \\
28	0.0125371758736916 \\
37	0.00362881127813515 \\
46	0.000770808887359466 \\
55	0.000217210319465723 \\
64	4.64901776305304e-05 \\
73	9.18733878112628e-06 \\
82	2.01489099634121e-06 \\
91	6.29647217273430e-07 \\
100	1.47644753155220e-07 \\
};

%% p-QLP
\addplot+[smooth,color=red,loosely dotted, every mark/.append style={solid}, mark=pentagon]
table[row sep=crcr]{
1	0.719939319556081  \\
10	0.171526678940468 \\
19	0.0395737500697207 \\
28	0.00899812937974519 \\
37	0.00192744484485872 \\
46	0.000416406240703834 \\
55	0.000100327982535088 \\
64	2.52393949478037e-05 \\
73	5.13720938944981e-06 \\
82	1.07189031839916e-06 \\
91	2.55415380173766e-07 \\
100	5.78290549669894e-08 \\
};

%%% R-SVD
\addplot+[smooth,color=teal,loosely dotted, every mark/.append style={solid}, mark=diamond]
table[row sep=crcr]{
1	0.728252096642333  \\
10	0.172674767318547 \\
19	0.0378295203668173 \\
28	0.00829920573399832 \\
37	0.00184364490253606 \\
46	0.000438335363710129 \\
55	0.000103871610908949 \\
64	2.26390832530141e-05 \\
73	4.98001642613320e-06 \\
82	1.08848264048088e-06 \\
91	2.44071134845848e-07 \\
100	5.42017889278780e-08 \\
};

%%% CoR-UTV
\addplot+[smooth,color=blue,loosely dotted, every mark/.append style={solid}, mark=star]
table[row sep=crcr]{
1	0.741238724638056  \\
10	0.174134659379740 \\
19	0.0380702900509528 \\
28	0.00892298438278627 \\
37	0.00196970559997219 \\
46	0.000468076930836910 \\
55	9.68653945435874e-05 \\
64	2.14647865364679e-05 \\
73	4.84676917743741e-06 \\
82	1.10356927203913e-06 \\
91	2.54337406052496e-07 \\
100	5.55079237202811e-08 \\
};

%%% PbP-QLP
\addplot+[smooth,color=green,loosely dotted, every mark/.append style={solid}, mark=triangle]
table[row sep=crcr]{
1	0.749895367722576  \\
10	0.176155553193464 \\
19	0.0385575874847150 \\
28	0.00889101509252599 \\
37	0.00200807957374070 \\
46	0.000436143681349587 \\
55	9.51201853730410e-05 \\
64	2.15378156483823e-05 \\
73	4.70449093773689e-06 \\
82	1.08399772752443e-06 \\
91	2.51692546472721e-07 \\
100	5.49441995442744e-08 \\
};

\end{axis}

\end{tikzpicture}% 
	\captionsetup{justification=centering,font=scriptsize}
		\caption{Low-rank approximation errors for Matrix 3. Left: basic PbP-QLP. Right: PI-coupled PbP-QLP with $q=2$.}
		\label{fig_LRAE_M3} 
	\end{center}
\end{figure}

\subsection{Low-Rank Image Reconstruction}
\begin{figure}[!htb]
	\centering
\includegraphics[width=0.42\textwidth,height=15cm]{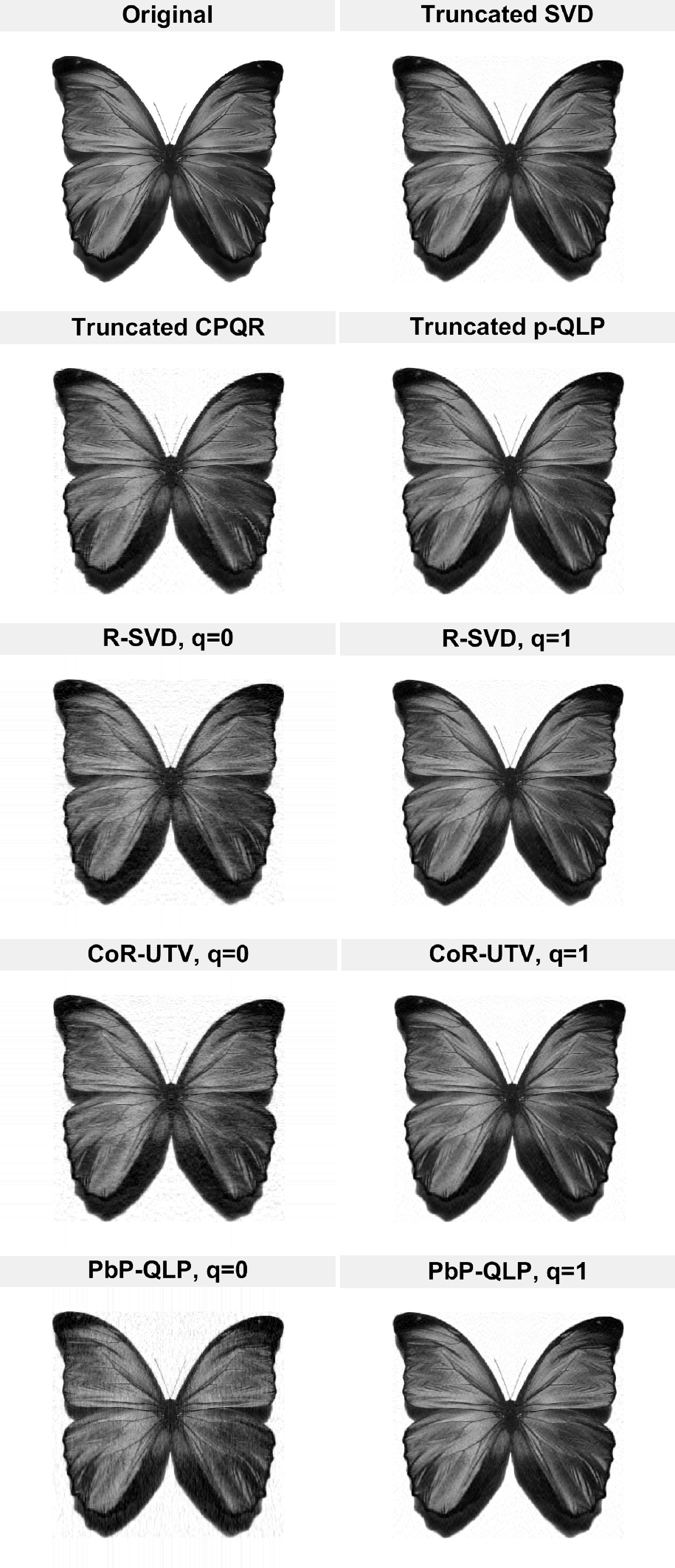}
	\caption{Low-rank image reconstruction. This figure shows the results of reconstructing a butterfly image with dimension $969\times 812$ using $rank=80$ by different methods.}
	\label{fig:Butterfly}
\end{figure}

This experiment investigates the performance of PbP-QLP on real-world data, where we reconstruct a gray-scale low-rank image of a butterfly of size $969\times 812$. We compare the results of PbP-QLP against those produced by the (optimal) truncated SVD, truncated CPQR and p-QLP, R-SVD, and CoR-UTV. The reconstructions with $rank = 80$ are displayed in Fig. \ref{fig:Butterfly}. We observe, with close scrutiny, that (i) the reconstructions by CPQR and the three randomized methods with no PI contains more noise than those of other methods, (ii) the reconstruction by CPQR, in addition to adding noise to the image, slightly distorts
butterfly’s antennas and upper parts of forewings (close to the head), and (iii) the approximation by PI-coupled PbP-QLP with $q = 1$ is as good as those of the p-QLP and SVD.

\begin{figure}[t]  
	\begin{center} 
		% This file was created by matlab2tikz v0.4.7 running on MATLAB 8.3.
% Copyright (c) 2008--2014, Nico Schlömer <nico.schloemer@gmail.com>
% All rights reserved.
% Minimal pgfplots version: 1.3
% 
% The latest updates can be retrieved from
%   http://www.mathworks.com/matlabcentral/fileexchange/22022-matlab2tikz
% where you can also make suggestions and rate matlab2tikz.
% 
%
% defining custom colors
\usetikzlibrary{positioning,calc}

\definecolor{mycolor1}{rgb}{0.00000,1.00000,1.00000}%
\definecolor{mycolor2}{rgb}{1.00000,0.00000,1.00000}%

\pgfplotsset{every axis label/.append style={font=\footnotesize},
every tick label/.append style={font=\footnotesize}
}

\begin{tikzpicture}[font=\footnotesize] 

\begin{axis}[%
name=ber,
ymode=log,
width  = 0.85\columnwidth,%5.63489583333333in,
height = 0.6\columnwidth,%4.16838541666667in,
scale only axis,
xmin  = 10,
xmax  = 197,
xlabel= {Approximation rank},
xmajorgrids,
xtick       ={40,100,160},
xticklabels ={$40$, $100$, $160$},
ymin = 5.55,
ymax = 116,
ylabel={$\|{\bf A}- {\bf A}_{\text{approx}}\|_F$},
ymajorgrids,
legend entries={Truncated SVD, Truncated CPQR, Truncated p-QLP, R-SVD $q=0$, R-SVD $q=1$, R-SVD $q=2$, 
	CoR-UTV $q=0$, CoR-UTV $q=1$, CoR-UTV $q=2$, PbP-QLP $q=0$, PbP-QLP $q=1$, PbP-QLP $q=2$},
%legend pos=outer north east
legend style={at={(0.62,1)},anchor=north,legend columns=2}
]

%% Truncaed SVD
\addplot+[smooth,color=black,loosely dotted, every mark/.append style={solid}, mark=x]
table[row sep=crcr]
{
10	71.8938652383786 \\
27	40.9796039258958 \\
44	30.4731405830893 \\
61	24.0204232128697 \\
78	19.5137994755066 \\
95	16.0756420291948 \\
112	13.3489616347212 \\
129	11.1522661178150 \\
146	9.34795750177788 \\
163	7.86112253249887 \\
180	6.61273455043875 \\
197	5.57526252238800 \\
}; 

%% Trun CPQR
\addplot+[smooth,color=gray, loosely dotted, every mark/.append style={solid}, mark=+]
  table[row sep=crcr]
  {
10	100.594121711605 \\
27	55.7823396287008 \\
44	42.1656166272928 \\
61	33.5976129669168 \\
78	27.9869009962126 \\
95	23.5744360378440 \\
112	19.8763840994694 \\
129	16.7349318416150 \\ 
146	14.4254519559253 \\
163	12.3895568939007 \\
180	10.4847350965635 \\
197	8.98164703573617 \\
}; 

%% Trun pQLP
\addplot+[smooth,color=red, loosely dotted, every mark/.append style={solid}, mark=pentagon]
  table[row sep=crcr]
  {
10	75.4948011136968\\
27	43.6857952396875\\
44	33.2381274456121\\
61	26.3019711787528\\
78	21.6350079289636\\
95	18.0493871828328\\
112	14.9898548891112\\
129	12.5773190070421\\
146	10.6467289642640\\
163	9.02668322306606\\
180	7.57668216510718\\
197	6.37378205822769\\
};

%% R-SVD 0
\addplot+[smooth,color=olive, dashed, every mark/.append style={solid}, mark = diamond]
  table[row sep=crcr]
{
10	111.238376859295\\
27	65.8261947125358\\
44	48.3446298956982\\
61	38.9875817044609\\
78	32.7975466854221\\
95	27.7326279519414\\
112	23.7648188308396\\
129	20.3679590161831\\
146	17.5733026961320\\
163	15.3715553608025\\
180	13.2979409681936\\
197	11.5759139771231 \\
};

%%  R-SVD 1
\addplot+[smooth,color = brown ,densely dotted, every mark/.append style={solid}, mark=diamond]
  table[row sep=crcr]
  {
10	74.5760112174502\\
27	43.8435123367740\\
44	32.6371334927984\\
61	25.8622919423058\\
78	21.0182995175622\\
95	17.2898394723038\\
112	14.4953755440535\\
129	12.1366372022877\\
146	10.2263209322586\\
163	8.60160110865441\\
180	7.25740032769414\\
197	6.12956571030006 \\
};

%%% R-SVD 2 column #7
\addplot+[smooth,color = teal ,loosely dotted, every mark/.append style={solid}, mark=diamond]
  table[row sep=crcr]
  {
10	74.4558829161015\\
27	41.9719712837795\\
44	31.0283783466096\\
61	24.6953669465060\\
78	20.0863429995545\\
95	16.5742446705570\\
112	13.8282391557237\\
129	11.5017679161720\\
146	9.64827061437541\\
163	8.11036069510972\\
180	6.81852825297636\\
197	5.76156406298492\\
};

%%% CoR-UTV 0 column #8
\addplot+[smooth,color = cyan ,dashed, every mark/.append style={solid}, mark=star]
  table[row sep=crcr]
  {
10	108.335569240298 \\
27	66.2905750743245 \\
44	48.8381033655969 \\
61	39.3080146538838 \\
78	32.9553001678587 \\
95	27.6498282906489 \\
112	23.6575383208730 \\
129	20.6317765233135 \\
146	17.6273736602665 \\
163	15.4499867915673 \\
180	13.3800132816276 \\
197	11.5437674290438 \\
};

%%% CoR-UTV 1
\addplot+[smooth,color = lightgray ,densely dotted, every mark/.append style={solid}, mark=star]
  table[row sep=crcr]
  {
10	75.5840490894803 \\
27	43.9498843173018\\
44	32.2877913903037\\
61	25.7728095779801\\
78	20.9548929594057\\
95	17.4556687166385\\
112	14.4850896670130\\
129	12.1390731847907\\
146	10.1861389205655\\
163	8.58265819246936\\
180	7.26144373288884\\
197	6.12828840000730 \\
};

%%% CoR-UTV 2
\addplot+[smooth,color = blue ,loosely dotted, every mark/.append style={solid}, mark=star]
table[row sep=crcr]
{
10	74.2899703144896 \\
27	42.3110873461690 \\
44	31.1934721902288 \\
61	24.5956435411121 \\
78	20.0717221710926 \\
95	16.6112711380008 \\
112	13.7542266549535 \\
129	11.4893732797832 \\
146	9.65277600376355 \\
163	8.09371376112062 \\
180	6.82928318361409 \\
197	5.77112226779693 \\
};

%%% PbP-QLP 0  column #11
\addplot+[smooth,color = pink ,dashed, every mark/.append style={solid}, mark=triangle]
table[row sep=crcr]
{
10	115.504497587907 \\
27	65.6484044613648 \\
44	49.1114055775783 \\
61	39.3813017374808 \\
78	32.9287498464777 \\
95	27.7050014820009 \\
112	23.6565946298749 \\
129	20.2647789065899 \\
146	17.6586877908692 \\
163	15.4367306014817 \\
180	13.1513117190143 \\
197	11.6563263017569 \\
};

%%% PbP-QLP 1
\addplot+[smooth,color = violet ,densely dotted, every mark/.append style={solid}, mark=triangle]
table[row sep=crcr]
{
10	73.7027265628626 \\
27	44.5324795135608 \\
44	32.5108240299238 \\
61	25.6658909745380 \\
78	21.0062819252344 \\
95	17.3849550156056 \\
112	14.4636463813625 \\
129	12.0905291092662 \\
146	10.1609086893895 \\
163	8.58812833278019 \\
180	7.22788906988576 \\
197	6.10041373688828 \\
};
%%% PbP-QLP 2
\addplot+[smooth,color = green, loosely dotted, every mark/.append style={solid}, mark=triangle]
table[row sep=crcr]
{
10	72.6153029350829 \\
27	41.8715189288629 \\
44	31.0722862357949 \\
61	24.6474724422023 \\
78	20.1063343190342 \\
95	16.5164757172376 \\
112	13.7966588816182 \\
129	11.5363498185160 \\
146	9.66956956529741 \\
163	8.09037161951101 \\
180	6.84960867122087 \\
197	5.77670606669432 \\
};

\end{axis}

\end{tikzpicture}%
		\caption{Image reconstruction Frobenius norm approximation error. This figure displays the errors incurred by different methods in reconstructing a butterfly image. Here, ${\bf A}_{\text{approx}}$ is an approximation generated by either method. The approximation by PI-coupled PbP-QLP with $q=2$ shows almost no loss of accuracy compared to the optimal truncated SVD.}
		\label{fig_Butterfly_FroErr}       % Give a unique label
	\end{center}
\end{figure}
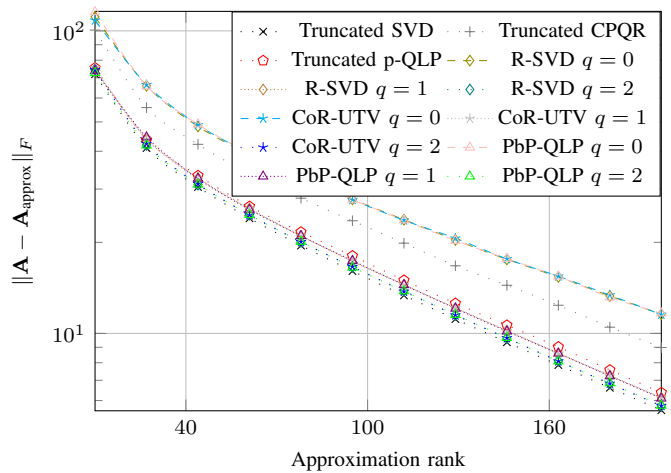

Fig. \ref{fig_Butterfly_FroErr} displays the reconstruction errors in terms of the Frobenius norm for considered methods against the
approximation rank. We observe that (i) randomized methods with no PI demonstrate similar performance and their reconstructions generate more errors, and (ii) PbP-QLP-coupled with $q = 2$ outperforms p-QLP and constructs approximations with almost no loss of accuracy compared to the optimal SVD.

\subsection{Test Matrices from Applications}
In this experiment, we investigate the effectiveness of the PbP-QLP algorithm on five matrices of size $256 \times 256$ described in Table \ref{Table_PracMat} from different applications \cite{Hansen_RegTools}. These matrices have been used in other works, e.g., \cite{DemGGX15, Ayala19}.
\begin{table}[!htb] 
\caption{Test matrices from applications.}
\begin{tabular}
{p{0.08cm} p{1cm} p{6.5cm} }
\noindent\rule{8.4cm}{0.4pt}\\
No. & Matrix & Description  \\
\noindent\rule{8.4cm}{0.4pt}\\
1 & \texttt{Baart} & Discretization of a first-kind Fredholm integral equation. \\
2 & \texttt{Deriv2} & Computation of the second derivative.  \\
3 & \texttt{Foxgood} & Severely ill-posed problem. \\
4 & \texttt{Gravity} & 1D gravity surveying problem. \\
5 & \texttt{Heat} & Inverse heat equation. \\
\noindent\rule{8.4cm}{0.4pt}
\end{tabular}
\label{Table_PracMat}
\end{table}

\subsubsection{Low-Rank Approximation}
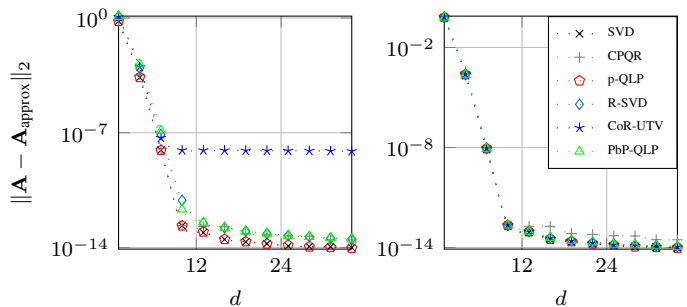
\begin{figure}[t]
	\begin{center}
			% This file was created by matlab2tikz v0.4.7 running on MATLAB 8.3.
\usetikzlibrary{positioning,calc}

\definecolor{mycolor1}{rgb}{0.00000,1.00000,1.00000}%
\definecolor{mycolor2}{rgb}{1.00000,0.00000,1.00000}%

\pgfplotsset{every axis label/.append style={font=\footnotesize},
every tick label/.append style={font=\footnotesize}
}

\begin{tikzpicture}[font=\footnotesize] 

\begin{axis}[%
name=ber,
ymode=log,
width  = 0.35\columnwidth,%5.63489583333333in,
height = 0.35\columnwidth,%4.16838541666667in,
scale only axis,
xmin  = 1,
xmax  = 34,
xlabel= {$d$},
xmajorgrids,
ymin = 8e-15 ,
ymax = 1.3,
xtick       ={12, 24},
xticklabels ={$12$,$24$},
ylabel={$\|{\bf A}- {\bf A}_{\text{approx}}\|_2$},
ymajorgrids,
]

%% SVD 
\addplot+[smooth,color=black,loosely dotted, every mark/.append style={solid}, mark=x]
table[row sep=crcr]{
1	0.631356459810302 \\
4	0.000236619381934864\\
7	8.62678543880795e-09\\
10	2.12926953267786e-13\\
13	8.16051627099115e-14\\
16	3.10198291367570e-14\\
19	2.19683542508400e-14\\
22	1.71071512723568e-14\\
25	1.23420393990765e-14\\
28	1.15144497329520e-14\\
31	1.12875795613212e-14\\
34	9.93268312214190e-15 \\
};
%% QRP
\addplot+[smooth,color=gray,loosely dotted, every mark/.append style={solid}, mark=+]
table[row sep=crcr]{
1	0.753006098070034\\
4	0.000341279865027706 \\
7	1.03324690787861e-08\\
10	2.16670431109900e-13\\
13	1.87940866858867e-13\\
16	1.86718330941574e-13\\
19	7.01066651848633e-14\\
22	6.01627461423962e-14\\
25	5.09817786512144e-14\\
28	4.86157036287819e-14\\
31	2.95072575486788e-14\\
34	2.89264436231876e-14 \\
};

%% p-QLP
\addplot+[smooth,color=red,loosely dotted, every mark/.append style={solid}, mark=pentagon]
table[row sep=crcr]{
1	0.636492651551068\\
4	0.000236713064089796\\
7	8.62701613903237e-09\\
10	2.12929387270626e-13\\
13	8.99318267187842e-14\\
16	3.14868627445912e-14\\
19	2.29108536793438e-14\\
22	1.64594260071232e-14\\
25	1.38688731858148e-14\\
28	1.10775756524197e-14\\
31	1.02835350091839e-14\\
34	9.25031337051791e-15 \\
  };

%%% R-SVD
\addplot+[smooth,color=teal,loosely dotted, every mark/.append style={solid}, mark=diamond]
table[row sep=crcr]{
1	1.29650400173732\\
4	0.000867743035424275\\
7	7.67689850158600e-08\\
10	7.54643938673490e-12\\
13	3.09392575795475e-13\\
16	1.60600957295090e-13\\
19	1.01092215609695e-13\\
22	7.19056222954474e-14\\
25	5.39586789974272e-14\\
28	4.55975248588149e-14\\
31	3.77482149345396e-14\\
34	3.36025572330687e-14 \\
};

%%% CoR-UTV
\addplot+[smooth,color=blue,loosely dotted, every mark/.append style={solid}, mark=star]
table[row sep=crcr]{
1	1.14048954767639\\
4	0.00103810202810154\\
7	4.89523287329943e-08\\
10	8.53665292544563e-09\\
13	8.40895987175296e-09\\
16	8.29260317918944e-09\\
19	8.13978661773754e-09\\
22	7.96802040791937e-09\\
25	7.97227014839113e-09\\
28	7.88486173882343e-09\\
31	7.64874653467647e-09\\
34	7.54639864019000e-09 \\
};

%%% PbP-QLP
\addplot+[smooth,color=green,loosely dotted, every mark/.append style={solid}, mark=triangle]
table[row sep=crcr]{
1	1.19439259718410\\
4	0.00152225684889984\\
7	1.34390441941530e-07\\
10	2.12325055368588e-12\\
13	3.39494073963284e-13\\
16	1.62838269508548e-13\\
19	9.49677454575297e-14\\
22	7.12747102473155e-14\\
25	5.28669440951402e-14\\
28	4.51743220883812e-14\\
31	3.85621691643848e-14\\
34	3.12924450072608e-14 \\
};

\end{axis}

\begin{axis}[%
name=SumRate,
at={($(ber.east)+(35,0em)$)},
		anchor= west,
ymode=log,
width  = 0.35\columnwidth,%5.63489583333333in,
height = 0.35\columnwidth,%4.16838541666667in,
scale only axis,
xmin  = 1,
xmax  = 34,
xlabel= {$d$},
xmajorgrids,
ymin = 8e-15 ,
ymax = 0.76,
xtick       = {12, 24},
xticklabels = {$12$,$24$},
ylabel={},
ymajorgrids,
%title = {\texttt{HighNoiseLowRank}}]
%ytick       ={0.0973831, 0.0973830 , 0.0973829, 0.0973828},
%yticklabels ={$9.73831$, $9.73830$ , $9.73829$, $9.73828$},
legend entries = {SVD,CPQR,p-QLP, R-SVD,CoR-UTV,PbP-QLP},
legend style={at={(1,1)},anchor=north east,draw=black,fill=white,legend cell align=left,font=\tiny}
]
%% SVD 
\addplot+[smooth,color=black,loosely dotted, every mark/.append style={solid}, mark=x]
table[row sep=crcr]{
1	0.631356459810302 \\
4	0.000236619381934864 \\
7	8.62678543880795e-09 \\
10	2.12926953267786e-13 \\
13	8.16051627099115e-14 \\
16	3.10198291367570e-14 \\
19	2.19683542508400e-14 \\
22	1.71071512723568e-14 \\
25	1.23420393990765e-14 \\
28	1.15144497329520e-14 \\
31	1.12875795613212e-14 \\
34	9.93268312214190e-15 \\
};
%% QRP
\addplot+[smooth,color=gray,loosely dotted, every mark/.append style={solid}, mark=+]
table[row sep=crcr]{
1	0.753006098070034 \\
4	0.000341279865027706 \\
7	1.03324690787861e-08 \\
10	2.16670431109900e-13 \\
13	1.87940866858867e-13 \\
16	1.86718330941574e-13 \\
19	7.01066651848633e-14 \\
22	6.01627461423962e-14 \\
25	5.09817786512144e-14 \\
28	4.86157036287819e-14 \\
31	2.95072575486788e-14 \\
34	2.89264436231876e-14 \\
};

%% p-QLP
\addplot+[smooth,color=red,loosely dotted, every mark/.append style={solid}, mark=pentagon]
table[row sep=crcr]{
1	0.636492651551068 \\
4	0.000236713064089796 \\
7	8.62701613903237e-09 \\
10	2.12929387270626e-13 \\
13	8.99318267187842e-14 \\
16	3.14868627445912e-14 \\
19	2.29108536793438e-14 \\
22	1.64594260071232e-14 \\
25	1.38688731858148e-14 \\
28	1.10775756524197e-14 \\
31	1.02835350091839e-14 \\
34	9.25031337051791e-15 \\
};

%%% R-SVD
\addplot+[smooth,color=teal,loosely dotted, every mark/.append style={solid}, mark=diamond]
table[row sep=crcr]{
1	0.631415830523519 \\
4	0.000236619383072629 \\
7	8.62678543557942e-09 \\
10	2.13450227713003e-13 \\
13	9.03259182571465e-14 \\
16	3.91910384268545e-14 \\
19	2.62703018811268e-14 \\
22	2.04684275430878e-14 \\
25	1.61795737259291e-14 \\
28	1.44358790544462e-14 \\
31	1.28072824648448e-14 \\
34	1.17224781465026e-14 \\
};

%%% CoR-UTV
\addplot+[smooth,color=blue,loosely dotted, every mark/.append style={solid}, mark=star]
table[row sep=crcr]{
1	0.631397195164583 \\
4	0.000236619382005766 \\
7	8.62678543479908e-09 \\
10	2.12926476492478e-13 \\
13	8.64803839954116e-14 \\
16	3.43602597508547e-14 \\
19	2.41600909033317e-14 \\
22	1.78621222611096e-14 \\
25	1.43560433848951e-14 \\
28	1.23246816785577e-14 \\
31	1.07975498826773e-14 \\
34	1.03929737149603e-14 \\
};

%%% PbP-QLP
\addplot+[smooth,color=green,loosely dotted, every mark/.append style={solid}, mark=triangle]
table[row sep=crcr]{
1	0.631401419480331 \\
4	0.000236619382187680 \\
7	8.62678543843367e-09 \\
10	2.12929396551456e-13 \\
13	9.02133885500544e-14 \\
16	3.82212739961993e-14 \\
19	2.56139468496872e-14 \\
22	2.01609424585018e-14 \\
25	1.51225157173219e-14 \\
28	1.37880891706407e-14 \\
31	1.20424631175211e-14 \\
34	1.08586729432261e-14 \\
};

\end{axis}

\end{tikzpicture}%
	\captionsetup{justification=centering,font=scriptsize}
		\caption{Low-rank approximation errors for \texttt{Baart}. Left: basic PbP-QLP. Right: PI-coupled PbP-QLP with $q=2$.}
		\label{fig_Baart}       % Give a unique label
	\end{center}
\end{figure}

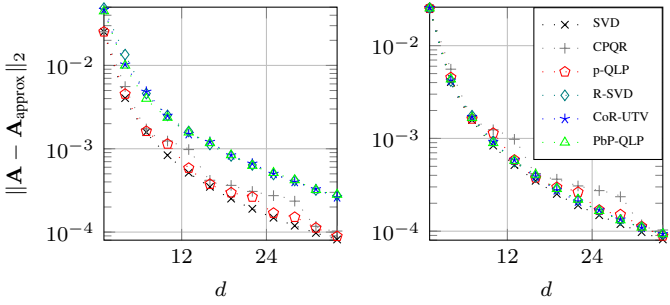
\begin{figure}[t]
	\begin{center}
			% This file was created by matlab2tikz v0.4.7 running on MATLAB 8.3.
\usetikzlibrary{positioning,calc}

\definecolor{mycolor1}{rgb}{0.00000,1.00000,1.00000}%
\definecolor{mycolor2}{rgb}{1.00000,0.00000,1.00000}%

\pgfplotsset{every axis label/.append style={font=\footnotesize},
every tick label/.append style={font=\footnotesize}
}

\begin{tikzpicture}[font=\footnotesize] 

\begin{axis}[%
name=ber,
ymode=log,
width  = 0.35\columnwidth,%5.63489583333333in,
height = 0.35\columnwidth,%4.16838541666667in,
scale only axis,
xmin  = 1,
xmax  = 34,
xlabel= {$d$},
xmajorgrids,
ymin = 8e-05 ,
ymax = 0.05,
xtick       ={12, 24},
xticklabels ={$12$,$24$},
ylabel={$\|{\bf A}- {\bf A}_{\text{approx}}\|_2$},
ymajorgrids,
]

%% SVD 
\addplot+[smooth,color=black,loosely dotted, every mark/.append style={solid}, mark=x]
table[row sep=crcr]{
1	0.0253290243831293 \\
4	0.00405157601934348 \\
7	0.00158187254167646 \\
10	0.000836094747699941 \\
13	0.000515675127603134 \\
16	0.000349323537083890 \\
19	0.000252035232435753 \\
22	0.000190266943794217 \\
25	0.000148618340790251 \\
28	0.000119213558961898 \\
31	9.76847675415860e-05 \\
34	8.14514200897467e-05 \\
};
%% QRP
\addplot+[smooth,color=gray,loosely dotted, every mark/.append style={solid}, mark=+]
table[row sep=crcr]{
1	0.0253314610712142 \\
4	0.00557707496060341 \\
7	0.00172880703648286 \\
10	0.00123878192984628 \\
13	0.000978227192515464 \\
16	0.000418504548886555 \\
19	0.000363417542451331 \\
22	0.000307454890517113 \\
25	0.000273791171134780 \\ 
28	0.000235180611385680 \\
31	0.000117257068803280 \\
34	0.000102932364418764 \\
};

%% p-QLP
\addplot+[smooth,color=red,loosely dotted, every mark/.append style={solid}, mark=pentagon]
table[row sep=crcr]{
1	0.0253291364504947 \\
4	0.00453695451002515 \\
7	0.00162278424610541 \\
10	0.00114228127275031 \\
13	0.000586402837939474 \\
16	0.000376264658868274 \\
19	0.000297394920744826 \\
22	0.000263721671171032 \\ 
25	0.000169179512300632 \\
28	0.000151056460408195 \\
31	0.000112326767566525 \\
34	8.90472755751952e-05 \\
  };

%%% R-SVD
\addplot+[smooth,color=teal,loosely dotted, every mark/.append style={solid}, mark=diamond]
table[row sep=crcr]{
1	0.0487413797543234 \\
4	0.0134744333048791 \\
7	0.00460320054872351 \\
10	0.00251700724849482 \\
13	0.00161265115000105 \\
16	0.00112555166132598 \\
19	0.000819704363268365 \\
22	0.000638871226324382 \\
25	0.000491457531600057 \\
28	0.000415982515169358 \\
31	0.000319830873394427 \\
34	0.000275140799645858 \\
};

%%% CoR-UTV
\addplot+[smooth,color=blue,loosely dotted, every mark/.append style={solid}, mark=star]
table[row sep=crcr]{
1	0.0465377534833647 \\
4	0.0102665301374834 \\
7	0.00484332027974424 \\
10	0.00249629078132136 \\
13	0.00150133214007614 \\
16	0.00120933657378883 \\
19	0.000835202955795869 \\
22	0.000664408347501932 \\
25	0.000501178795375682 \\
28	0.000400471226794863 \\
31	0.000327394025050551 \\
34	0.000261807935786718 \\ 
};

%%% PbP-QLP
\addplot+[smooth,color=green,loosely dotted, every mark/.append style={solid}, mark=triangle]
table[row sep=crcr]{
1	0.0447530925997986 \\
4	0.00999622739243675 \\
7	0.00399618677651289 \\
10	0.00236335050225437 \\
13	0.00157448004391229 \\
16	0.00117266207769932 \\
19	0.000838897486890489 \\
22	0.000626149127117550 \\
25	0.000514354682564742 \\
28	0.000420979973478752 \\
31	0.000320994661540461 \\
34	0.000281909400330113 \\
};

\end{axis}

\begin{axis}[%
name=SumRate,
at={($(ber.east)+(35,0em)$)},
		anchor= west,
ymode=log,
width  = 0.35\columnwidth,%5.63489583333333in,
height = 0.35\columnwidth,%4.16838541666667in,
scale only axis,
xmin  = 1,
xmax  = 34,
xlabel= {$d$},
xmajorgrids,
ymin = 8e-5 ,
ymax = 0.026,
xtick       = {12, 24},
xticklabels = {$12$,$24$},
ylabel={},
ymajorgrids,
%title = {\texttt{HighNoiseLowRank}}]
%ytick       ={0.0973831, 0.0973830 , 0.0973829, 0.0973828},
%yticklabels ={$9.73831$, $9.73830$ , $9.73829$, $9.73828$},
legend entries = {SVD,CPQR,p-QLP, R-SVD,CoR-UTV,PbP-QLP},
legend style={at={(1,1)},anchor=north east,draw=black,fill=white,legend cell align=left,font=\tiny}
]
%% SVD 
\addplot+[smooth,color=black,loosely dotted, every mark/.append style={solid}, mark=x]
table[row sep=crcr]{
1	0.0253290243831293 \\
4	0.00405157601934348 \\
7	0.00158187254167646 \\
10	0.000836094747699941 \\
13	0.000515675127603134 \\
16	0.000349323537083890 \\ 
19	0.000252035232435753 \\
22	0.000190266943794217 \\
25	0.000148618340790251 \\
28	0.000119213558961898 \\
31	9.76847675415860e-05 \\
34	8.14514200897467e-05 \\
};
%% QRP
\addplot+[smooth,color=gray,loosely dotted, every mark/.append style={solid}, mark=+]
table[row sep=crcr]{
1	0.0253314610712142 \\
4	0.00557707496060341 \\
7	0.00172880703648286 \\
10	0.00123878192984628 \\
13	0.000978227192515464 \\
16	0.000418504548886555 \\
19	0.000363417542451331 \\
22	0.000307454890517113 \\
25	0.000273791171134780 \\
28	0.000235180611385680 \\
31	0.000117257068803280 \\
34	0.000102932364418764 \\
};

%% p-QLP
\addplot+[smooth,color=red,loosely dotted, every mark/.append style={solid}, mark=pentagon]
table[row sep=crcr]{
1	0.0253291364504947 \\
4	0.00453695451002515 \\
7	0.00162278424610541 \\
10	0.00114228127275031 \\ 
13	0.000586402837939474 \\
16	0.000376264658868274 \\
19	0.000297394920744826 \\
22	0.000263721671171032 \\
25	0.000169179512300632 \\
28	0.000151056460408195 \\
31	0.000112326767566525 \\
34	8.90472755751952e-05 \\
};

%%% R-SVD
\addplot+[smooth,color=teal,loosely dotted, every mark/.append style={solid}, mark=diamond]
table[row sep=crcr]{
1	0.0253409416490037 \\
4	0.00415482024223132 \\
7	0.00175814632266733 \\
10	0.000931436500304695 \\
13	0.000564838056099219 \\
16	0.000384152473407484 \\
19	0.000288374821722182 \\
22	0.000216618855494341 \\
25	0.000166729948097564 \\
28	0.000132007197598508 \\
31	0.000110643435384033 \\
34	9.33900792815765e-05 \\
};

%%% CoR-UTV
\addplot+[smooth,color=blue,loosely dotted, every mark/.append style={solid}, mark=star]
table[row sep=crcr]{
1	0.0254985572255141 \\
4	0.00420639734212619 \\
7	0.00170604688002418 \\
10	0.000919913136076917 \\
13	0.000582791200047881 \\
16	0.000389197640211558 \\
19	0.000276170852384856 \\
22	0.000209082213386052 \\
25	0.000168336518730148 \\
28	0.000133854303937680 \\
31	0.000109151619810504 \\
34	9.31082924684984e-05 \\
};

%%% PbP-QLP
\addplot+[smooth,color=green,loosely dotted, every mark/.append style={solid}, mark=triangle]
table[row sep=crcr]{
1	0.0253296795700171 \\
4	0.00434267293128975 \\
7	0.00166719804554977 \\
10	0.000894396492756013 \\
13	0.000574659881077722 \\
16	0.000407779806570969 \\
19	0.000285755033240783 \\
22	0.000219340548037172 \\
25	0.000165653112677761 \\
28	0.000131561944153018 \\
31	0.000109669227269415 \\
34	9.31861047657430e-05 \\
};

\end{axis}

\end{tikzpicture}%
	\captionsetup{justification=centering,font=scriptsize}
		\caption{Low-rank approximation errors for \texttt{Derive2}. Left: basic PbP-QLP. Right: PI-coupled PbP-QLP with $q=2$.}
		\label{fig_Derive2}       % Give a unique label
	\end{center} 
\end{figure}

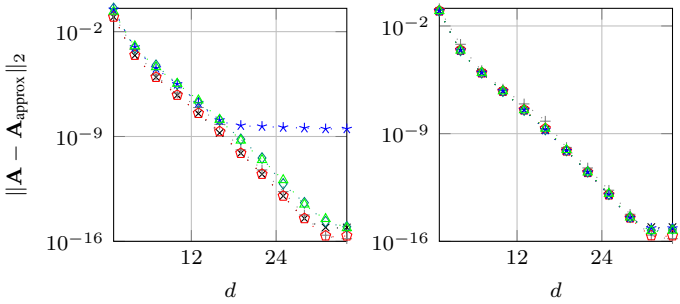
\begin{figure}[t]
	\begin{center}
			% This file was created by matlab2tikz v0.4.7 running on MATLAB 8.3.
\usetikzlibrary{positioning,calc}

\definecolor{mycolor1}{rgb}{0.00000,1.00000,1.00000}%
\definecolor{mycolor2}{rgb}{1.00000,0.00000,1.00000}%

\pgfplotsset{every axis label/.append style={font=\footnotesize},
every tick label/.append style={font=\footnotesize}
}

\begin{tikzpicture}[font=\footnotesize] 

\begin{axis}[%
name=ber,
ymode=log,
width  = 0.35\columnwidth,%5.63489583333333in,
height = 0.35\columnwidth,%4.16838541666667in,
scale only axis,
xmin  = 1,
xmax  = 34,
xlabel= {$d$},
xmajorgrids,
ymin = 1e-16 ,
ymax = 0.36,
xtick       ={12, 24},
xticklabels ={$12$,$24$},
ylabel={$\|{\bf A}- {\bf A}_{\text{approx}}\|_2$},
ymajorgrids,
]

%% SVD 
\addplot+[smooth,color=black,loosely dotted, every mark/.append style={solid}, mark=x]
table[row sep=crcr]{
1	0.0956716192948380 \\
4	0.000257600119379796 \\
7	9.10762762333226e-06 \\
10	5.73240027087217e-07 \\
13	3.39891920144121e-08 \\
16	1.72516560746357e-09 \\
19	7.65954357669311e-11 \\
22	3.01582701016826e-12 \\
25	1.06188850192086e-13 \\
28	3.36304751830057e-15 \\
31	8.42759470427519e-16 \\
34	8.42484345895808e-16 \\
};
%% QRP
\addplot+[smooth,color=gray,loosely dotted, every mark/.append style={solid}, mark=+]
table[row sep=crcr]{
1	0.134270058274722 \\
4	0.000644037351761653 \\
7	1.26955141221978e-05 \\
10	7.91887308174803e-07 \\
13	8.57842671368455e-08 \\
16	6.45624953691529e-09 \\
19	1.26798696206190e-10 \\
22	5.88196329974884e-12 \\
25	2.09563463602859e-13 \\
28	5.72871300417339e-15 \\
31	2.55832974092800e-16 \\
34	1.49031108076020e-16 \\
};

%% p-QLP
\addplot+[smooth,color=red,loosely dotted, every mark/.append style={solid}, mark=pentagon]
table[row sep=crcr]{
1	0.0963224308998651 \\
4	0.000271090314634495 \\
7	9.12112037110349e-06 \\
10	5.80936974275390e-07 \\
13	3.59785266920599e-08 \\
16	2.15531043984068e-09 \\
19	7.67034386081641e-11 \\
22	3.07512284294602e-12 \\
25	1.06981772519936e-13 \\
28	3.38668847598600e-15 \\
31	2.37537295172259e-16 \\
34	2.36113349541483e-16 \\
  };

%%% R-SVD
\addplot+[smooth,color=teal,loosely dotted, every mark/.append style={solid}, mark=diamond]
table[row sep=crcr]{
1	0.357810041308114 \\
4	0.000818200184179023 \\
7	5.72312897457107e-05 \\
10	3.21287814521894e-06 \\
13	1.83009287220744e-07 \\
16	1.26449805981652e-08 \\
19	6.53235453834285e-10 \\
22	3.66080706177309e-11 \\
25	6.91513085335440e-13 \\
28	3.76866133221617e-14 \\
31	2.08704768075344e-15 \\
34	9.42735452334095e-16 \\
};

%%% CoR-UTV
\addplot+[smooth,color=blue,loosely dotted, every mark/.append style={solid}, mark=star]
table[row sep=crcr]{
1	0.301238544189366 \\
4	0.000906336087362746 \\
7	3.61068813229988e-05 \\
10	3.12449768227715e-06 \\
13	1.73395696753896e-07 \\
16	1.22203884910667e-08 \\
19	5.41999438031248e-09 \\
22	4.69104992838089e-09 \\
25	4.04363364543578e-09 \\
28	3.66628085551699e-09 \\
31	3.23488407187953e-09 \\
34	3.25167810685580e-09 \\
};

%%% PbP-QLP
\addplot+[smooth,color=green,loosely dotted, every mark/.append style={solid}, mark=triangle]
table[row sep=crcr]{
1	0.221238609403774 \\
4	0.00108218003866119 \\
7	4.34045802877293e-05 \\
10	3.37923334787084e-06 \\
13	2.63766787394679e-07 \\
16	1.33211459916580e-08 \\
19	5.46347770128864e-10 \\
22	2.72830408040728e-11 \\
25	1.24882640977736e-12 \\
28	2.91652880635164e-14 \\
31	3.15851783749670e-15 \\
34	7.78882673904361e-16 \\
};

\end{axis}

\begin{axis}[%
name=SumRate,
at={($(ber.east)+(35,0em)$)},
		anchor= west,
ymode=log,
width  = 0.35\columnwidth,%5.63489583333333in,
height = 0.35\columnwidth,%4.16838541666667in,
scale only axis,
xmin  = 1,
xmax  = 34,
xlabel= {$d$},
xmajorgrids,
ymin = 1e-16 ,
ymax = 0.14,
xtick       = {12, 24},
xticklabels = {$12$,$24$},
ylabel={},
ymajorgrids,
%title = {\texttt{HighNoiseLowRank}}]
%ytick       ={0.0973831, 0.0973830 , 0.0973829, 0.0973828},
%yticklabels ={$9.73831$, $9.73830$ , $9.73829$, $9.73828$},
% legend entries = {SVD,CPQR,p-QLP, R-SVD,CoR-UTV,PbP-QLP},
% legend style={at={(1,1)},anchor=north east,draw=black,fill=white,legend cell align=left,font=\tiny}
]
%% SVD 
\addplot+[smooth,color=black,loosely dotted, every mark/.append style={solid}, mark=x]
table[row sep=crcr]{
1	0.0956716192948380 \\
4	0.000257600119379796 \\
7	9.10762762333226e-06 \\
10	5.73240027087217e-07 \\
13	3.39891920144121e-08 \\
16	1.72516560746357e-09 \\
19	7.65954357669311e-11 \\
22	3.01582701016826e-12 \\
25	1.06188850192086e-13 \\
28	3.36304751830057e-15 \\
31	8.42759470427519e-16 \\
34	8.42484345895808e-16 \\
};
%% QRP
\addplot+[smooth,color=gray,loosely dotted, every mark/.append style={solid}, mark=+]
table[row sep=crcr]{
1	0.134270058274722  \\
4	0.000644037351761653 \\
7	1.26955141221978e-05 \\
10	7.91887308174803e-07 \\
13	8.57842671368455e-08 \\
16	6.45624953691529e-09 \\
19	1.26798696206190e-10 \\
22	5.88196329974884e-12 \\
25	2.09563463602859e-13 \\
28	5.72871300417339e-15 \\
31	2.55832974092800e-16 \\
34	1.49031108076020e-16 \\
};

%% p-QLP
\addplot+[smooth,color=red,loosely dotted, every mark/.append style={solid}, mark=pentagon]
table[row sep=crcr]{
1	0.0963224308998651 \\
4	0.000271090314634495 \\
7	9.12112037110349e-06 \\
10	5.80936974275390e-07 \\
13	3.59785266920599e-08 \\
16	2.15531043984068e-09 \\
19	7.67034386081641e-11 \\
22	3.07512284294602e-12 \\
25	1.06981772519936e-13 \\
28	3.38668847598600e-15 \\
31	2.37537295172259e-16 \\
34	2.36113349541483e-16 \\
};

%%% R-SVD
\addplot+[smooth,color=teal,loosely dotted, every mark/.append style={solid}, mark=diamond]
table[row sep=crcr]{
1	0.0956716459690191 \\
4	0.000257603950371812 \\
7	9.12838871104879e-06 \\
10	5.77560972200774e-07 \\
13	3.40506733038672e-08 \\
16	1.86013758174468e-09 \\
19	8.11071413910035e-11 \\
22	3.34559634817789e-12 \\
25	1.15655794216857e-13 \\
28	3.75702839638612e-15 \\
31	5.36785462444291e-16 \\
34	6.45851266985400e-16 \\
};

%%% CoR-UTV
\addplot+[smooth,color=blue,loosely dotted, every mark/.append style={solid}, mark=star]
table[row sep=crcr]{
1	0.0956732955771514 \\
4	0.000257603947065809 \\
7	9.11926946305147e-06 \\
10	5.75461245191545e-07 \\ 
13	3.40114653769158e-08 \\
16	1.73782818252858e-09 \\
19	7.67864910568822e-11 \\
22	3.01670015170364e-12 \\
25	1.06214382697837e-13 \\
28	3.36556691222560e-15 \\
31	7.17638990325583e-16 \\
34	7.08398027130433e-16 \\
};

%%% PbP-QLP
\addplot+[smooth,color=green,loosely dotted, every mark/.append style={solid}, mark=triangle]
table[row sep=crcr]{
1	0.0956716652696529 \\
4	0.000257627567693187 \\
7	9.15143264057559e-06 \\
10	5.82001889069071e-07 \\
13	3.41844105062355e-08 \\
16	2.16353501350910e-09 \\
19	8.78972135896319e-11 \\
22	3.10976548954442e-12 \\
25	1.26566674138327e-13 \\
28	3.61958758631711e-15 \\
31	4.77188369444617e-16 \\
34	5.02174054709763e-16 \\
};

\end{axis}

\end{tikzpicture}%
        \captionsetup{justification=centering,font=scriptsize}
		\caption{Low-rank approximation errors for \texttt{Foxgood}. Left: basic PbP-QLP. Right: PI-coupled PbP-QLP with $q=2$.}
		\label{fig_Foxgood}       % Give a unique label
	\end{center}
\end{figure}

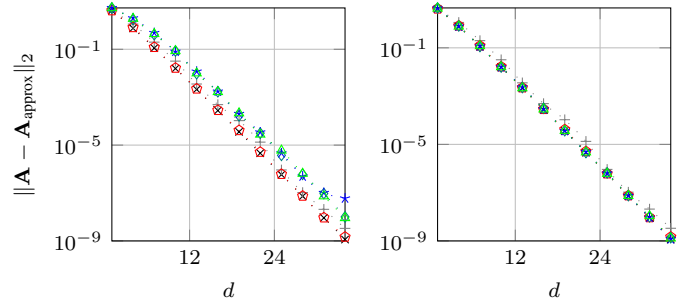
\begin{figure}[t]
	\begin{center}
			% This file was created by matlab2tikz v0.4.7 running on MATLAB 8.3.
\usetikzlibrary{positioning,calc}

\definecolor{mycolor1}{rgb}{0.00000,1.00000,1.00000}%
\definecolor{mycolor2}{rgb}{1.00000,0.00000,1.00000}%

\pgfplotsset{every axis label/.append style={font=\footnotesize},
every tick label/.append style={font=\footnotesize}
}

\begin{tikzpicture}[font=\footnotesize] 

\begin{axis}[%
name=ber,
ymode=log,
width  = 0.35\columnwidth,%5.63489583333333in,
height = 0.35\columnwidth,%4.16838541666667in,
scale only axis,
xmin  = 1,
xmax  = 34,
xlabel= {$d$},
xmajorgrids,
ymin = 1e-9 ,
ymax = 5.32,
xtick       ={12, 24},
xticklabels ={$12$,$24$},
ylabel={$\|{\bf A}- {\bf A}_{\text{approx}}\|_2$},
ymajorgrids,
]

%% SVD 
\addplot+[smooth,color=black,loosely dotted, every mark/.append style={solid}, mark=x]
table[row sep=crcr]{
1	4.13280231549347 \\
4	0.750560667599720 \\
7	0.112396226471918 \\
10	0.0156827216556171 \\
13	0.00210798471431051 \\
16	0.000277098446260804 \\
19	3.58561275426095e-05 \\
22	4.58827918508173e-06 \\
25	5.81872378599350e-07 \\
28	7.32789553254048e-08 \\
31	9.17241495485784e-09 \\
34	1.14238648100648e-09 \\
};
%% QRP
\addplot+[smooth,color=gray,loosely dotted, every mark/.append style={solid}, mark=+]
table[row sep=crcr]{
1	4.13300462468023 \\
4	1.12699801712495 \\
7	0.198725948861991 \\
10	0.0320065407254139 \\
13	0.00354836741389733 \\
16	0.000489150448219918 \\
19	0.000104989172679000 \\
22	1.33108338616250e-05 \\
25	9.18426766241547e-07 \\
28	1.04633348993537e-07 \\
31	2.08359333883915e-08 \\
34	3.34247671612049e-09 \\
};

%% p-QLP
\addplot+[smooth,color=red,loosely dotted, every mark/.append style={solid}, mark=pentagon]
table[row sep=crcr]{
1	4.13285507122188 \\
4	0.820851303289579 \\
7	0.121807893574546 \\
10	0.0165724724733656 \\
13	0.00224006517801533 \\
16	0.000285369280138156 \\
19	4.12715040054313e-05 \\
22	5.12342975418468e-06 \\
25	6.11029071954657e-07 \\
28	7.54392629954976e-08 \\
31	9.34497243606642e-09 \\
34	1.46687974395360e-09 \\
  };

%%% R-SVD
\addplot+[smooth,color=teal,loosely dotted, every mark/.append style={solid}, mark=diamond]
table[row sep=crcr]{
1	5.30776701050844 \\
4	1.77438420335113 \\
7	0.407227787584078 \\
10	0.0846731881664590 \\
13	0.00956932950649088 \\
16	0.00169341500444783 \\
19	0.000167685325183451 \\
22	3.14797628190164e-05 \\
25	3.73601286197991e-06 \\
28	6.03228658336309e-07 \\
31	9.49429178004782e-08 \\
34	1.00518739307991e-08 \\
};

%%% CoR-UTV
\addplot+[smooth,color=blue,loosely dotted, every mark/.append style={solid}, mark=star]
table[row sep=crcr]{
1	5.27670471277568 \\
4	1.97805144539808 \\
7	0.481332784739154 \\
10	0.0798009220885210 \\
13	0.0118122037055102 \\
16	0.00172984198672768 \\
19	0.000201323969774295 \\
22	3.42334332545650e-05 \\
25	4.60244141788299e-06 \\
28	4.94701305518421e-07 \\
31	1.01299051747020e-07 \\
34	5.96484079263365e-08 \\
};

%%% PbP-QLP
\addplot+[smooth,color=green,loosely dotted, every mark/.append style={solid}, mark=triangle]
table[row sep=crcr]{
1	5.25316239508665 \\
4	1.89952853828671 \\
7	0.438363012553688 \\
10	0.0852075224618425 \\
13	0.00999847706035529 \\
16	0.00165791078899728 \\
19	0.000216646473928695 \\
22	2.78120680867136e-05 \\
25	6.00488756255166e-06 \\ 
28	6.47331251185362e-07 \\
31	7.37240349490582e-08 \\
34	9.07432100218845e-09 \\
};

\end{axis}

\begin{axis}[%
name=SumRate,
at={($(ber.east)+(35,0em)$)},
		anchor= west,
ymode=log,
width  = 0.35\columnwidth,%5.63489583333333in,
height = 0.35\columnwidth,%4.16838541666667in,
scale only axis,
xmin  = 1,
xmax  = 34,
xlabel= {$d$},
xmajorgrids,
ymin = 1e-9 ,
ymax = 4.5,
xtick       = {12, 24},
xticklabels = {$12$,$24$},
ylabel={},
ymajorgrids,
%title = {\texttt{HighNoiseLowRank}}]
%ytick       ={0.0973831, 0.0973830 , 0.0973829, 0.0973828},
%yticklabels ={$9.73831$, $9.73830$ , $9.73829$, $9.73828$},
% legend entries = {SVD,CPQR,p-QLP, R-SVD,CoR-UTV,PbP-QLP},
% legend style={at={(1,1)},anchor=north east,draw=black,fill=white,legend cell align=left,font=\tiny}
]
%% SVD 
\addplot+[smooth,color=black,loosely dotted, every mark/.append style={solid}, mark=x]
table[row sep=crcr]{
1	4.13280231549347 \\
4	0.750560667599720 \\
7	0.112396226471918 \\
10	0.0156827216556171 \\
13	0.00210798471431051 \\
16	0.000277098446260804 \\
19	3.58561275426095e-05 \\
22	4.58827918508173e-06 \\
25	5.81872378599350e-07 \\
28	7.32789553254048e-08 \\
31	9.17241495485784e-09 \\
34	1.14238648100648e-09 \\
};
%% QRP
\addplot+[smooth,color=gray,loosely dotted, every mark/.append style={solid}, mark=+]
table[row sep=crcr]{
1	4.13300462468023 \\
4	1.12699801712495 \\
7	0.198725948861991 \\
10	0.0320065407254139 \\
13	0.00354836741389733 \\ 
16	0.000489150448219918 \\
19	0.000104989172679000 \\
22	1.33108338616250e-05 \\
25	9.18426766241547e-07 \\
28	1.04633348993537e-07 \\
31	2.08359333883915e-08 \\
34	3.34247671612049e-09 \\
};

%% p-QLP
\addplot+[smooth,color=red,loosely dotted, every mark/.append style={solid}, mark=pentagon]
table[row sep=crcr]{
1	4.13285507122188 \\
4	0.820851303289579 \\
7	0.121807893574546 \\
10	0.0165724724733656 \\
13	0.00224006517801533 \\
16	0.000285369280138156 \\
19	4.12715040054313e-05 \\
22	5.12342975418468e-06 \\
25	6.11029071954657e-07 \\
28	7.54392629954976e-08 \\ 
31	9.34497243606642e-09 \\
34	1.46687974395360e-09 \\
};

%%% R-SVD
\addplot+[smooth,color=teal,loosely dotted, every mark/.append style={solid}, mark=diamond]
table[row sep=crcr]{
1	4.48788272287812 \\
4	0.808970755895977 \\
7	0.113985220813345 \\
10	0.0160483655687348 \\
13	0.00227157636688030 \\
16	0.000315228965963038 \\
19	3.70195067065118e-05 \\
22	4.61299084518886e-06 \\
25	6.60157830460212e-07 \\
28	7.55497666543928e-08 \\
31	9.99209904642847e-09 \\
34	1.36455883066675e-09 \\
};

%%% CoR-UTV
\addplot+[smooth,color=blue,loosely dotted, every mark/.append style={solid}, mark=star]
table[row sep=crcr]{
1	4.35027079140640 \\
4	0.770363471048884 \\
7	0.115520896677420 \\ 
10	0.0157951943845542 \\
13	0.00232463590147341 \\
16	0.000283884700274534 \\
19	3.59339147981232e-05 \\
22	4.85304798739909e-06 \\
25	6.30167311136662e-07 \\
28	7.42226302275993e-08 \\
31	9.28218099214756e-09 \\
34	1.18651582751099e-09 \\
};

%%% PbP-QLP
\addplot+[smooth,color=green,loosely dotted, every mark/.append style={solid}, mark=triangle]
table[row sep=crcr]{
1	4.29308360689616 \\
4	0.772938495544837 \\
7	0.118815935633072 \\
10	0.0159522155154107 \\
13	0.00215012152362933 \\
16	0.000285195753062016 \\
19	3.68865752897043e-05 \\
22	4.72894792163353e-06 \\
25	6.11080014277020e-07 \\
28	7.40261883920451e-08 \\
31	1.07408003681279e-08 \\
34	1.28456999882227e-09 \\
};

\end{axis}

\end{tikzpicture}%
	\captionsetup{justification=centering,font=scriptsize}
		\caption{Low-rank approximation errors for \texttt{Gravity}. Left: basic PbP-QLP. Right: PI-coupled PbP-QLP  with $q=2$.}
		\label{fig_Gravity}      
	\end{center}
\end{figure}

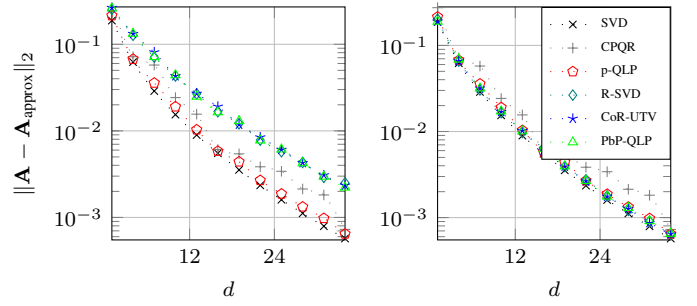
\begin{figure}[t] 
	\begin{center}
			% This file was created by matlab2tikz v0.4.7 running on MATLAB 8.3.
\usetikzlibrary{positioning,calc}

\definecolor{mycolor1}{rgb}{0.00000,1.00000,1.00000}%
\definecolor{mycolor2}{rgb}{1.00000,0.00000,1.00000}%

\pgfplotsset{every axis label/.append style={font=\footnotesize},
every tick label/.append style={font=\footnotesize}
}

\begin{tikzpicture}[font=\footnotesize] 

\begin{axis}[%
name=ber,
ymode=log,
width  = 0.35\columnwidth,%5.63489583333333in,
height = 0.35\columnwidth,%4.16838541666667in,
scale only axis,
xmin  = 1,
xmax  = 34,
xlabel= {$d$},
xmajorgrids,
ymin = 5.5e-4 ,
ymax = 0.27,
xtick       ={12, 24},
xticklabels ={$12$,$24$},
ylabel={$\|{\bf A}- {\bf A}_{\text{approx}}\|_2$},
ymajorgrids,
]

%% SVD 
\addplot+[smooth,color=black,loosely dotted, every mark/.append style={solid}, mark=x]
table[row sep=crcr]{
1	0.187893411176044 \\
4	0.0622491560619530 \\
7	0.0288357010335887 \\
10	0.0154105445748422 \\
13	0.00895924902938751 \\
16	0.00551616556836378 \\
19	0.00354284691116643 \\
22	0.00235138711705871 \\
25	0.00160242913335762 \\
28	0.00111620328937081 \\
31	0.000792043935762179 \\
34	0.000571047570608606 \\
};
%% QRP
\addplot+[smooth,color=gray,loosely dotted, every mark/.append style={solid}, mark=+]
table[row sep=crcr]{
1	0.274347466817268 \\
4	0.0718432541745160 \\
7	0.0575832588969413 \\
10	0.0242258181315438 \\
13	0.0155920509684725 \\
16	0.00608252967624553 \\
19	0.00543034901530827 \\
22	0.00384671799002933 \\
25	0.00341315414061041 \\
28	0.00213174768557187 \\
31	0.00181901640177864 \\
34	0.000727979296983441 \\
};

%% p-QLP
\addplot+[smooth,color=red,loosely dotted, every mark/.append style={solid}, mark=pentagon]
table[row sep=crcr]{
1	0.214624359532227 \\
4	0.0673511089256924 \\
7	0.0357805445445160 \\
10	0.0190138524179073 \\
13	0.0103123627753634 \\
16	0.00583563346947835 \\
19	0.00439744368023929 \\
22	0.00268332023371397 \\ 
25	0.00187220644027527 \\
28	0.00132695424890531 \\
31	0.000977203371746612 \\
34	0.000645771789476848 \\
  };

%%% R-SVD
\addplot+[smooth,color=teal,loosely dotted, every mark/.append style={solid}, mark=diamond]
table[row sep=crcr]{
1	0.253228504947032 \\
4	0.127372591486177 \\
7	0.0710225427674054 \\
10	0.0430698557594240 \\
13	0.0268036056801756 \\
16	0.0166614257956345 \\
19	0.0120965938187349 \\
22	0.00782051005978686 \\
25	0.00574908640760473 \\
28	0.00432315819895953 \\
31	0.00290834872160175 \\
34	0.00250638382333760 \\
};

%%% CoR-UTV
\addplot+[smooth,color=blue,loosely dotted, every mark/.append style={solid}, mark=star]
table[row sep=crcr]{
1	0.263589160145221 \\
4	0.131815143390645 \\
7	0.0809264520944593 \\
10	0.0429014277659937 \\
13	0.0269391056149635 \\
16	0.0190771986006617 \\
19	0.0118201072687176 \\
22	0.00842865162421442 \\
25	0.00606937741955195 \\
28	0.00425736641092803 \\
31	0.00307273232332830 \\
34	0.00234013785817400 \\
};

%%% PbP-QLP
\addplot+[smooth,color=green,loosely dotted, every mark/.append style={solid}, mark=triangle]
table[row sep=crcr]{
1	0.260640142314392 \\
4	0.132702524386254 \\
7	0.0714458872566242 \\
10	0.0440268770399225 \\
13	0.0249104376884569 \\
16	0.0164421617639594 \\
19	0.0130184563262043 \\
22	0.00777583770163915 \\
25	0.00612182489744210 \\
28	0.00424435214647495 \\
31	0.00296703685360733 \\
34	0.00222023653906745 \\
};

\end{axis}

\begin{axis}[%
name=SumRate,
at={($(ber.east)+(35,0em)$)},
		anchor= west,
ymode=log,
width  = 0.35\columnwidth,%5.63489583333333in,
height = 0.35\columnwidth,%4.16838541666667in,
scale only axis,
xmin  = 1,
xmax  = 34,
xlabel= {$d$},
xmajorgrids,
ymin = 5.5e-4 ,
ymax = 0.28,
xtick       = {12, 24},
xticklabels = {$12$,$24$},
ylabel={},
ymajorgrids,
%title = {\texttt{HighNoiseLowRank}}]
%ytick       ={0.0973831, 0.0973830 , 0.0973829, 0.0973828},
%yticklabels ={$9.73831$, $9.73830$ , $9.73829$, $9.73828$},
legend entries = {SVD,CPQR,p-QLP, R-SVD,CoR-UTV,PbP-QLP},
legend style={at={(1,1)},anchor=north east,draw=black,fill=white,legend cell align=left,font=\tiny}
]
%% SVD 
\addplot+[smooth,color=black,loosely dotted, every mark/.append style={solid}, mark=x]
table[row sep=crcr]{
1	0.187893411176044 \\
4	0.0622491560619530 \\
7	0.0288357010335887 \\
10	0.0154105445748422 \\
13	0.00895924902938751 \\
16	0.00551616556836378 \\
19	0.00354284691116643 \\
22	0.00235138711705871 \\
25	0.00160242913335762 \\
28	0.00111620328937081 \\
31	0.000792043935762179 \\
34	0.000571047570608606 \\
};
%% QRP
\addplot+[smooth,color=gray,loosely dotted, every mark/.append style={solid}, mark=+]
table[row sep=crcr]{
1	0.274347466817268 \\
4	0.0718432541745160 \\
7	0.0575832588969413 \\
10	0.0242258181315438 \\
13	0.0155920509684725 \\
16	0.00608252967624553 \\
19	0.00543034901530827 \\
22	0.00384671799002933 \\
25	0.00341315414061041 \\
28	0.00213174768557187 \\
31	0.00181901640177864 \\
34	0.000727979296983441 \\
};

%% p-QLP
\addplot+[smooth,color=red,loosely dotted, every mark/.append style={solid}, mark=pentagon]
table[row sep=crcr]{
1	0.214624359532227 \\
4	0.0673511089256924 \\
7	0.0357805445445160 \\
10	0.0190138524179073 \\
13	0.0103123627753634 \\
16	0.00583563346947835 \\
19	0.00439744368023929 \\
22	0.00268332023371397 \\
25	0.00187220644027527 \\
28	0.00132695424890531 \\
31	0.000977203371746612 \\
34	0.000645771789476848 \\
};

%%% R-SVD
\addplot+[smooth,color=teal,loosely dotted, every mark/.append style={solid}, mark=diamond]
table[row sep=crcr]{
1	0.202149602008033 \\
4	0.0658263091512935 \\
7	0.0304300571634164 \\
10	0.0166005127830860 \\
13	0.00974117077250112 \\
16	0.00639794468922561 \\
19	0.00391341769982865 \\
22	0.00274508162605611 \\
25	0.00178495152393761 \\
28	0.00121450328965939 \\
31	0.000907373270160875 \\
34	0.000635790512174031 \\
};

%%% CoR-UTV
\addplot+[smooth,color=blue,loosely dotted, every mark/.append style={solid}, mark=star]
table[row sep=crcr]{
1	0.190078529151726 \\
4	0.0656607745076157 \\
7	0.0316506016129677 \\
10	0.0169177964355518 \\
13	0.0101703879754911 \\
16	0.00596401286449128 \\
19	0.00391987436174913 \\
22	0.00263909160748768 \\
25	0.00175404737567475 \\
28	0.00126063981748932 \\
31	0.000866806461568089 \\
34	0.000640601317625521 \\
};

%%% PbP-QLP
\addplot+[smooth,color=green,loosely dotted, every mark/.append style={solid}, mark=triangle]
table[row sep=crcr]{
1	0.189659041391145 \\
4	0.0688017782122353 \\
7	0.0317334493506520 \\
10	0.0171761869649230 \\
13	0.0100686637176157 \\
16	0.00632221896907596 \\
19	0.00394086850450326 \\
22	0.00267501750208093 \\
25	0.00176052309777271 \\ 
28	0.00129868211299128 \\
31	0.000912991039334673 \\
34	0.000651003755358936 \\
};

\end{axis}

\end{tikzpicture}%
    \captionsetup{justification=centering,font=scriptsize}
		\caption{Low-rank approximation errors for \texttt{Heat}. Left: basic PbP-QLP. Right: PI-coupled PbP-QLP with $q=2$.}
		\label{fig_Heat}      
	\end{center}
\end{figure}

With matrices from Table \ref{Table_PracMat} as our inputs, similar to the experiment in Section \ref{subsec_LRA_SynData}, we construct rank-$d$ approximations using different algorithms and compute the approximation errors.  The results are shown in Figs. \ref{fig_Baart}-\ref{fig_Heat}. We observe that while PbP-QLP with no PI scheme provides fairly accurate low-rank approximations for \texttt{Baart}, \texttt{Foxgood} and \texttt{Gravity}, the approximations for \texttt{Deriv2} and \texttt{Heat}, similar to those of R-SVD, are rather poor. This is because these two matrices have slowly decaying singular values. However, the errors incurred by PI-coupled PbP-QLP overlap those produced by the optimal SVD for all five matrices, showing the high accuracy of PbP-QLP. We further observe in Figs. \ref{fig_Baart} and \ref{fig_Foxgood} that for \texttt{Baart} and \texttt{Foxgood}, basic CoR-UTV produces less accurate results as the rank parameter increases. This is due to the application of input matrix to the sample matrix (Step 4 of Algorithm \ref{Alg_CoRUTV}) without any orthonormalization being applied (see Section \ref{subsec_PIPbPQLP}). To be precise, in this case, the best accuracy attainable by the algorithm is $\sigma_1 \epsilon_\text{machine}^{1/2}$. Let $\epsilon_\text{machine}=10^{-16}$, and we have the largest singular values of \texttt{Baart} and \texttt{Foxgood} equal to $3.2$ and $0.8$, respectively. Thus, the best accuracy attinable by basic CoR-UTV for \texttt{Baart} is $3.2 \times 10^{-8}$, and for  \texttt{Foxgood} is $8\times 10^{-9}$. This issue, however, is resolved by applying the orthonormalization scheme, as the plots show.

\subsubsection{Matrix $\ell_2$-norm Estimation}
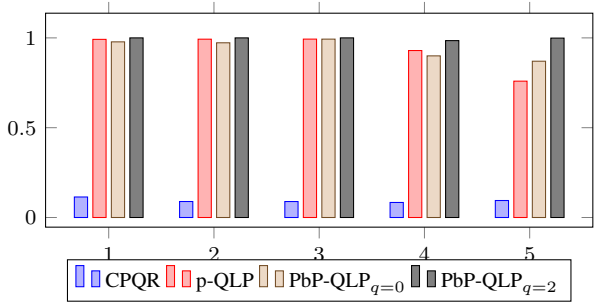
\begin{figure}[t]
	\begin{center}
			% This file was created by matlab2tikz v0.4.7 running on MATLAB 8.3.
% Copyright (c) 2008--2014, Nico Schlömer <nico.schloemer@gmail.com>
% All rights reserved.
% Minimal pgfplots version: 1.3
% 
% The latest updates can be retrieved from
%   http://www.mathworks.com/matlabcentral/fileexchange/22022-matlab2tikz
% where you can also make suggestions and rate matlab2tikz.
% 
%
% defining custom colors
\usetikzlibrary{positioning,calc}

\definecolor{mycolor1}{rgb}{0.00000,1.00000,1.00000}%
\definecolor{mycolor2}{rgb}{1.00000,0.00000,1.00000}%

\pgfplotsset{every axis label/.append style={font=\footnotesize},
every tick label/.append style={font=\footnotesize}
}

\begin{tikzpicture}[font=\footnotesize] 

\begin{axis}[
width  = 1\columnwidth,
height = 0.5\columnwidth,
x tick label style={
	/pgf/number format/1000 sep=},
%ylabel=Population,
enlargelimits=0.15,
legend style={at={(0.5,-0.15)},
	anchor=north,legend columns=-1},
ybar,
bar width=5pt,
]
\addplot
coordinates {(1,0.1142)    (2,0.0890)
(3,0.0889)          (4,0.0837)      (5,0.0946)};
\addplot
coordinates {(1,0.9918)    (2,0.9928)
(3,0.9932)          (4,0.9296)      (5,0.7591)};
\addplot
coordinates {(1,0.9779)        (2,0.9727)
(3,0.9931)             (4,0.8999)      (5,0.8701)};
\addplot
coordinates {(1,1.0)        (2,1.0)
(3,1.0)             (4,0.9848)      (5,0.9988)};
\legend{CPQR,p-QLP,$\text{PbP-QLP}_{q=0}$, $\text{PbP-QLP}_{q=2}$}
\end{axis}

\end{tikzpicture}% 
 	\captionsetup{justification=centering,font=scriptsize}
    \caption{Ratios of the estimated $\ell_2$-norm to the exact norm for matrices of Table \ref{Table_PracMat}.} 
		\label{fig_Ratio} 
	\end{center}
\end{figure}

For any matrix $\bf A$, $\|{\bf A}\|_2$ is equal to its largest singular value. We compute an estimation to the largest singular value of each matrix from Table \ref{Table_PracMat} using PbP-QLP, and compare the results with those of CPQR and p-QLP. Fig. \ref{fig_Ratio} displays the ratios of the estimated singular values to the exact norms. It is seen that (i) basic PbP-QLP substantially outperforms CPQR in estimating the matrix $\ell_2$-norm, while it produces comparable results with p-QLP, and (ii) PI-coupled PbP-QLP provides excellent estimations to the first singular values and its  performance exceeds that of p-QLP for all matrices.

\section{Conclusion}
\label{section_concl}
We presented in this paper the rank-revealing PbP-QLP algorithm, which, by utilizing randomization, constructs an approximation to the pivoted QLP and truncated SVD. PbP-QLP is primarily designed to approximate low-rank matrices. It consists of two stages, each performing only the unpivoted QR factorization to factor the associated small matrices. With theoretical analysis, we showed that the numerical rank of a given matrix is revealed in the first stage. We further furnished a detailed theoretical analysis for PbP-QLP, which brings an insight into the rank-revealing property as well as the accuracy of the algorithm. Through numerical tests conducted on several classes of matrices, we showed our proposed PbP-QLP (i) outperforms R-SVD and CoR-UTV in runtime, and (ii) establishes highly accurate approximations, as accurate as those of the optimal SVD, to the matrices.  

% in the first stage, by utilizing randomization, an input matrix is first compressed and then factored using a QR algorithm. With theoretical analysis, we showed the R-factor reveals the numerical rank of the matrix. In the second stage, the transpose of the R-factor is deomposed with a QR algorithm, and the final approximation is constrcuted by projecting the reduced matrices back to the original space.

\appendices
\section{Proof of Theorem \ref{Th_R22bound}}
\label{Proo_R22bound}
We first prove \eqref{eq_2nR22_bound} for the case $q = 0$ (the basic PbP-QLP). To do so, we write $\bf D$ and its $\bf Q$ and $\bf R$ factors, equations \eqref{eq_D} and \eqref{eq_DQR}, with partitioned matrices:
\begin{equation}\label{key}
\begin{aligned}
{\bf D}  = {\bf A}\bar{\bf P} = {\bf QR}  =[{\bf D}_1\quad {\bf D}_2]=[{\bf Q}_1\quad {\bf Q}_2]
\begin{bmatrix}
{\bf R}_{11} & {\bf R}_{12}  \\
{\bf 0} & {\bf R}_{22}
\end{bmatrix},
\end{aligned}
\notag
\end{equation}
where ${\bf D}_1$ and ${\bf Q}_1$ contain the first $k$ columns, and ${\bf D}_2$ and ${\bf Q}_2$ contain the remaining $d-k$ columns of ${\bf D}$ and ${\bf Q}$, respectively.
We now define a matrix $\bf X$ constructed by the interaction of $\bf V$ (right singular vectors of $\bf A$) and $\bar{\bf P}$:
\begin{equation}
 {\bf X} \coloneqq {\bf V}^T\bar{\bf P} = \begin{bmatrix}
{\bf V}_k^T  \\
{\bf V}_\perp^T
\end{bmatrix}[\bar{\bf P}_1\quad \bar{\bf P}_2]
= \begin{bmatrix}
{\bf X}_{11} & {\bf X}_{12}  \\
{\bf X}_{21} & {\bf X}_{22}
\end{bmatrix},
\label{eq_X}
\end{equation}
where $\bar{\bf P}_1$ contains the first $k$ columns, and $\bar{\bf P}_2$ contains the remaining $d-k$ columns of $\bar{\bf P}$. With a simple computation, we will obtain the following four equalities:
\begin{equation}\label{eq_D1}
{\bf D}_1 = {\bf U}_k{\bf \Sigma}_k{\bf X}_{11} + {\bf U}_{\perp}{\bf \Sigma}_{\perp}{\bf X}_{21}.
\end{equation}

\begin{equation}\label{eq_D2}
\begin{aligned}
 {\bf D}_2 = {\bf U}_k{\bf \Sigma}_k{\bf X}_{12} + {\bf U}_{\perp}{\bf \Sigma}_{\perp}{\bf X}_{22}.
\end{aligned}
\end{equation}

\begin{equation}\label{key}
{\bf D}_1 = {\bf Q}_1{\bf R}_{11}.
\notag
\end{equation}

\begin{equation}\label{key}
\begin{aligned}
{\bf P}_{{\bf D}_1^\perp} {\bf D}_2 = {\bf Q}_2{\bf R}_{22}  = ({\bf I} - {\bf D}_1{\bf D}_1^\dagger){\bf D}_2  =
({\bf I} - {\bf Q}_1{\bf Q}_1^T){\bf D}_2.
\end{aligned}
\notag
\end{equation}

Hence, ${\bf R}_{22}$ is obtained as
\begin{equation}
{\bf R}_{22} = {\bf Q}_2^T ({\bf I} - {\bf Q}_1{\bf Q}_1^T){\bf D}_2.
\notag
\end{equation}

We take the $\ell_2$-norm of the above identity:
\begin{equation}\label{key}
\begin{aligned}
\|{\bf R}_{22} \|_2 & \le \|{\bf Q}_2^T\|_2\|({\bf I} - {\bf Q}_1{\bf Q}_1^T){\bf D}_2\|_2\\&
\le \|({\bf I} - {\bf Q}_1{\bf Q}_1^T){\bf D}_2\|_2.
\end{aligned}
\notag
\end{equation}
The last relation follows since for any orthonormal matrix $\bf M$, $\|{\bf M}\|_2\le 1$. By replacing ${\bf D}_2$ \eqref{eq_D2} and applying  the triangle inequality, we obtain
\begin{equation}\label{key}
\begin{aligned}
\|{\bf R}_{22}\|_2 &\le \|({\bf I} - {\bf Q}_1{\bf Q}_1^T){\bf U}_k{\bf \Sigma}_k{\bf X}_{12}\|_2 \\ &+\|({\bf I} - {\bf Q}_1{\bf Q}_1^T){\bf U}_{\perp}{\bf \Sigma}_{\perp}{\bf X}_{22}\|_2.
\end{aligned}
\notag
\end{equation}
By substituting ${\bf X}_{12}$ and ${\bf X}_{22}$ \eqref{eq_X} into the above equation, and from the orthonormality of $\bar{\bf P}_2$, we will have
\begin{equation}\label{eq_2nR22_2terms}
\|{\bf R}_{22}\|_2 \le \|({\bf I} - {\bf Q}_1{\bf Q}_1^T){\bf A}_k\|_2 +\|({\bf I} - {\bf Q}_1{\bf Q}_1^T){\bf U}_{\perp}{\bf \Sigma}_\perp \|_2,
\end{equation}
where ${\bf A}_k \coloneqq {\bf U}_k{\bf \Sigma}_k{\bf V}_k^T$. We now bound the two terms on the right-hand side of \eqref{eq_2nR22_2terms}. To bound
the first term, firstly
\begin{equation}
\mathcal{R}({\bf D}) = \mathcal{R}({\bf A}\bar{\bf P})= \mathcal{R}({\bf AA}^T{\bf \Phi}).
\notag
\end{equation}
The second equality holds as $\mathcal{R}(\bar{\bf P})=\mathcal{R}({\bf A}^T{\bf \Phi})$. Hence,
\begin{equation}
\begin{aligned}
{\bf AA}^T{\bf \Phi}  = {\bf U \Sigma}^2{\bf U}^T {\bf \Phi} ={\bf U}\begin{bmatrix}
{\bf \Sigma}_1^{2} & {\bf 0} & {\bf 0} \\
{\bf 0} & {\bf \Sigma}_2^{2} & {\bf 0}\\
{\bf 0} & {\bf 0} & {\bf \Sigma}_3^{2}
\end{bmatrix}
\begin{bmatrix}
{\bf \Phi}_1\\
\\
{\bf \Phi}_2
\end{bmatrix}= {\bf Q}{\bf R},
\end{aligned}
\notag
\end{equation}
where ${\bf \Sigma}_1 \in \mathbb R^{k \times k}$, ${\bf \Sigma}_2 \in \mathbb R^{(d-p-k) \times (d-p-k)}$, and ${\bf \Sigma}_3 \in \mathbb R^{(n_2-d+p) \times (n_2-d+p)}$. Assuming that ${\bf \Phi}_1$ is full row rank and its Moore-Penrose inverse satisfies
\begin{equation}
{\bf \Phi}_1{\bf \Phi}_1^\dagger  ={\bf I},
\notag
\end{equation}
we define a non-singular matrix $\bf Y$ as follows:
\begin{equation}
{\bf Y} = \bigg[
{\bf \Phi}_1^\dagger
\begin{pmatrix}
{\bf \Sigma}_1^2 & {\bf 0}\\
{\bf 0} & {\bf \Sigma}_2^2
\end{pmatrix}^{-1}, \bar{\bf Y}\bigg],
\notag
\end{equation}
where $\bar{\bf Y}\in \mathbb R^{d \times p}$ is chosen so that ${\bf Y}\in \mathbb R^{d \times d}$ is non-singular and ${\bf \Phi}_1\bar{\bf Y} = \bf 0$. We then compute the matrix product:
\begin{equation}
\begin{aligned}
{\bf A}{\bf A}^T {\bf \Phi Y} =
& {\bf U}\begin{bmatrix}
\begin{pmatrix}
{\bf \Sigma}_1^2 & {\bf 0}\\
{\bf 0} & {\bf \Sigma}_2^2
\end{pmatrix}{\bf \Phi}_1\\
{\bf \Sigma}_3^2 {\bf \Phi}_2
\end{bmatrix}{\bf Y}= {\bf U} \begin{bmatrix}
{\bf I} & {\bf 0} & {\bf 0} \\
{\bf 0} & {\bf I} & {\bf 0}\\
{\bf Z}_1 & {\bf Z}_2 & {\bf Z}_3
\end{bmatrix},
\end{aligned}
\label{eq_AAtPhiY_part}
\end{equation}
where ${\bf Z}_1 = {\bf \Sigma}_3^2{\bf \Phi}_2{\bf \Phi}_1^\dagger {\bf \Sigma}_1^{-2} \in \mathbb R^{(n_2-d+p) \times k}$, ${\bf Z}_2 = {\bf \Sigma}_3^2{\bf \Phi}_2{\bf \Phi}_1^\dagger {\bf \Sigma}_2^{-2} \in \mathbb R^{(n_2-d+p) \times (d-p-k)}$, and
${\bf Z}_3 = {\bf \Sigma}_3^2{\bf \Phi}_2 \bar{\bf Y} \in \mathbb R^{(n_2-d+p) \times p}$. Let the matrix in \eqref{eq_AAtPhiY_part} have a QR factorization:
\begin{equation}
\begin{aligned}
{\bf U} \begin{bmatrix}
{\bf I} & {\bf 0} & {\bf 0} \\
{\bf 0} & {\bf I} & {\bf 0}\\
{\bf Z}_1 & {\bf Z}_2 & {\bf Z}_3
\end{bmatrix} =
{\bf Q}^\prime {\bf R}^\prime = \begin{bmatrix}
{{\bf Q}_1^\prime}^T\\
{{\bf Q}_2^\prime}^T\\
{{\bf Q}_3^\prime}^T
\end{bmatrix}^T
\begin{bmatrix}
{\bf R}_{11}^\prime & {\bf R}_{12}^\prime & {\bf R}_{13}^\prime \\
{\bf 0} & {\bf R}_{22}^\prime & {\bf R}_{23}^\prime\\
{\bf 0} & {\bf 0} & {\bf R}_{33}^\prime
\end{bmatrix}
\end{aligned}
\notag
\end{equation}
which gives
\begin{equation}\label{eq_Q1prime}
{\bf U} \begin{bmatrix}
{\bf I} \\
{\bf 0} \\
{\bf Z}_1
\end{bmatrix} = {\bf Q}_1^\prime{\bf R}_{11}^\prime.
\end{equation}
Since matrix $\bf Y$ is non-singular, by \cite[Lemma 4.1]{Gu2015}, we have ${\bf Q}{\bf Q}^T = {\bf Q}^\prime {{\bf Q}^\prime}^T$. Exploiting \eqref{eq_Q1prime}, it follows that
\begin{equation}
\begin{aligned}
&{\bf I} - {\bf Q}_1 {{\bf Q}_1}^T = {\bf U}
\begin{bmatrix}
{\bf I} - \bar{\bf Z}^{-1} & {\bf 0} & -\bar{\bf Z}^{-1}{\bf Z}_1^T \\
{\bf 0} & {\bf I} & {\bf 0}\\
- {\bf Z}_1 \bar{\bf Z}^{-1} & {\bf 0} &
{\bf I} - {\bf Z}_1 \bar{\bf Z}^{-1}{\bf Z}_1^T
\end{bmatrix} {\bf U}^T,
\end{aligned}
\notag
\end{equation}
where ${\bf Z}_1$ is defined in \eqref{eq_AAtPhiY_part}, and $\bar{\bf Z}^{-1}$ is defined as follows:
\begin{equation}\label{eq_ZbarInv}
\begin{aligned}
\bar{\bf Z}^{-1} \coloneqq {{\bf R}_{11}^\prime}^{-1} {{\bf R}_{11}^\prime}^{-T} =
({{\bf R}_{11}^\prime}^T {\bf R}_{11}^\prime)^{-1} = ({\bf I}+{\bf Z}_1^T{\bf Z}_1) ^{-1}.
\end{aligned}
\notag
\end{equation}
Writing ${\bf A}_k = {\bf U}[{\bf \Sigma}_1 \quad {\bf 0}\quad {\bf 0}]^T{\bf V}^T$, we obtain
\begin{equation}
\begin{aligned}
({\bf I} - {\bf Q}_1{\bf Q}_1^T){\bf A}_k = {\bf U}
	\begin{bmatrix}
	({\bf I} - \bar{\bf Z}^{-1}){\bf \Sigma}_1  \\
	{\bf 0} \\
	- {\bf Z}_1 \bar{\bf Z}^{-1}{\bf \Sigma}_1
	\end{bmatrix} {\bf V}^T.
\end{aligned}
\notag
\end{equation}
It follows that
\begin{equation}
\begin{aligned}\label{eq_2nIminQ1Q1tAk2}
\|({\bf I} - {\bf Q}_1{\bf Q}_1^T){\bf A}_k \|_2^2 & = \|{\bf \Sigma}_1 ({\bf I} - \bar{\bf Z}^{-1}){\bf \Sigma}_1\|_2 \\ & \le \|{\bf \Sigma}_1\|_2^2\|{\bf I} - \bar{\bf Z}^{-1}\|_2,
\end{aligned}
\end{equation}
where we have used the following relation that holds for any
matrix $\bf M$ with ${\bf I} +{\bf M}$ being non-singular \cite{HendersonS81}:
\begin{equation}
({\bf I} +{\bf M})^{-1} = {\bf I} - {\bf M}({\bf I} +{\bf M})^{-1} = {\bf I} - ({\bf I} +{\bf M})^{-1}{\bf M}.
\notag
\end{equation}
Let ${\bf K}\coloneqq {\bf I}-\bar{\bf Z} ^{-1}$. Matrix $\bf K$ is positive semidefinite, and its eigenvalues satisfy \cite[p.148]{StewartSun90}:
\begin{equation}
\begin{aligned}
\lambda_i({\bf K}) = \frac{{\sigma}_i^2({\bf Z}_1)}{1+{\sigma}_i^2({\bf Z}_1)}, \quad i=1,...k.
\end{aligned}
\notag
\end{equation}
The largest singular value of ${\bf Z}_1$ satisfies:
\begin{equation}\notag
 \sigma_1 ({\bf Z}_1) \le \delta_k^2\|{\bf \Phi}_2\|_2\|{\bf \Phi}_1^\dagger\|_2.
\end{equation}
Accordingly,
\begin{equation}
\begin{aligned}
\lambda_1({\bf K}) \le \frac{\delta_k^4\|{\bf \Phi}_2\|_2^2\|{\bf \Phi}_1^\dagger\|_2^2}{1+\delta_k^4\|{\bf \Phi}_2\|_2^2\|{\bf \Phi}_1^\dagger\|_2^2},
\end{aligned}
\notag
\end{equation}
Plugging this result into \eqref{eq_2nIminQ1Q1tAk2} and taking the square root, it follows
\begin{equation}\label{eq_2n1ndTermR22}
\begin{aligned}
\|({\bf I} - {\bf Q}_1{\bf Q}_1^T){\bf A}_k \|_2\le \frac{\delta_k^{2}\sigma_1\|{\bf \Phi}_2\|_2\|{\bf \Phi}_1^\dagger\|_2}{\sqrt{1 + \delta_k^{4}\|{\bf \Phi}_2\|_2^2\|{\bf \Phi}_1^\dagger\|_2^2}},
\end{aligned}
\end{equation}

For the second term on the right-hand side of \eqref{eq_2nR22_2terms}, we have
\begin{equation}
\begin{aligned}
\|({\bf I} - {\bf Q}_1{\bf Q}_1^T){\bf U}_{\perp}{\bf \Sigma}_\perp\|_2 & \le \|{\bf I} - {\bf Q}_1{\bf Q}_1^T\|_2\|{\bf U}_\perp\|_2\|{\bf \Sigma}_\perp\|_2\\ & \le \sigma_{k+1}.
\end{aligned}  
\label{eq_2n2ndTermR22} 
\end{equation}  
By substituting the results in \eqref{eq_2n1ndTermR22} and \eqref{eq_2n2ndTermR22} into \eqref{eq_2nR22_2terms}, the
theorem for the basic version of PbP-QLP follows.

When the PI scheme with power parameter $q$ is used, $\bf A$ is supplanted by $({\bf A}^T{\bf A})^q{\bf A}^T$, and as a result
\begin{equation}
\mathcal{R}({\bf D}) = \mathcal{R}({\bf A}\bar{\bf P})= \mathcal{R}({\bf A}({\bf A}^T{\bf A})^q{\bf A}^T{\bf \Phi}).
\notag
\end{equation}
Therefore, considering
\begin{equation}\notag
	{\bf A}({\bf A}^T{\bf A})^q{\bf A}^T{\bf \Phi} = {\bf U \Sigma}^{2q+2}{\bf U}^T {\bf \Phi},
\end{equation}
we now define a non-singular matrix $\bf Y$ as:
\begin{equation}\notag
{\bf Y} = \bigg[
{\bf \Phi}_1^\dagger
\begin{pmatrix}
{\bf \Sigma}_1^{2q+2} & {\bf 0}\\
{\bf 0} & {\bf \Sigma}_2^{2q+2}
\end{pmatrix}^{-1}, \bar{\bf Y}\bigg],
\end{equation}
and compute the product
\begin{equation}\notag
\begin{aligned}
{\bf A}({\bf A}^T{\bf A})^q{\bf A}^T{\bf \Phi Y}  =
{\bf U} \begin{bmatrix}
{\bf I} & {\bf 0} & {\bf 0} \\
{\bf 0} & {\bf I} & {\bf 0}\\
{\bf Z}_1 & {\bf Z}_2 & {\bf Z}_3
\end{bmatrix},
\end{aligned}
\label{}
\end{equation}
where ${\bf Z}_1 = {\bf \Sigma}_3^{2q+2}{\bf \Phi}_2{\bf \Phi}_1^\dagger {\bf \Sigma}_1^{-({2q+2})}$, ${\bf Z}_2 = {\bf \Sigma}_3^{2q+2}{\bf \Phi}_2{\bf \Phi}_1^\dagger {\bf \Sigma}_2^{-({2q+2})}$, and ${\bf Z}_3 = {\bf \Sigma}_3^{2q+2}{\bf \Phi}_2 \bar{\bf Y}$. Proceeding further with the procedure as described for $q=0$, the result for PI-incorporated PbP-QLP follows. \QEDB

\section{Proof of Theorem \ref{Th_boundssigmak}}
\label{Proo_boundssigmak}
The left inequality in \eqref{eq_Th_boundssigmak} is easily obtained by the Cauchy’s interlacing theorem \cite{StewartSun90}. To prove the right inequality, we exploit the following theorem from \cite{HornJohnson94}.
\begin{theorem}	(Horn and Johnson \cite{HornJohnson94}). Let ${\bf M}, {\bf N} \in \mathbb R^{n_1 \times n_2}$ and $\ell=min\{n_1, n_2\}$. Then for $1\le i, j \le \ell$ and $i+j \le \ell+1$,
	\begin{equation}\notag
	\sigma_i({\bf MN}) \le \sigma_1({\bf M})\sigma_i({\bf N}).
	\end{equation}
	\begin{equation}\notag
	\sigma_{i+j-1}({\bf M}+{\bf N}) \le \sigma_i({\bf M})+ \sigma_j({\bf N}).
	\end{equation}
	\begin{equation}
	\sigma_{i+j-1}({\bf MN}) \le \sigma_i({\bf M})\sigma_j({\bf N}).
	\label{eq1_HornJohn94}
	\end{equation} 	
	\label{Thr_HornJohn94}
\end{theorem}

By the definition of ${\bf D}_1$ \eqref{eq_D1}, we have
\begin{equation}\label{eq_R11_exp}
\begin{aligned}
{\bf R}_{11} & = {\bf Q}_1^T{\bf D}_1 = {\bf Q}_1^T({\bf U}_k{\bf \Sigma}_k{\bf X}_{11} + {\bf U}_{\perp}{\bf \Sigma}_{\perp}{\bf X}_{21})\\ &
= {\bf Q}_1^T{\bf U}_k{\bf \Sigma}_k{\bf X}_{11} + {\bf Q}_1^T{\bf U}_{\perp}{\bf \Sigma}_{\perp}{\bf X}_{21}.
\end{aligned}
\end{equation}
Exploiting Theorem \ref{Thr_HornJohn94} results in
\begin{equation}\label{eq_sigmak_R11_proo}
\begin{aligned}
	\sigma_k({\bf R}_{11}) & \le \sigma_k({\bf Q}_1^T{\bf U}_k{\bf \Sigma}_k{\bf X}_{11}) + \sigma_1({\bf Q}_1^T{\bf U}_{\perp}{\bf \Sigma}_{\perp}{\bf X}_{21})\\
& \le \sigma_k({\bf Q}_1^T{\bf U}_k{\bf \Sigma}_k{\bf X}_{11}) + \sigma_{k+1},
\end{aligned}
\end{equation}
and $\sigma_k$ of ${\bf Q}_1^T{\bf U}_k{\bf \Sigma}_k{\bf X}_{11}$ is bounded above by:
\begin{equation}\notag
\begin{aligned}
\sigma_k({\bf Q}_1^T{\bf U}_k{\bf \Sigma}_k{\bf X}_{11}) & \le \sigma_1({\bf Q}_1^T)\sigma_k({\bf U}_k{\bf \Sigma}_k{\bf X}_{11})\\ & \le \sigma_1({\bf U}_k)\sigma_k({\bf \Sigma}_k{\bf X}_{11}) \le
\sigma_k.
\end{aligned}
\end{equation}
The last relation follows by applying \eqref{eq1_HornJohn94} with $i=k$ and $j=1$. Plugging the result into \eqref{eq_sigmak_R11_proo} gives
\begin{equation}\notag
	\sigma_k({\bf R}_{11}) \le \sigma_k + \sigma_{k+1},
\end{equation}
which completes the proof. \QEDB

\section{Proof of Theorem \ref{Th_distUkVkQ1P1}}
\label{Proo_distance}
We provide proofs for the case the PI is used (for the simple form of PbP-QLP, $q = 0$). According to Theorem 2.6.1 of \cite{GolubVanLoan96}, we have for the distance between $\mathcal{R}({\bf U}_k)$ and $\mathcal{R}({\bf Q}_1)$
\begin{equation}
\begin{aligned}
\text{dist}(\mathcal{R}({\bf U}_k), \mathcal{R}({\bf Q}_1))  = \|{\bf U}_k{\bf U}_k^T - {\bf Q}_1{\bf Q}_1^T\|_2  = \|{\bf U}_\perp^T{\bf Q}_1\|_2.
\label{eq_dist_def}
\end{aligned}
\end{equation}
For the range of $\bf Q$, we have
\begin{equation}\notag
\mathcal{R}({\bf Q}) = \mathcal{R}({\bf D}) = \mathcal{R}({\bf A}\bar{\bf P})= \mathcal{R}({\bf A}({\bf A}^T{\bf A})^q{\bf A}^T{\bf \Phi}).
\label{}
\end{equation}
We therefore have
\begin{equation}\notag
\begin{aligned}
{\bf A}({\bf A}^T{\bf A})^q{\bf A}^T{\bf \Phi} & = {\bf U}\begin{bmatrix}
{\bf \Sigma}_k^{2q+2} & {\bf 0}  \\
{\bf 0} & {\bf \Sigma}_\perp^{2q+2}
\end{bmatrix} \begin{bmatrix}
{\bf \Phi}_{11} & {\bf \Phi}_{12} \\
{\bf \Phi}_{21} & {\bf \Phi}_{22}
\end{bmatrix}\\
& = [{\bf Q}_1\quad {\bf Q}_2]
\begin{bmatrix}
{\bf R}_{11} & {\bf R}_{12}  \\
{\bf 0} & {\bf R}_{22}
\end{bmatrix}.
\end{aligned}
\end{equation}
Let 
\begin{equation}\notag
\begin{aligned}
{\bf H}\coloneqq {\bf U}\begin{bmatrix}
{\bf \Sigma}_k^{2q+2} & {\bf 0}  \\
{\bf 0} & {\bf \Sigma}_\perp^{2q+2}
\end{bmatrix} \begin{bmatrix}
{\bf \Phi}_{11} \\
{\bf \Phi}_{21}
\end{bmatrix} = {\bf Q}_1{\bf R}_{11}.
\end{aligned}
\end{equation}
We now define a matrix ${\bf J} \in \mathbb R^{k \times k}$ as:
\begin{equation}\notag
	{\bf J} \coloneqq {\bf \Phi}_{11}^{-1}{\bf \Sigma}_k^{-(2q+2)},
\end{equation}
and compute a QR factorization of the product $\bf HJ$:
\begin{equation}\notag
\begin{aligned}
{\bf HJ}= {\bf U}\begin{bmatrix}
{\bf I}_k  \\
{\bf \Sigma}_\perp^{2q+2}{\bf \Phi}_{21}{\bf \Phi}_{11}^{-1} {\bf \Sigma}_k^{-(2q+2)}
\end{bmatrix} = \ddot{\bf Q}\ddot{\bf R},
\end{aligned}
\end{equation}
which gives

\begin{equation}\notag
\begin{aligned}
\begin{bmatrix}
{\bf I}_k  \\
{\bf \Sigma}_\perp^{2q+2}{\bf \Phi}_{21}{\bf \Phi}_{11}^{-1} {\bf \Sigma}_k^{-(2q+2)}
\end{bmatrix} = \begin{bmatrix}
{\bf U}_k^T \\
{\bf U}_\perp^T
\end{bmatrix} \ddot{\bf Q}\ddot{\bf R}.
\end{aligned}
\end{equation}

From ${\bf U}_k^T\ddot{\bf Q}\ddot{\bf R} = {\bf I}_k$, it is derived that $\ddot{\bf R}^{-1} = {\bf U}_k^T\ddot{\bf Q}$. Accordingly,
\begin{equation}\notag
{\bf U}_\perp^T \ddot{\bf Q} = {\bf \Sigma}_\perp^{2q+2}{\bf \Phi}_{21}{\bf \Phi}_{11}^{-1} {\bf \Sigma}_k^{-(2q+2)}{\bf U}_k^T\ddot{\bf Q}.
\end{equation}
From \eqref{eq_dist_def}, it follows that
\begin{equation}\notag
\|{\bf U}_\perp^T \ddot{\bf Q}\|_2 \coloneqq  \|{\bf U}_k{\bf U}_k^T - \ddot{\bf Q}\ddot{\bf Q}^T\|_2
\end{equation}
By \cite[Lemma 4.1]{Gu2015}, ${\bf Q}_1{\bf Q}_1^T=\ddot{\bf Q}\ddot{\bf Q}^T$. As a result,
\begin{equation}\notag
\begin{aligned}
\text{dist}(\mathcal{R}({\bf U}_k), \mathcal{R}({\bf Q}_1)) & =
\|{\bf U}_\perp^T \ddot{\bf Q}\|_2  \\ & \le \|{\bf \Sigma}_\perp^{2q+2}{\bf \Phi}_{21}{\bf \Phi}_{11}^{-1} {\bf \Sigma}_k^{-(2q+2)}\|_2\|{\bf U}_k^T\ddot{\bf Q}\|_2,
\end{aligned}
\end{equation}
from which the theorem for \eqref{eq_Th_distance1} follows. 

To prove \eqref{eq_Th_distance2}, since ${\bf P} = \bar{\bf P}\widetilde{\bf P}$, we have
\begin{equation}\notag
\mathcal{R}({\bf P}) = \mathcal{R}(\bar{\bf P})= \mathcal{R}(({\bf A}^T{\bf A})^q{\bf A}^T{\bf \Phi}).
\end{equation}
The first relation follows because $\widetilde{\bf P}$ is an orthogonal matrix of order $d$. Thus, it suffices to bound from above $\text{dist}(\mathcal{R}({\bf V}_k), \mathcal{R}(\bar{\bf P}_1))$, where $\bar{\bf P}_1$ is the first $k$ columns of $\bar{\bf P}$. The rest of the proof is similat to that of \eqref{eq_Th_distance1}, and we therefore omit it.
\QEDB
 
\section{Proof of Theorem \ref{Th_LRapprox}}
\label{Proo_LRapprox}
For the left-hand side term in \eqref{eq_2nLRAErrbound}, we write
\begin{equation}\label{eq_LRAErr_proof}
\begin{aligned}
\|({\bf I} - {\bf QQ}^T){\bf A}\|_2 & \le \|({\bf I} - {\bf QQ}^T){\bf A}_k\|_2 + \|({\bf I} - {\bf QQ}^T){\bf A}_\perp\|_2\\ & \le \|({\bf I} - {\bf Q}_1{\bf Q}_1^T){\bf A}_k\|_2 + \|{\bf \Sigma}_\perp\|_2.
\end{aligned}
\end{equation}
In the last relation, the first term on the right-hand side results from the fact that $\mathcal{R}({\bf Q}_1) \subset \mathcal{R}({\bf Q})$, which for any matrix $\bf A$ yields \cite[Proposition 8.5]{HMT2011}:
\begin{equation}\notag
\|({\bf I} - {\bf QQ}^T){\bf A}\|_2 \le \|({\bf I} - {\bf Q}_1{\bf Q}_1^T){\bf A}\|_2.
\end{equation}
By substituting the bound in \eqref{eq_2n1ndTermR22} into \eqref{eq_LRAErr_proof}, the theorem follows for the case $q = 0$. The theorem for the case when the PI technique is used follows similarly. \QEDB

\section{Proof of Theorem \ref{Theorem_SinValueApprox}}
\label{Proo_SinValueApprox1}

First, we write ${\bf D}= {\bf Q}{\bf Q}^T{\bf D}$. It follows that
\begin{equation}\notag
\begin{aligned}
 {\bf D}^T{\bf D} & = {\bf D}^T{\bf Q}{\bf Q}^T{\bf D}\\
 {\bf D}^T{\bf D} & \succeq {\bf D}^T{\bf Q}_1{\bf Q}_1^T{\bf D}.
\end{aligned}
\end{equation}
By the Cauchy’s interlacing theorem, we therefore have

\begin{equation}\notag
\lambda_i({\bf A}^T{\bf A}) \ge \lambda_i({\bf D}^T{\bf D}) \ge \lambda_i({\bf D}^T{\bf Q}_1{\bf Q}_1^T{\bf D}).
\end{equation}
Furthermore, we have ${\bf Q}_1{\bf Q}_1^T = {\bf Q}_1^\prime {{\bf Q}_1^\prime}^T$. Thus, by replacing $\bf D$ and ${\bf Q}_1^\prime$, for the last term of the above equation we obtain
\begin{equation}\notag
\begin{aligned}
 & {\bf D}^T{\bf Q}_1^\prime {{\bf Q}_1^\prime}^T{\bf D} = \bar{\bf P}{\bf A}^T{\bf Q}_1^\prime {{\bf Q}_1^\prime}^T{\bf A}\bar{\bf P}\\ &
 = \bar{\bf P}{\bf V}^T
 \begin{bmatrix}
{\bf \Sigma}_1\bar{\bf Z}^{-1}{\bf \Sigma}_1 & {\bf 0} & {\bf \Sigma}_1\bar{\bf Z}^{-1}{\bf Z}_1^T{\bf \Sigma}_3 \\
{\bf 0} & {\bf 0} & {\bf 0}\\
{\bf \Sigma}_3{\bf Z}_1 \bar{\bf Z}^{-1}{\bf \Sigma}_1 & {\bf 0} &
{\bf \Sigma}_3{\bf Z}_1 \bar{\bf Z}^{-1}{\bf Z}_1^T{\bf \Sigma}_3
\end{bmatrix}{\bf V}^T\bar{\bf P}.
\end{aligned}
\end{equation}
It is seen that ${\bf \Sigma}_1\bar{\bf Z}^{-1}{\bf \Sigma}_1 $ is a submatrix of ${\bf D}^T{\bf Q}_1^\prime {{\bf Q}_1^\prime}^T{\bf D}$. By replacing $\bar{\bf Z}^{-1}$, we thus obtain the following relation:
\begin{equation}\notag
\begin{aligned}
\lambda_i({\bf A}^T{\bf A}) \ge \lambda_i({\bf D}^T{\bf D}) & \ge \lambda_i({\bf D}^T{\bf Q}_1{\bf Q}_1^T{\bf D})\\
& \ge \lambda_i({\bf \Sigma}_1({\bf I} + {\bf Z}_1^T{\bf Z}_1)^{-1}{\bf \Sigma}_1).
\end{aligned}
\end{equation}
                                                                                                                                                                            By applying the properties of partial ordering, it follows that
                                                                                                                                                                            \begin{equation}\notag
                                                                                                                                                                            \begin{aligned}
                                                                                                                                                                            {\bf Z}_1^T{\bf Z}_1 & \preceq \sigma_{d-p+1}^{4q+4}\|{\bf \Phi}_2\|_2^2\|{\bf \Phi}_1^\dagger\|_2^2 {\bf \Sigma}_1^{-(4q+4)} \\
                                                                                                                                                                            & = {\bf \Delta}^{4q+4}\|{\bf \Phi}_2\|_2^2\|{\bf \Phi}_1^\dagger\|_2^2,
                                                                                                                                                                            \end{aligned}
                                                                                                                                                                            \end{equation}
                                                                                                                                                                            where ${\bf \Delta} = \text{diag}(\delta_1, ..., \delta_k)$ is a $k \times k$ matrix with entries $\delta_i= \frac{\sigma_{d-p+1}}{\sigma_i}$. In addition, the following relation holds
                                                                                                                                                                            \begin{equation}\notag
                                                                                                                                                                            {\bf \Sigma}_1({\bf I} + {\bf Z}_1^T{\bf Z}_1)^{-1}{\bf \Sigma}_1 \succeq {\bf \Sigma}_1({\bf I} + {\bf \Delta}^{4q+4}\|{\bf \Phi}_2\|_2^2\|{\bf \Phi}_1^\dagger\|_2^2)^{-1}{\bf \Sigma}_1,
                                                                                                                                                                            \end{equation}
                                                                                                                                                                            which results in
                                                                                                                                                                            \begin{equation}\notag
                                                                                                                                                                            \begin{aligned}
                                                                                                                                                                            \lambda_i({\bf A}^T{\bf A}) \ge \lambda_i({\bf D}^T{\bf D})  & \ge \lambda_i({\bf \Sigma}_1({\bf I} + {\bf Z}_1^T{\bf Z}_1)^{-1}{\bf \Sigma}_1) \\ & \ge \frac{\sigma_i^2}{1 + \delta_i^{4q+4}\|{\bf \Phi}_2\|_2^2\|{\bf \Phi}_1^\dagger\|_2^2}.
                                                                                                                                                                            \end{aligned}
                                                                                                                                                                            \end{equation}
By taking the square root of the last identity, the theorem
follows. \QEDB

\section{Proof of Theorem \ref{Th_SinValueApprox2}}
\label{Proo_SinValueApprox2}
According to Theorem 2.1 of  \cite{HuckabyChan05}, the singular values of ${\bf L}_{11}$ and $\bf D$ are related as follows:
\begin{equation} \label{eq_sL11si}
\begin{aligned}
\frac{\sigma_i({\bf L}_{11})}{\sigma_i({\bf D})} \ge 1 - \mathcal{O}\Bigg(\frac{\|{\bf L}_{12}\|_2^2}{(1-\rho^2)\sigma_k^2({\bf L}_{11})}\Bigg), \quad i=1,..., k.
\end{aligned}
\end{equation}
We also have the following relation:
\begin{equation}\label{eq_quotientMult}
\begin{aligned}
\frac{\sigma_i({\bf L}_{11})}{\sigma_i}= \frac{\sigma_i({\bf L}_{11})}{\sigma_i({\bf D})} \times \frac{\sigma_i({\bf D})}{\sigma_i}.
\end{aligned}
\end{equation}
By substituting bounds \eqref{eq_Th_SinValD_bound} and \eqref{eq_sL11si} into \eqref{eq_quotientMult}, the theorem follows.\QEDB

\section{Proof of Theorem \ref{Th_boundsA'ssigmak}}
\label{Proo_boundsA'ssigmak}
Let ${\bf A}_k$ and $\hat{\bf A}_k$ be rank-$k$ approximations of $\bf A$ constructed by the SVD and PbP-QLP, respectively. From the perturbation theory for singular values 
\cite[Corollary 7.3.5]{HornJohnson852012}, for $i=1,...,k$, we have
\begin{equation}\notag      
\begin{aligned}
|\sigma_i - \sigma_i({\bf L}_{11})| \le \|{\bf A}_k - \hat{\bf A}_k\|_2.
\end{aligned}
\end{equation}
Writing ${\bf A}_k={\bf U}_k{\bf U}_k^T{\bf A}$, $\hat{\bf A}_k={\bf Q}_1{\bf Q}_1^T{\bf A}$, and applying Theorem \ref{Th_distUkVkQ1P1}, we obtain
\begin{equation}\notag 
|\sigma_i - \sigma_i({\bf L}_{11})| \le \|{\bf A}\|_2\text{dist}(\mathcal{R}({\bf U}_k), \mathcal{R}({\bf Q}_1)),
\end{equation}
from which the theorem follows. \QEDB

\section{Proof of Theorem \ref{Th_SinVal_PbP}}
\label{Proo_SinVal_PbP}
To prove this theorem, we use a key result from \cite[Theorem 5.8]{Gu2015}, stated with our notation as follows:
\begin{equation}\label{eq_HighProGu}
\mathbb P\{\|{\bf \Phi}_2\|_2\|{\bf \Phi}_1^\dagger\|_2\ge C_\Upsilon\}\le \Upsilon.
\end{equation}

To prove \eqref{eq_HighProB_R22}, we observe that \eqref{eq_2nR22_bound} is simplified to
\begin{equation}\notag 
\begin{aligned}
\|{\bf R}_{22}\|_2 \le \sigma_{k+1} + {\delta_k^{2q+2}\sigma_1\|{\bf \Phi}_2\|_2\|{\bf \Phi}_1^\dagger\|_2}.
\end{aligned}
\end{equation}
Substituting the bound in \eqref{eq_HighProGu} into the above equation gives the desired result. The bounds \eqref{eq_HighProB_SVs} and \eqref{eq_HighProB_LRAE} are likewise obtained, by plugging \eqref{eq_HighProGu} into \eqref{eq_sigmaL11sigmai} and \eqref{eq_2nLRAErrbound}, respectively. \QEDB

\bibliographystyle{IEEEtran}
\bibliography{mybibfile}

\begin{thebibliography}{10}
\providecommand{\url}[1]{#1}
\csname url@samestyle\endcsname
\providecommand{\newblock}{\relax}
\providecommand{\bibinfo}[2]{#2}
\providecommand{\BIBentrySTDinterwordspacing}{\spaceskip=0pt\relax}
\providecommand{\BIBentryALTinterwordstretchfactor}{4}
\providecommand{\BIBentryALTinterwordspacing}{\spaceskip=\fontdimen2\font plus
\BIBentryALTinterwordstretchfactor\fontdimen3\font minus
  \fontdimen4\font\relax}
\providecommand{\BIBforeignlanguage}[2]{{%
\expandafter\ifx\csname l@#1\endcsname\relax
\typeout{** WARNING: IEEEtran.bst: No hyphenation pattern has been}%
\typeout{** loaded for the language `#1'. Using the pattern for}%
\typeout{** the default language instead.}%
\else
\language=\csname l@#1\endcsname
\fi
#2}}
\providecommand{\BIBdecl}{\relax}
\BIBdecl

\bibitem{FazelPST13}
M.~Fazel, T.~K. Pong, D.~Sun, and P.~Tseng, ``{Hankel matrix rank minimization
  with applications to system identification and realization},'' \emph{SIAM. J.
  Matrix Anal. \& Appl.}, vol.~34, no.~3, pp. 946--977, Apr 2013.

\bibitem{CaiLi15}
T.~{Cai} and X.~{Li}, ``{Robust and computationally feasible community
  detection in the presence of arbitrary outlier nodes},'' \emph{The Annals of
  Statistics}, vol.~43, no.~3, pp. 1027--1059, 2015.

\bibitem{EftekhariYW18}
A.~{Eftekhari}, D.~{Yang}, and M.~B. {Wakin}, ``Weighted matrix completion and
  recovery with prior subspace information,'' \emph{IEEE Transactions on
  Information Theory}, vol.~64, no.~6, pp. 4044--4071, June 2018.

\bibitem{AybatI15}
N.~S. Aybat and G.~Iyengar, ``{An alternating direction method with increasing
  penalty for stable principal component pursuit},'' \emph{Comput Optim Appl},
  vol.~61, no.~3, p. 635–668, Jul 2015.

\bibitem{MFKDeJSTSP18}
M.~F. Kaloorazi and R.~C. de~Lamare, ``{Compressed randomized UTV
  decompositions for low-rank matrix approximations},'' \emph{IEEE J. Sel.
  Topics Signal Process.}, vol.~12, no.~6, pp. 1--15, Dec 2018.

\bibitem{ClarWood2017}
K.~L. Clarkson and D.~P. Woodruff, ``{Low-rank approximation and regression in
  input sparsity time},'' \emph{J. ACM}, vol.~63, no.~6, pp. 54:1--54:45, Jan.
  2017.

\bibitem{Rasti18}
B.~{Rasti}, P.~{Scheunders}, P.~{Ghamisi}, G.~{Licciardi}, and J.~{Chanussot},
  ``{Noise reduction in hyperspectral imagery: Overview and application},''
  \emph{Remote Sens.}, vol.~10, no.~3, 2018.

\bibitem{ErichsonMWK18}
N.~B. {Erichson}, A.~{Mendible}, S.~{Wihlborn}, and J.~N. {Kutz}, ``{Randomized
  nonnegative matrix factorization},'' \emph{Pattern Recognition Letters}, vol.
  104, pp. 1--7, 2018.

\bibitem{FowlerDu12}
J.~E. {Fowler} and Q.~{Du}, ``Anomaly detection and reconstruction from random
  projections,'' \emph{IEEE Transactions on Image Processing}, vol.~21, no.~1,
  pp. 184--195, Jan 2012.

\bibitem{KaDeICASSP17}
M.~F. Kaloorazi and R.~C. de~Lamare, ``{Anomaly detection in IP networks based
  on randomized subspace methods},'' in \emph{ICASSP, USA}, Mar 2017, pp.
  4222--4226.

\bibitem{ChenRS7859344}
J.~Chen, C.~Richard, and A.~H. Sayed, ``{Multitask diffusion adaptation over
  networks with common latent representations},'' \emph{IEEE J. Sel. Topics
  Signal Process.}, vol.~11, no.~3, pp. 563--579, Apr 2017.

\bibitem{DarnellGME17}
G.~{Darnell}, S.~{Georgiev}, S.~{Mukherjee}, and B.~E. {Engelhardt},
  ``{Adaptive randomized dimension reduction on massive data},'' \emph{JMLR},
  vol.~18, pp. 1--30, 2017.

\bibitem{UbaruSaad19}
S.~{Ubaru} and Y.~{Saad}, ``{Sampling and multilevel coarsening algorithms for
  fast matrix approximations},'' \emph{Numerical Linear Algebra with
  Applications}, vol.~26, no.~3, 2019.

\bibitem{7038247}
A.~{Cichocki}, D.~{Mandic}, L.~{De Lathauwer}, G.~{Zhou}, Q.~{Zhao},
  C.~{Caiafa}, and H.~A. {Phan}, ``Tensor decompositions for signal processing
  applications: From two-way to multiway component analysis,'' \emph{IEEE
  Signal Processing Magazine}, vol.~32, no.~2, pp. 145--163, Mar 2015.

\bibitem{DavisRS2016}
T.~A. Davis, S.~Rajamanickam, and W.~M. Sid-Lakhdar, ``{A survey of direct
  methods for sparse linear systems},'' \emph{Acta Numerica}, vol.~25, pp.
  383--566, 2016.

\bibitem{GolubVanLoan96}
G.~H. Golub and C.~F. van Loan, \emph{{Matrix computations}}, 3rd ed., Johns
  Hopkins Univ. Press, Baltimore, MD, 1996.

\bibitem{Chan87}
T.~F. Chan, ``{Rank revealing QR factorizations},'' \emph{Linear Algebra and
  its Applications}, vol. 88-89, pp. 67--82, Apr 1987.

\bibitem{StewartQLP}
G.~W. Stewart, ``{The QLP approximation to the singular value decomposition},''
  \emph{SIAM J. Sci. Comput.}, vol.~20, no.~4, pp. 1336--1348, 1999.

\bibitem{DemGGX15}
J.~Demmel, L.~Grigori, M.~Gu, and H.~Xiang, ``{Communication avoiding rank
  revealing QR factorization with column pivoting},'' \emph{SIAM J. Matrix
  Anal. \& Appl.}, vol.~36, no.~1, pp. 55--89, 2015.

\bibitem{DuerschGu2017}
J.~A. Duersch and M.~Gu, ``{Randomized QR with column pivoting},'' \emph{SIAM
  J. Sci. Comput.}, vol.~39, no.~4, pp. C263--C291, 2017.

\bibitem{Dongarraetal18}
J.~{Dongarra}, M.~{Gates}, A.~{Haider}, J.~{Kurzak}, P.~{Luszczek}, S.~{Tomov},
  and I.~{Yamazaki}, ``{The singular value decomposition: Anatomy of optimizing
  an algorithm for extreme scale},'' \emph{SIAM Rev}, vol.~60, no.~4, p.
  808–865, 2018.

\bibitem{MartinssonQH2019}
P.~Martinsson, G.~Quintana-Ort\'{\i}, and N.~Heavner, ``{randUTV: A blocked
  randomized algorithm for computing a rank-revealing UTV factorization},''
  \emph{ACM Trans. Math. Softw.}, vol.~45, no.~1, pp. 4:1--4:26, Mar. 2019.

\bibitem{Rokhlin09}
V.~Rokhlin, A.~Szlam, and M.~Tygert, ``{A randomized algorithm for principal
  component analysis},'' \emph{SIAM. J. Matrix Anal. \& Appl.}, vol.~31, no.~3,
  pp. 1100--1124, 2009.

\bibitem{HMT2011}
N.~Halko, P.-G. Martinsson, and J.~Tropp, ``{Finding structure with randomness:
  Probabilistic algorithms for constructing approximate matrix
  decompositions},'' \emph{SIAM Review}, vol.~53, no.~2, pp. 217--288, Jun
  2011.

\bibitem{Gu2015}
M.~Gu, ``{Subspace iteration randomization and singular value problems},''
  \emph{SIAM J. Sci. Comput.}, vol.~37, no.~3, pp. A1139--A1173, 2015.

\bibitem{MFKDeTSP18}
M.~F. Kaloorazi and R.~C. de~Lamare, ``{Subspace-orbit randomized decomposition
  for low-rank matrix approximations},'' \emph{IEEE Trans. Signal Process.},
  vol.~66, no.~16, pp. 4409--4424, Aug 2018.

\bibitem{Martinsson18}
P.-G. Martinsson, ``{Randomized methods for matrix computations},'' \emph{The
  Mathematics of Data, IAS/Park City Mathematics Series}, vol.~25, no.~4, pp.
  187--231, 2018.

\bibitem{Saibaba19}
A.~K. Saibaba, ``{Randomized subspace iteration: Analysis of canonical angles
  and unitarily invariant norms},'' \emph{SIAM. J. Matrix Anal. \& Appl.},
  vol.~40, no.~1, p. 23–48, Jan 2019.

\bibitem{MFKJC19}
M.~F. Kaloorazi and J.~Chen, ``{Randomized truncated pivoted QLP factorization
  for low-rank matrix recovery},'' \emph{IEEE Signal Processing Letters},
  vol.~26, no.~7, pp. 1075--1079, Jul 2019.

\bibitem{Stewart98}
G.~W. Stewart, \emph{{Matrix algorithms: volume 1: basic decompositions}},
  SIAM, Philadelphia, PA, 1998.

\bibitem{MathiasStewart93}
R.~Mathias and S.~G. W., ``{A block QR algorithm and the singular value
  decomposition},'' \emph{Linear Algebra and its App.}, vol. 182, pp. 91--100,
  1993.

\bibitem{HuckabyChan05}
D.~A. Huckaby and T.~F. Chan, ``{Stewart's pivoted QLP decomposition for
  low-rank matrices},'' \emph{Numer. Linear Algebra Appl.}, vol.~12, p.
  153–159, 2005.

\bibitem{FierroHan97}
R.~D. Fierro and P.~C. Hansen, ``{Low-rank revealing UTV decompositions},''
  \emph{Numerical Algorithms}, vol.~15, no.~1, pp. 37--–55, Jul 1997.

\bibitem{GoreinovTZ97}
S.~Goreinov, E.~Tyrtyshnikov, and N.~Zamarashkin, ``{A theory of pseudoskeleton
  approximations},'' \emph{Linear Algebra and its Applications}, vol. 261, pp.
  1--21, Aug 1997.

\bibitem{FriezeKVS04}
A.~Frieze, R.~Kannan, and S.~Vempala, ``{Fast Monte-Carlo algorithms for
  finding low-rank approximations},'' \emph{J. ACM}, vol.~51, no.~6, pp.
  1025--1041, Nov. 2004.

\bibitem{MahDri09}
M.~{Mahoney} and P.~{Drineas}, ``{CUR matrix decompositions for improved data
  analysis},'' \emph{PNAS}, vol. 106, no.~3, pp. 697--702, Jan 2009.

\bibitem{DeshpandeV2006}
A.~Deshpande and S.~Vempala, ``{Adaptive sampling and fast low-rank matrix
  approximation},'' \emph{Diaz J., Jansen K., Rolim J.D.P., Zwick U. (eds)
  Approximation, Randomization, and Combinatorial Optimization. Algorithms and
  Techniques}, vol. 4110, pp. 292--303, 2006.

\bibitem{RudelsonV07}
M.~Rudelson and R.~Vershynin, ``{Sampling from large matrices: An approach
  through geometric functional analysis},'' \emph{J. ACM}, vol.~54, no.~4, Jul.
  2007.

\bibitem{Friedland11}
S.~{Friedland}, V.~{Mehrmann}, A.~{Miedlar}, and M.~{Nkengla}, ``{Fast low rank
  approximations of matrices and tensors},'' \emph{ELA}, vol.~22, pp.
  1031--1048, Oct 2011.

\bibitem{BoutsidisWood17}
C.~Boutsidis and D.~P. Woodruff, ``{Optimal CUR matrix decompositions},''
  \emph{SIAM J. Comput.}, vol.~46, no.~2, pp. 543--589, 2017.

\bibitem{Sarlos06}
T.~Sarl{\'{o}}s, ``{Improved approximation algorithms for large matrices via
  random projections},'' in \emph{47th Ann. IEEE Symp. on Foundations of
  Computer Science. FOCS '06.}, vol.~1, Oct. 2006.

\bibitem{PapaRTV00}
C.~Papadimitrioua, P.~Raghvan, H.~Tamaki, and S.~Vempalad, ``{Latent semantic
  indexing: A probabilistic analysis},'' \emph{J. of Comput. System Sci.},
  vol.~61, no.~2, pp. 217--235, Oct 2000.

\bibitem{NelsonNg2013}
J.~Nelson and H.~L. Nguyen, ``{OSNAP: Faster numerical linear algebra
  algorithms via sparser subspace embeddings},'' in \emph{Proc. of 54th Annual
  Symp. on FOCS '13}, 2013, pp. 117--126.

\bibitem{AilonCh09}
N.~Ailon and B.~Chazelle, ``{The fast Johnson-Lindenstrauss transform and
  approximate nearest neighbors},'' \emph{SIAM J. Comput.}, vol.~39, no.~1, pp.
  302--322, 2009.

\bibitem{XiaoGL17}
J.~{Xiao}, M.~{Gu}, and J.~{Langou}, ``{Fast parallel randomized QR with column
  pivoting algorithms for reliable low-rank matrix approximations},'' in
  \emph{24th IEEE International Conference on High Performance Computing
  (HiPC)}, Dec. 2017.

\bibitem{FengXG19}
Y.~{Feng}, J.~{Xiao}, and M.~{Gu}, ``{Flip-flop spectrum-revealing QR
  factorization and its applications on singular value decomposition},''
  \emph{Electronic Transactions on Numerical Analysis}, vol.~51, pp. 469--494,
  2019.

\bibitem{WuX20}
N.~{Wu} and H.~{Xiang}, ``{Randomized QLP decomposition,},'' \emph{Linear
  Algebra and its App.}, vol. 599, p. 18–35, Aug 2020.

\bibitem{VoroninMartin16}
S.~Voronin and P.-G. Martinsson, ``{A randomized blocked algorithm for
  efficiently computing rank-revealing factorizations of matrices},''
  \emph{SIAM J. Sci. Comput.}, vol.~38, no.~5, p. S458–S507, 2016.

\bibitem{YuLi18}
W.~{Yu} and Y.~{Li}, ``{Efficient randomized algorithms for the fixed-precision
  low-rank matrix approximation},'' \emph{SIAM J. Matrix Anal. \& Appl.},
  vol.~39, no.~3, pp. 1339--1359, 2018.

\bibitem{ChandrasekaranIpsen94}
S.~{Chandrasekaran} and I.~C.~F. {Ipsen}, ``{On rank-revealing QR
  factorizations},'' \emph{SIAM J. Matrix Anal. \& Appl.}, vol.~15, no.~2, pp.
  592--622, 1994.

\bibitem{Edelman88}
A.~{Edelman}, ``{Eigenvalues and Condition Numbers of Random Matrices},''
  \emph{SIAM J. Matrix Anal. \& Appl.}, vol.~9, no.~4, pp. 543--560, 1988.

\bibitem{Szarek91}
S.~{Szarek}, ``{Condition Numbers of Random Matrices},'' \emph{J. Complex},
  vol.~7, pp. 131--149, 1991.

\bibitem{Hansen_RegTools}
P.~Hansen, \emph{{Regularization tools version 4.1 for matlab 7.3.}},
  http://www.imm.dtu.dk/~pcha/Regutools.

\bibitem{Ayala19}
A.~{Ayala}, X.~{Claeys}, and L.~{Grigori}, ``{ALORA: Affine low-rank
  approximations},'' \emph{J Sci Comput}, vol.~79, pp. 1135--1160, 2019.

\bibitem{HendersonS81}
H.~{Henderson} and R.~{Searle}, ``{On deriving the inverse of a sum of
  matrices},'' \emph{SIAM Rev}, vol.~23, no.~1, pp. 53--60, 1981.

\bibitem{StewartSun90}
G.~W. Stewart and J.-g. Sun, \emph{{Matrix perturbation theory}}, Academic
  Press 1990.

\bibitem{HornJohnson94}
R.~A. Horn and C.~R. Johnson, \emph{{Topics in matrix analysis}}, Cambridge
  Univ. Press, 1994.

\bibitem{HornJohnson852012}
------, \emph{{Matrix analysis}}, 2nd ed., Cambridge Univ. Press, 2012.

\end{thebibliography}
%\bibliography{iqi}
\end{document}